\documentclass[a4paper,fleqn]{cas-dc}

\usepackage{amsmath}
\usepackage[numbers]{natbib}
\usepackage{subfig}
\usepackage[percent]{overpic}
\usepackage{picins}
\def\tsc#1{\csdef{#1}{\textsc{\lowercase{#1}}\xspace}}
\tsc{WGM}
\tsc{QE}
\begin{document}
\let\WriteBookmarks\relax
\def\floatpagepagefraction{1}
\def\textpagefraction{.001}

\pagestyle{plain}	% 去除页眉
%% Short title
%\shorttitle{short title of the paper for running head}    

%% Short author
%\shortauthors{short author list for running head}  

% Main title of the paper
\title[mode = title]{Adaptive Weighted Guided Image Filtering for Depth Enhancement in Shape-From-Focus}

%\tnotemark[1]
%\tnotetext[1]{}

\author[1,4]{Yuwen Li}
\ead{yuwen.li@nit.edu.cn} 
\author[2]{Zhengguo Li}
\ead{ezgli@i2r.a-star.edu.sg}
\author[3]{Chaobing Zheng}
\ead{zhengchaobing@wust.edu.cn}
\author[3]{Shiqian Wu}
\cormark[1] 
\ead{shiqian.wu@wust.edu.cn} 

\address[1]{Hubei Key Laboratory of Mechanical Transmission and Manufacturing Engineering, Wuhan University of Science and Technology, Wuhan 430081, China}
\address[2]{SRO Department, Institute for Infocomm Research, 138632, Singapore}
\address[3]{Institute of Robotics and Intelligent Systems, School of Information Science and Engineering, Wuhan University of Science and Technology, Wuhan 430081, China}
\address[4]{School of Electrical Engineering, Nanchang Institute of Technology, Nanchang 330099, China}

\cortext[cor1]{Corresponding author}  

% Here goes the abstract
\begin{abstract}
Existing shape from focus (SFF) techniques cannot preserve depth edges and fine structural details from a sequence of multi-focus images. Moreover, noise in the sequence of multi-focus images affects the accuracy of the depth map. In this paper, a novel depth enhancement algorithm for the SFF based on an adaptive weighted guided image filtering (AWGIF) is proposed to address the above issues. The AWGIF is applied to decompose an initial depth map which is estimated by the traditional SFF into a base layer and a detail layer. In order to preserve the edges accurately in the refined depth map, the guidance image is constructed from the multi-focus image sequence, and the coefficient of the AWGIF is utilized to suppress the noise while enhancing the fine depth details. Experiments on real and synthetic objects demonstrate the superiority of the proposed algorithm in terms of anti-noise, and the ability to preserve depth edges and fine structural details compared to existing methods.
\end{abstract}

%% Use if graphical abstract is present
%\begin{graphicalabstract}
%\includegraphics{}
%\end{graphicalabstract}

%% Research highlights
%\begin{highlights}
%\item 
%\item 
%\item 
%\end{highlights}

% Keywords
% Each keyword is seperated by \sep

\maketitle

% Main text
\section{Introduction}\label{Introduction}
Estimating depth information of a scene from 2D images is a challenging task in computer vision. Shape from focus (SFF) is a representative passive optical method utilizing 2D focus information as a cue to estimate the depth map of a scene. The advantages of SFF are simplicity and easy implementation, more accurate and less expensive than other depth estimation methods. In addition, it allows for a more compact electronic system. Therefore, it is widely used to acquire depth information of various scenarios, such as synthetic focus, autonomousc navigation\cite{greg2021dual,chao2020multi}, 3D parallax \cite{jeon2020ring}, microelectronics \cite{mahmood2012nonlinear}, industrial inspection \cite{alicona2021}, augmented or virtual reality (AR/VR) on mobile devices \cite{surh2017noise,suwajanakorn2015depth}, etc.

SFF techniques are usually performed by the following steps. The first step is to acquire multi-images in different focuses by changing the focus setting of the imaging device gradually \cite{nayar1994shape}. Since there is one-to-one correspondence between the depth of one point in the scene and the focus setting, the depth can be inferred from the best focused pixels  as long as the focus quality of each pixel is figured out. Therefore, in the second step, a criterion, commonly referred to as the focus measure (FM) operator, is applied to effectively calculate the focus quality of each pixel in the multi-focus images sequence, and the resulting volume is called as image focus volume \cite{mahmood2012nonlinear}. At the third step of SFF, the depth map can be estimated by finding the image number of the maximum focus measurement along the optical axis in the image focus volume \cite{ali2019image}. In the literature, the FM operators can be affected by noise level, contrast, the scene texture and other factors \cite{pertuz2013analysis}, which yields the focus volume to contain erroneous focus values, eventually leads to the depth map noisy and inaccurate. Therefore, a number of algorithms have been proposed to improve the focus volume to obtain the accurate depth map. This ranges from the simple linear filtering by averaging focus values in a small neighborhood window \cite{nayar1994shape,thelen2009improvements,jang2019optimal} to complex nonlinear methods \cite{mahmood2012nonlinear,ali2019image,mahmood2019cross,ma2020shape}. In addition, some algorithms attempt to describe SFF problem as depth reconstruction, which can directly obtain more accurate and reliable depth information from the noisy depth map. For instance, Hariharan and Rajagopalan \cite{hariharan2012shape} adopted smoothness constraints to improve the depth map. Tseng and Wang \cite{tseng2014shape} proposed a local learning scheme with a spatially consistent model to well restore the depth. Moeller et al. \cite{moeller2015variational} proposed to improve the depth map by employing an efficient nonconvex minimization scheme. Kumar and Sahay \cite{prashanth2017accurate} used the low rank prior as a regularizer to recover the depth map in the form of weighted kernel norm minimization. However, all these algorithms have one basic problem, i.e., they do not consider any additional information about the structure of the scene to improve the depth map. Once the initial depth estimation is erroneous, a very little improvement is expected.

\begin{figure*}[htp]
	\centering
	\vspace*{0.5cm}
	\includegraphics[height=6cm,width=16cm]{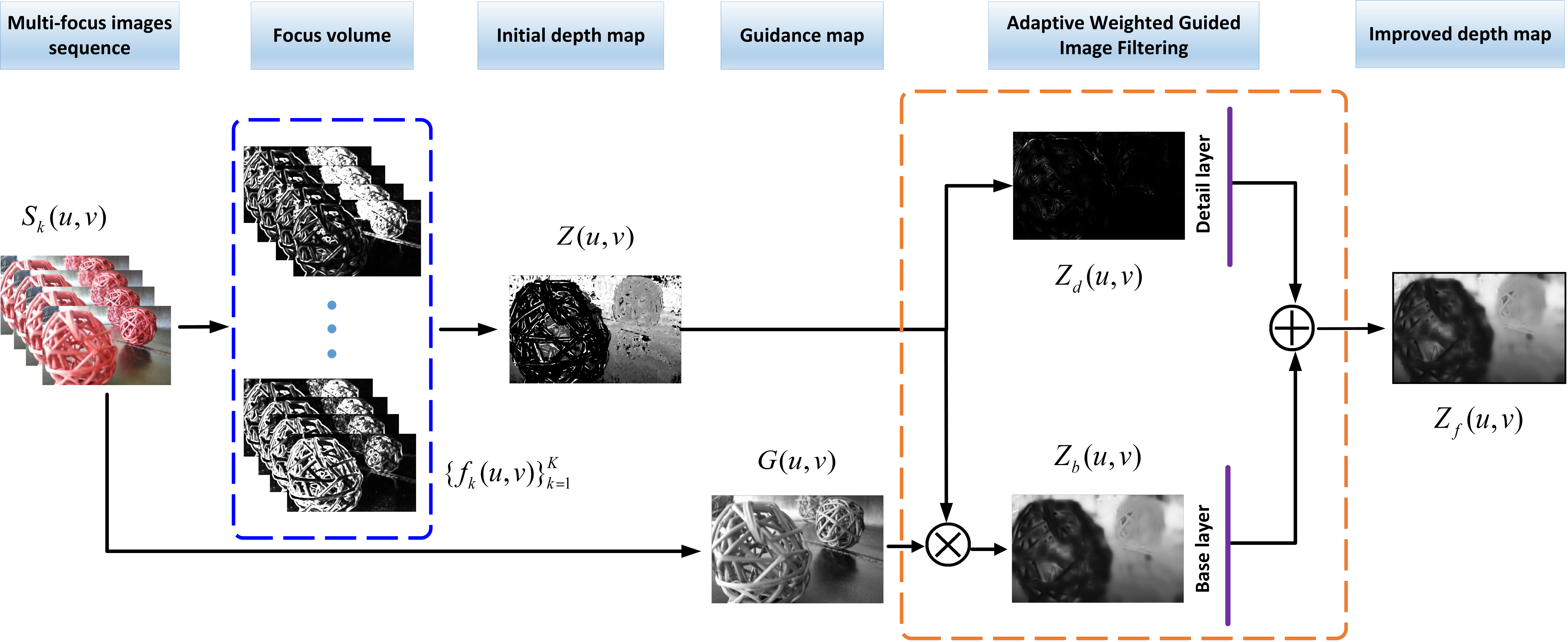}
	\caption{Framework of proposed AWGIF for depth enhancement in SFF}
	\label{fig:framework}
	\vspace*{-0.4cm}
\end{figure*}

Recently, guided filtering algorithms have been applied for SFF to improve depth maps \cite{ali2021guided,ali2021depth}. Guided filtering is a technique to obtain the target image with richer details via the structural information of the guided image. The technique extracts the guide image from the input image, and then guides the enhancement processing of the target image, so that the target image maintains the structure information of the input image. Therefore, the guided filtering algorithms are employed to the depth enhancement of SFF, which make the depth map more accurate by retaining the structural information of the input image sequences. Guided filtering is an edge - preserving smoothing filtering, which is widely used in detail enhancement \cite{kou2015content,kou2018itelligent}, Multi-scale exposure fusion \cite{kou2017multi},  etc. Guided filtering algorithms are usually divided into two categories. One is the global filtering algorithms, such as Least Squares (FWLS) filters \cite{kim2017fast}, Markov Random Field-based Filter (MRFF) \cite{diebel2005an}, Large Sparse Fusion (LSF) \cite{harrison2010image}. etc. The objective function of the global optimization filtering algorithm is mainly composed of the fidelity term and the regularization term, which filters the target image by calculating regularization weights of the guidance image. The quality of the enhanced image is often excellent by the global filters, but more computational time will be consumed. 

Another kind of guided filtering algorithms is served as a local filtering which uses the weight of pixel similarity in the local window of the reference image to filter the target image. Bilateral filter (BF) \cite{tomasil1998bilateral}, trilateral filter (TF) \cite{choudhury2003the} and Guided Image Filter (GIF) \cite{he2013guided} are the typical local filtering algorithms. Among them GIF is a local linear transformation in mathematics, with the advantages of simple and fast edge-preserving smoothing filtering, which is very suitable for SFF depth estimation, VR/AR and other handheld devices for online real-time operation. As pointed in \cite{ali2021guided}, it is important to adopt a simple edge-preserving smoothing filter for the refinement of the initial depth map. The GIF and its variants, such as weighted guided image filter (WGIF)\cite{li2015weighted}, weighted aggregation based GIF (WAGIF) \cite{chen2020weighted}, etc, are the most important edge-preserving smoothing filters. However, they do not preserve edge well. The edges of the improved depth map may suffer from the halo artifacts due to the fixed regularization parameters of the GIF \cite{he2013guided} in SFF. The WGIF \cite{li2015weighted} and WAGIF\cite{chen2020weighted} can address this problem by incorporating an edge-aware weighting, explicit first-order edge aware and weighted aggregation into the constraint term of the GIF, respectively. Nevertheless, the fixed amplification factor of these filters could amplify the inherent noise in the SFF \cite{ali2021guided}, which yields the improved depth map to be noisy and inaccurate.

Inspired by experimental results in \cite{ali2021guided} and the above challenging problems, a novel depth enhancement method based on adaptive WGIF (AWGIF) for the SFF is proposed in this paper. There are two steps in the estimation of the improved depth map. In the first step, an initial value of the depth map is estimated by the traditional SFF. In the second step, the improved depth map is recovered via the proposed AWGIF algorithm which is used to extract structural information from the input multi-focus image sequence as a guidance to compensate for incorrect depth estimation. The proposed adaptive WGIF can better suppress the halo artifacts and preserve clear edges by further improving the existing guidance filtering \cite{he2013guided,li2015weighted,chen2020weighted}. An adaptive amplification factor on top of the proposed AWGIF is designed and discussed for suppressing the noise and enhance depth details by using the coefficient of the proposed AWGIF intelligently. Experimental results indicate that the proposed AWGIF strategy has superiority on higher structural retention performance and better anti-noise capability than the existing related algorithm in \cite{he2013guided,li2015weighted,chen2020weighted,lu2018effictive}. In summary, the paper has the following contributions:

1) A novel AWGIF algorithm which outperforms the existing guided filters \cite{he2013guided,li2015weighted,chen2020weighted,lu2018effictive} to preserve accurate edges is proposed.

2) An adaptive adjustment algorithm for the amplification factor is proposed to effectively remove the noise while enhancing depth details.

3) A new strategy for depth enhancement in the SFF is proposed to obtain a high-quality depth map from 2D images. The proposed method can be applied to the emerging AR/VR on hand-held devices. 

The rest of this paper is organized as below. The proposed method of depth enhancement for SFF is described in great detail in Section \ref{Method}. The experimental results are compared and analyzed to verify the effectiveness of the proposed algorithm in Section \ref{Experiment}. conclusion remarks are provided in Section \ref{Conclusions}.

\begin{figure*}[htp]
	\centering
	%\vspace*{-0.5cm}
	\subfloat[input]{
		\begin{minipage}[b]{0.162\textwidth}
			\includegraphics[width=3cm,height=2cm]{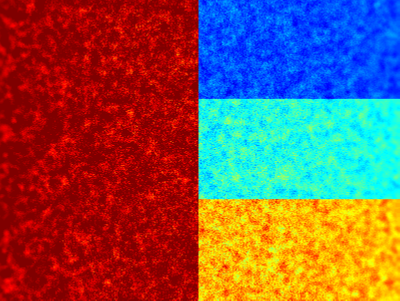}
		\end{minipage}
	}
	\subfloat[GIF]{
		\begin{minipage}[b]{0.162\textwidth}
			\includegraphics[width=3cm,height=2cm]{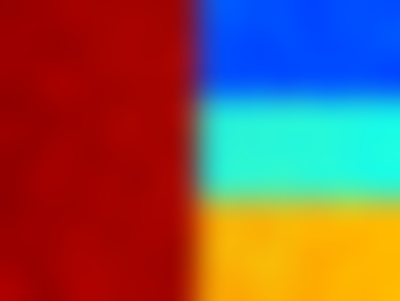}
		\end{minipage}
	}%\hspace{1in}
	\subfloat[WGIF]{
		\begin{minipage}[b]{0.162\textwidth}
			\includegraphics[width=3cm,height=2cm]{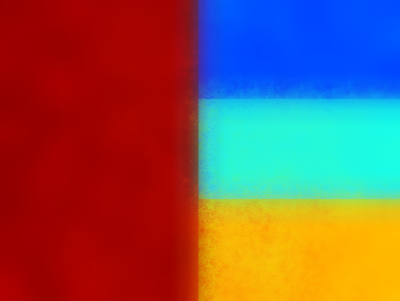}
		\end{minipage}
	}
	\subfloat[EGIF]{
		\begin{minipage}[b]{0.162\textwidth}
			\includegraphics[width=3cm,height=2cm]{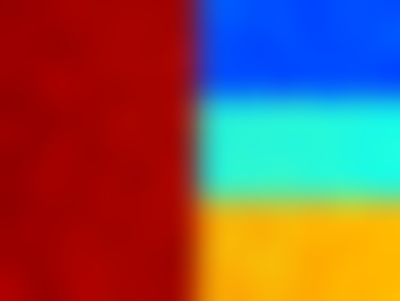}
		\end{minipage}
	}
	\subfloat[WAGIF]{
		\begin{minipage}[b]{0.162\textwidth}
			\includegraphics[width=3cm,height=2cm]{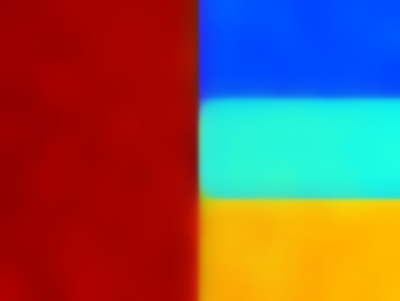}
		\end{minipage}
	}
	\subfloat[Ours]{
		\begin{minipage}[b]{0.162\textwidth}
			\includegraphics[width=3cm,height=2cm]{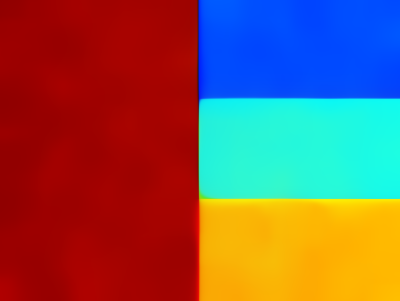}
		\end{minipage}
	}
		%\vspace{0.1cm}
	\caption{Comparison of different guided image filters in Case1 algorithm. (a) Input noise image. Edge-preserving smooth images by (b) GIF, (c) WGIF, (d) EGIF, (e) WAGIF, and (f) OursCase1. $\zeta = 15$, $\lambda_0 = 10^3$ in all the five filters.}
	\label{fig:case1_comparison}
	\vspace*{-0.5cm}
\end{figure*}

\begin{figure*}[htp]
	\centering
	%\vspace*{-0.5cm}
	\subfloat[input]{
		\begin{minipage}[b]{0.117\textwidth}
			\includegraphics[width=2.21cm,height=3.5cm]{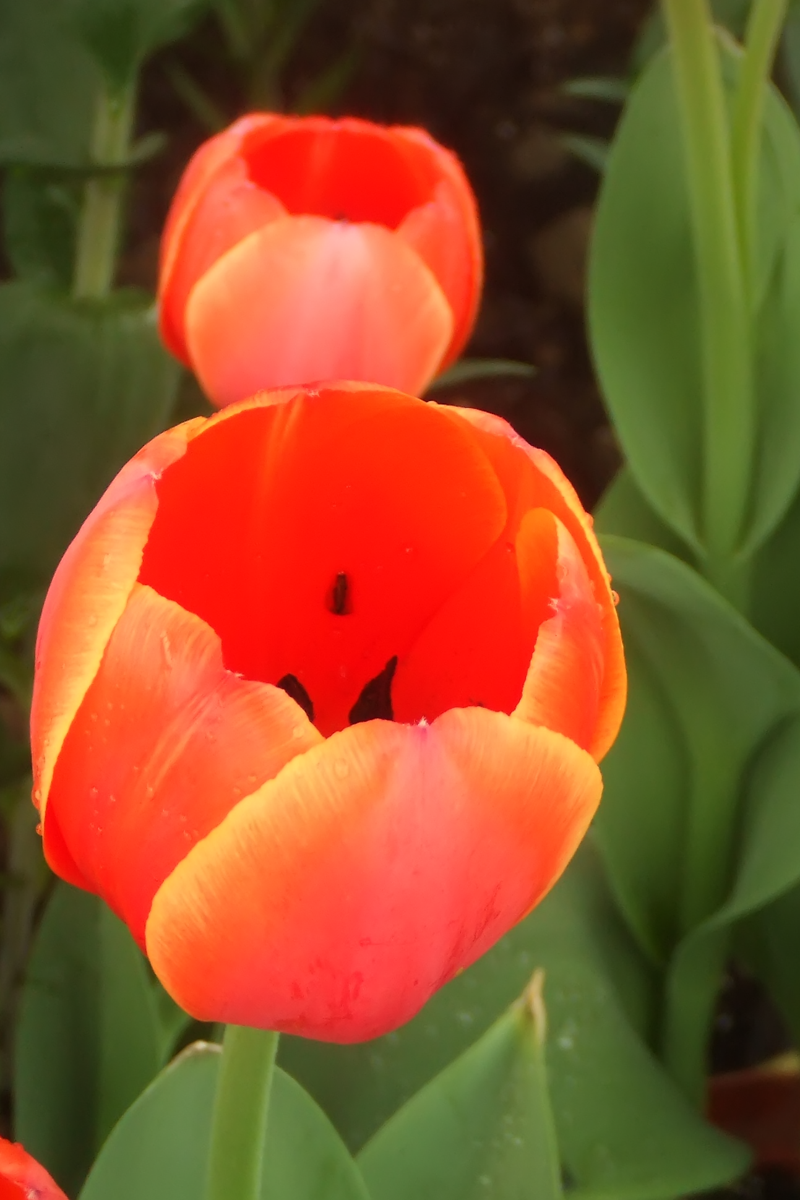}
		\end{minipage}
	}
	\subfloat[GIF]{
		\begin{minipage}[b]{0.117\textwidth}
			\includegraphics[width=2.21cm,height=3.5cm]{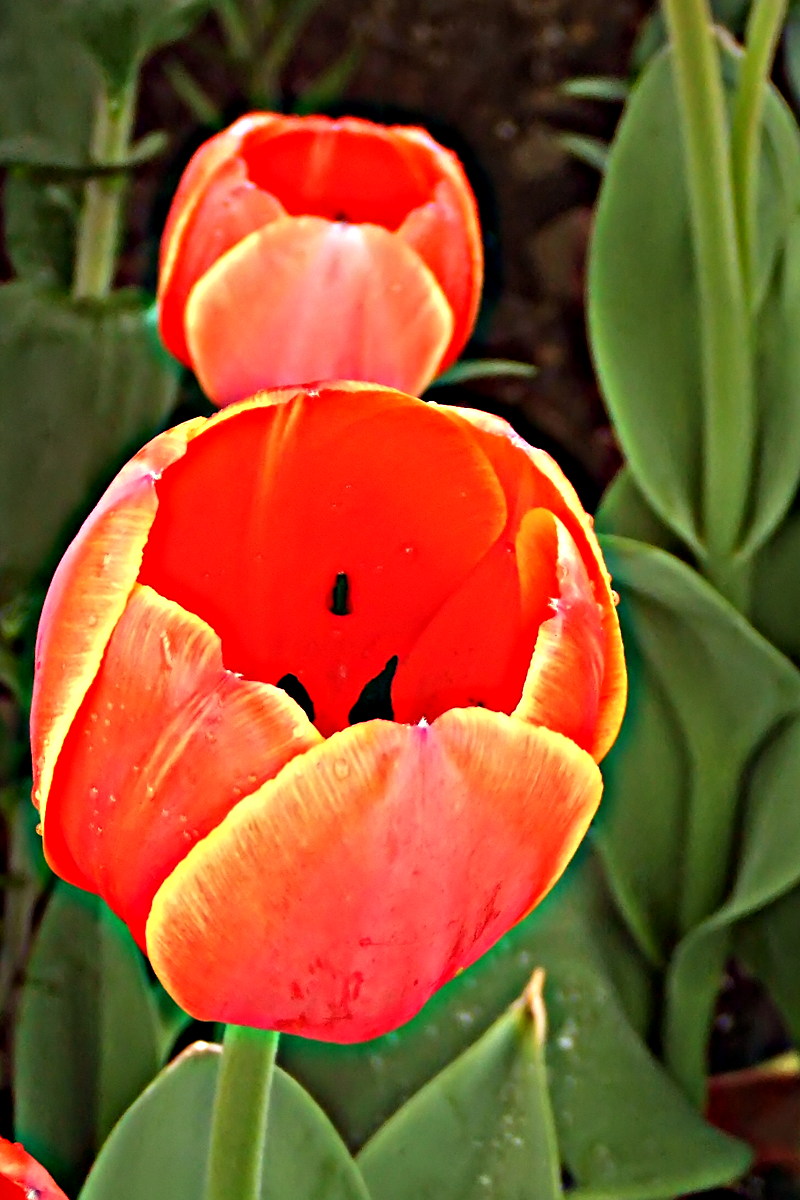}
		\end{minipage}
	}
	\subfloat[WGIF]{
		\begin{minipage}[b]{0.117\textwidth}
			\includegraphics[width=2.21cm,height=3.5cm]{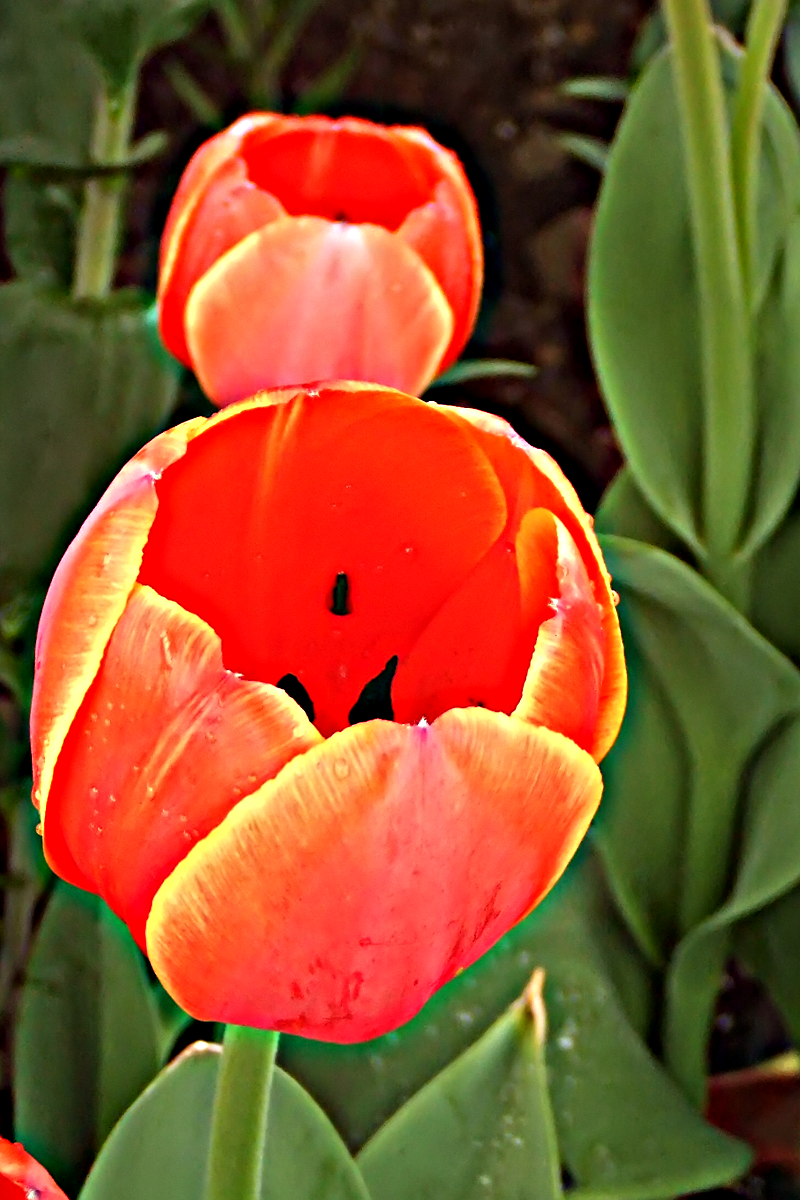}
		\end{minipage}
	}
	\subfloat[EGIF]{
		\begin{minipage}[b]{0.117\textwidth}
			\includegraphics[width=2.21cm,height=3.5cm]{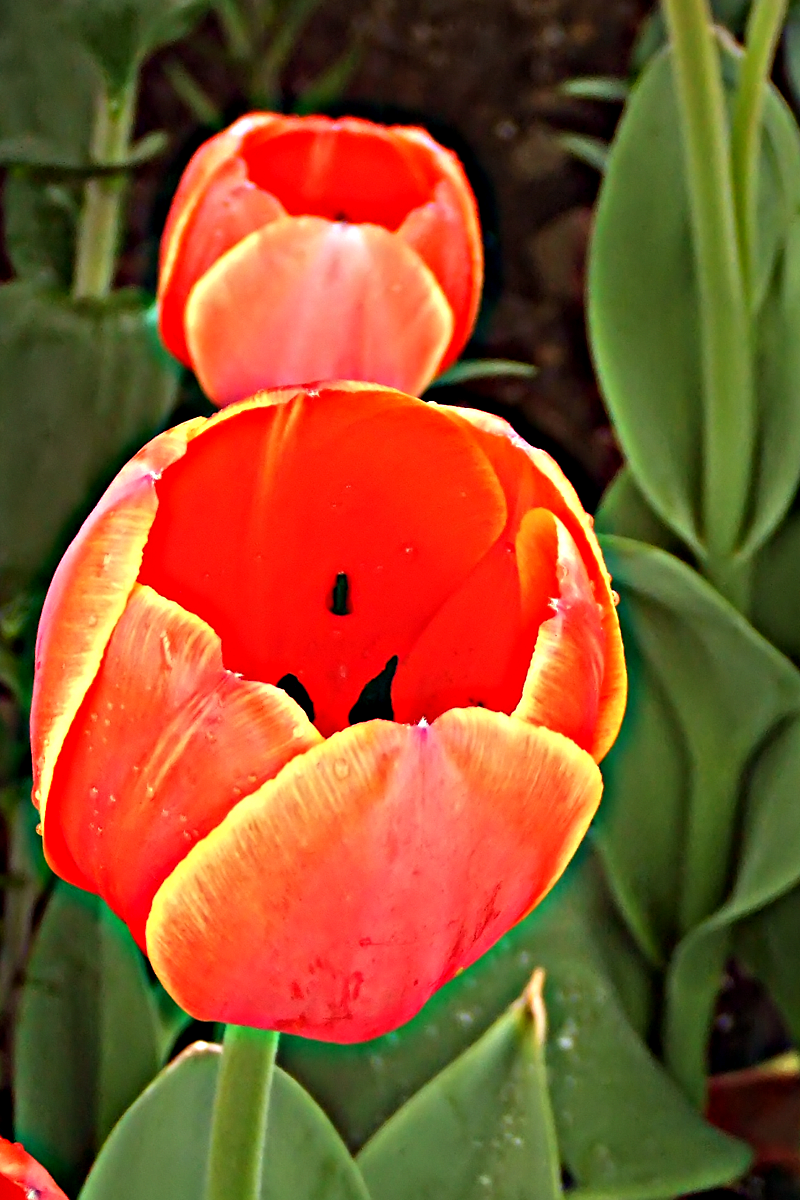}
		\end{minipage}
	}%\hspace{1in}
	\subfloat[WAGIF]{
		\begin{minipage}[b]{0.117\textwidth}
			\includegraphics[width=2.21cm,height=3.5cm]{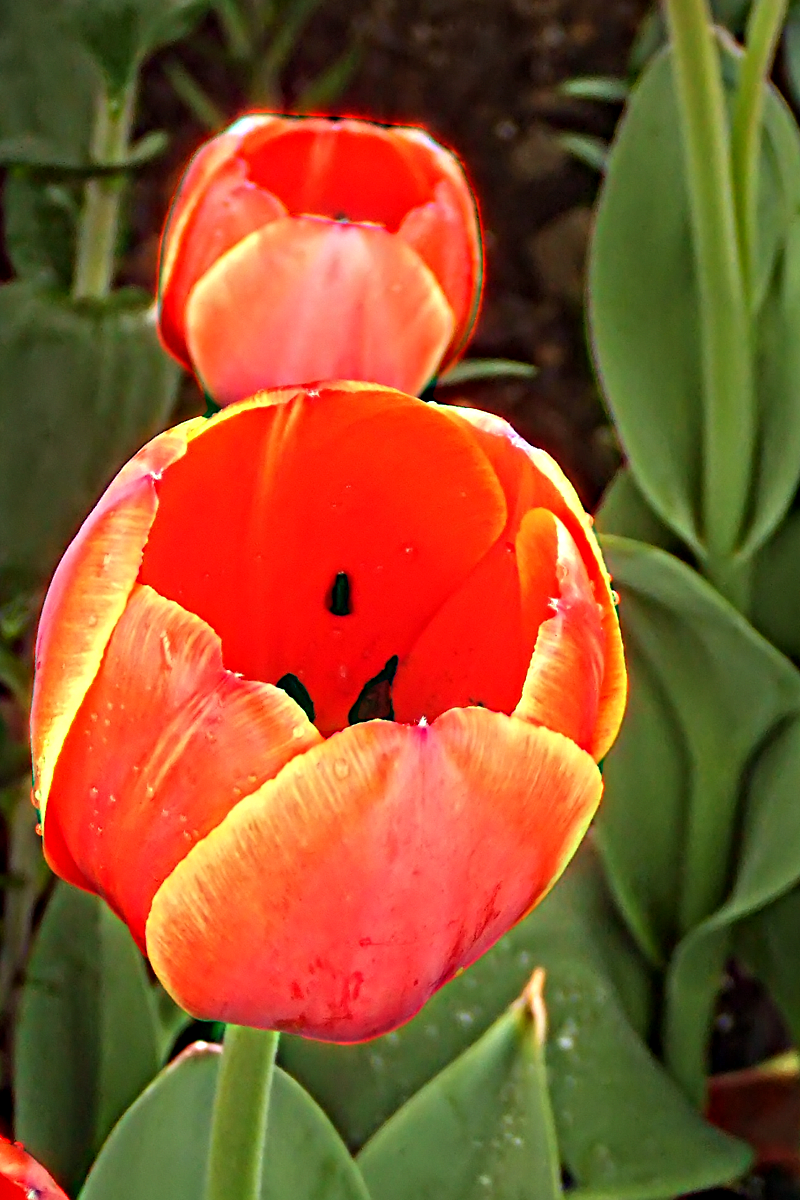}
		\end{minipage}
	}
	\subfloat[OursCase2]{
		\begin{minipage}[b]{0.117\textwidth}
			\includegraphics[width=2.21cm,height=3.5cm]{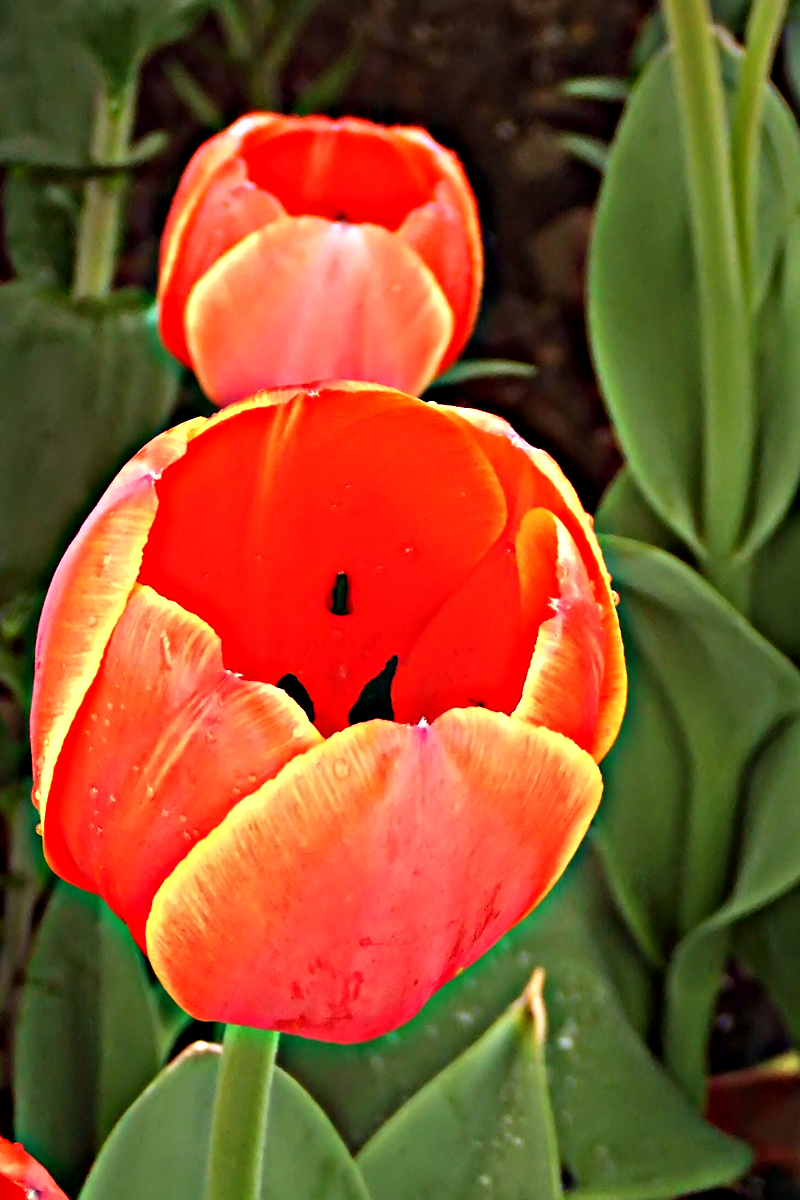}
		\end{minipage}
	}
	\subfloat[OursCase3]{
		\begin{minipage}[b]{0.117\textwidth}
			\includegraphics[width=2.21cm,height=3.5cm]{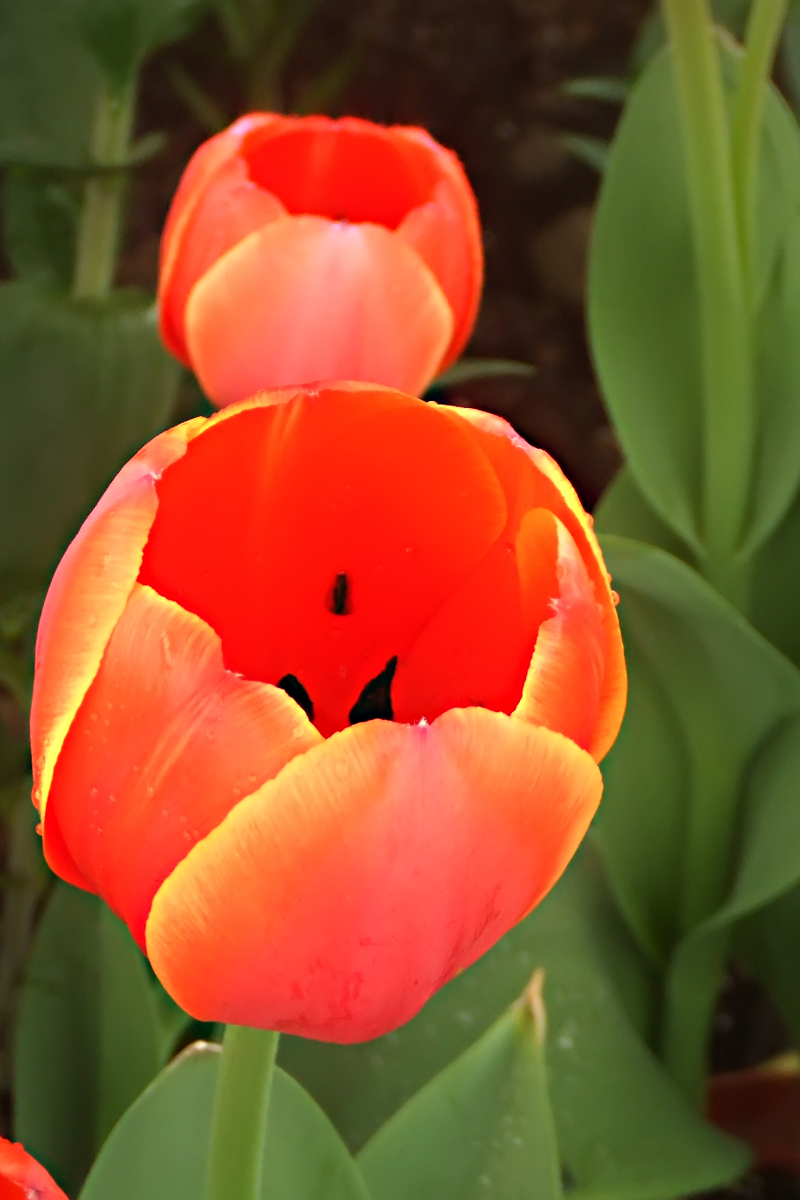}
		\end{minipage}
	}
	\subfloat[OursCase4]{
		\begin{minipage}[b]{0.11\textwidth}
			\includegraphics[width=2.21cm,height=3.5cm]{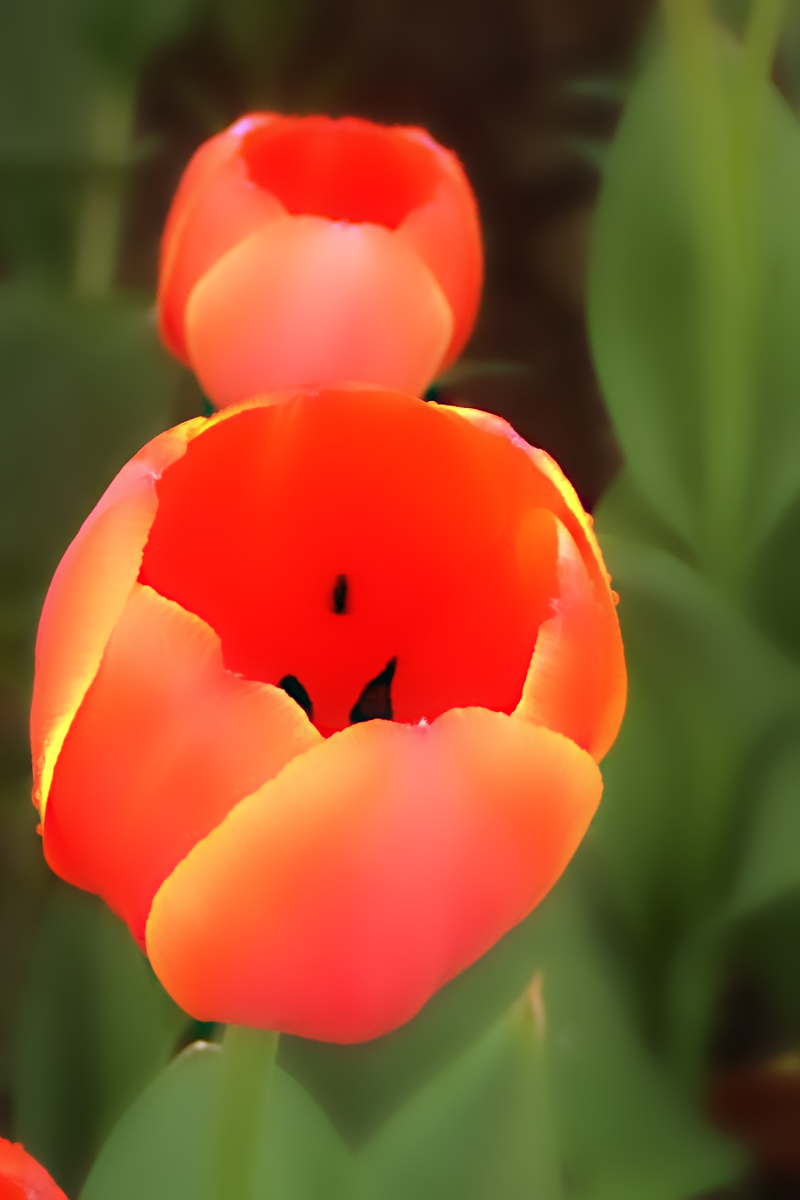}
		\end{minipage}
	}
		%\vspace{0.1cm}
	\caption{Comparison of different guided image filters in Case2 algorithm and the results of our proposed algorithm under different cases. (a) Input image. Conventional detail enhancement images by (b) GIF, (c) WGIF, (d) EGIF, (e) WAGIF, and (f) OursCase2. $\alpha = 3$, $\beta = 0$, $\zeta = 15$, $\lambda_0 = 10^3$ in all above five filters. Selective detail enhancement image by (g) OursCase3 ($\alpha = 1$, $\beta = 255^2/48$, $\zeta = 15$, $\lambda_0 = 10^3$). Hybrid smoothing and detail enhancement image by (h) OursCase4 ($\alpha = 0$, $\beta = 255^2/20$, $\zeta = 15$, $\lambda_0 = 10^3$).}
	\label{fig:case2_comparison}
	\vspace*{-0.4cm}
\end{figure*}

\section{The Proposed Method}\label{Method}
In this section, the proposed new method is described for the enhancement of depth map in the SFF technique. The framework is depicted in Fig. \ref{fig:framework}. Firstly, the initial depth map is estimated via the focus volume which is obtained by utilizing a FM operator. Then, the proposed AWGIF algorithm is elaborated and four special cases of the adaptive adjustment algorithm are designed and discussed to verify the effective of the proposed algorithm. Finally, the depth map is reconstructed by applying the AWGIF. These steps are described in detail as follow.
\subsection{Estimation of Initial Depth Map}\label{Initial}
Let the captured multi-focus images sequence be represented as 3D data ${S_k}(u,v)$ which is composed of $K$ grayscale images with the size of $U \times V$, where $k \in \{ 1,...,K\}$, $u \in \{ 1,...,U\}$, $v \in \{ 1,...,V\}$. The focus quality of each pixel $p = (u,v)$ in the images sequence is measured by employing any suitable FM operator. In the paper, we have applied the well-known statistics-based FM operator gray level variance (GLV) [9] which computes the variance of image gray levels within a small local window. Thus the initial focus volume ${f_k}(p)$ is obtained which is described as
\begin{flalign}
\label{eq:SFF_focusvolume}
&{f_k}(p) = {S_k}(p) \otimes FM ,&
\end{flalign}
where $\otimes$ is the 2D convolution operator. In order to reduce the influence of noise in the FM, the linear filtering of focus values in $5 \times 5$ neighborhood was aggregated to obtain the enhanced focus volume ${\tilde f_k}(p)$. From the enhanced focus volume, the initial depth map $Z(p)$ is estimated by finding the image number corresponding to the maximum focus values along the k-direction, as given by
\begin{flalign}
\label{eq:SFF_initialdepth}
&Z(p) = \arg \max ({\tilde f_k}(p)) .&
\end{flalign}

Due to the limited capability of the FM operator, the initial depth map might be erroneous at certain points \cite{pertuz2013analysis}. In order to compensate for these errors and make the depth map have better performance in denoising, preserving edges and retaining details, a novel strategy based on the adaptive weighted guided image filtering (AWGIF) is proposed to enhance the depth map.

\subsection{Adaptive Weighted Guided Image Filtering}\label{Adaptive}
In this subsection, an AWGIF is introduced to improve the WGIF such that depth edges are preserved better, noise is reduced more, and structural details is retained richer.

Let $Z$ be an input image to be processing and $G$ be a guidance image. $\Omega_{\zeta}(p')$ is  a square window centered at the pixel $p'$
of a radius $\zeta$. Same as  the GIF, $Z(p)$ is assumed to be a linear transform of the guidance image $G(p)$ in the window $\Omega_{\zeta}(p')$:
\begin{flalign}
\label{linearmodel}
&Z(p)=a_{p'}G(p)+b_{p'}, \forall p\in
\Omega_{\zeta}(p'),&
\end{flalign}
where $a_{p'}$ and $b_{p'}$ are two constants in the
window $\Omega_{\zeta}(p')$.

The optimal values of $a_{p'}$ and $b_{p'}$ are derived by minimizing a quadratic cost function $E(a_{p'},b_{p'})$ which is defined as \cite{li2015weighted}
\begin{flalign}
\label{novel_costfunction}
&\sum_{p\in
	\Omega_{\zeta}(p')}[\Gamma^{G}_{p'}(a_{p'}G(p)+b_{p'}-Z(p))^2+\lambda  a_{p'}^2],&
\end{flalign}

\begin{figure*}[htp]
	\centering
	%\vspace*{-0.5cm}
	\subfloat[SyntheticClean]{
		\begin{minipage}[b]{0.24\textwidth}
			\includegraphics[width=4.0cm,height=2.8cm]{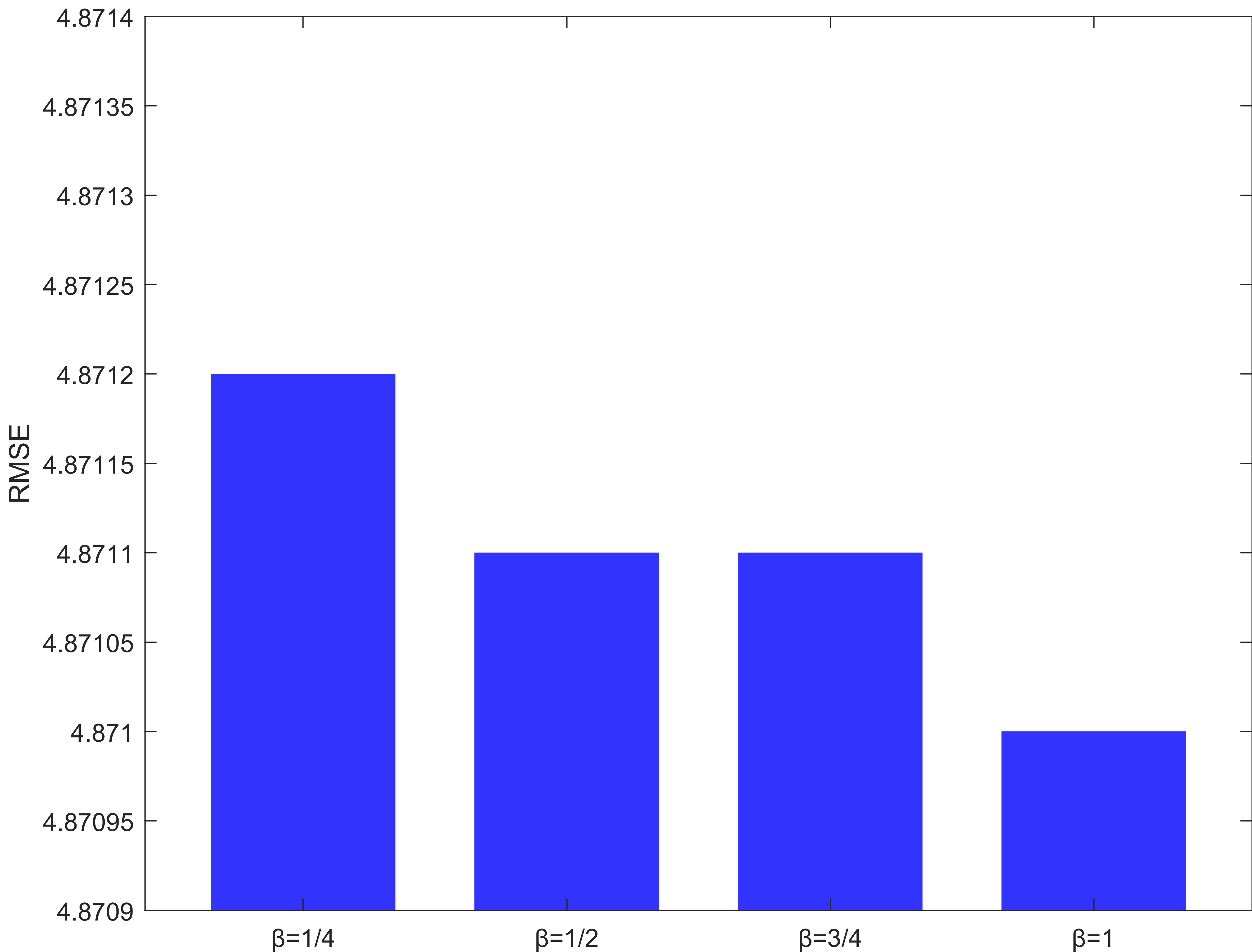}
		\end{minipage}
	}
	\subfloat[SyntheticNoise]{
		\begin{minipage}[b]{0.24\textwidth}
			\includegraphics[width=4.0cm,height=2.8cm]{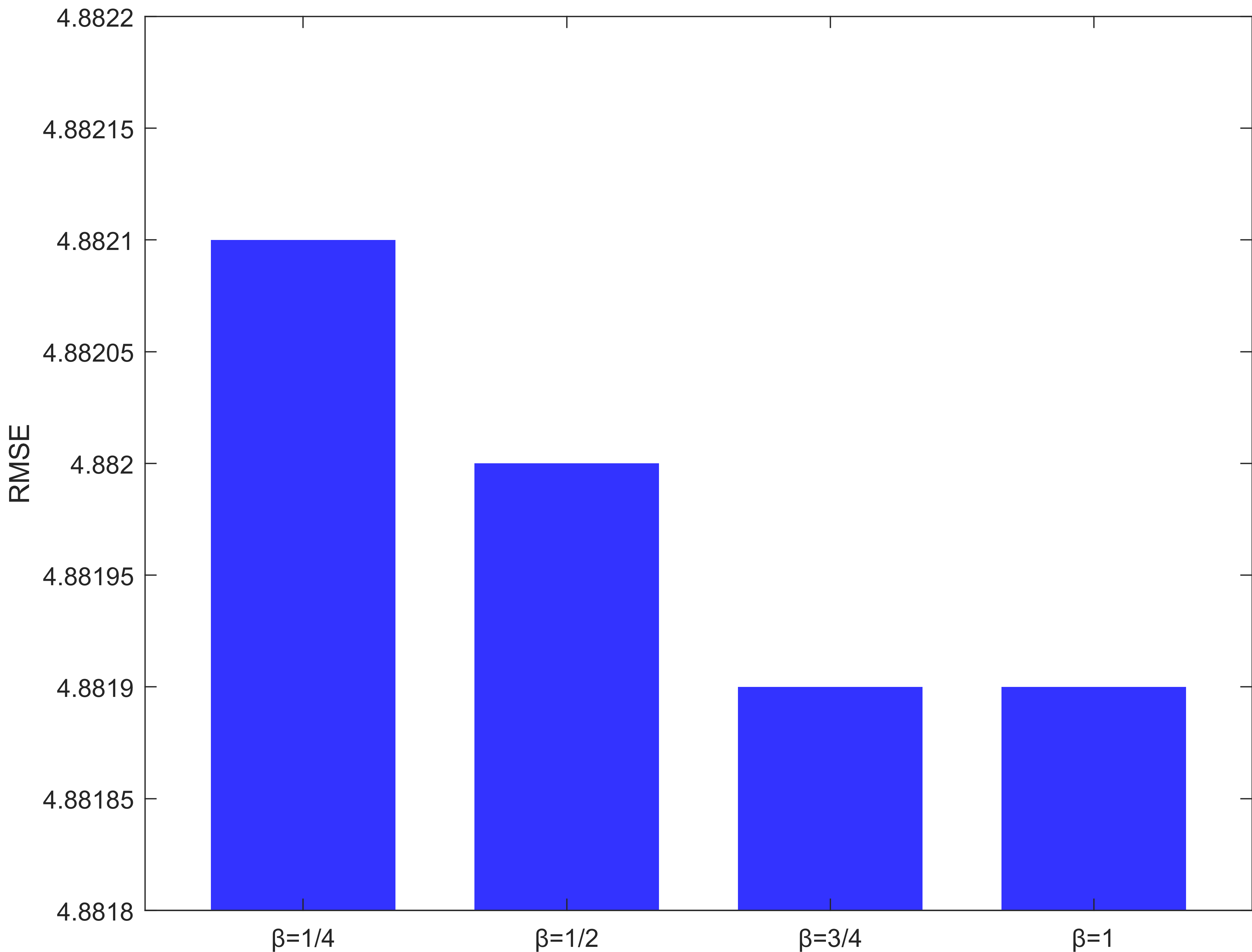}
		\end{minipage}
	}%\hspace{1in}
	\subfloat[RealClean]{
		\begin{minipage}[b]{0.24\textwidth}
			\includegraphics[width=4.0cm,height=2.8cm]{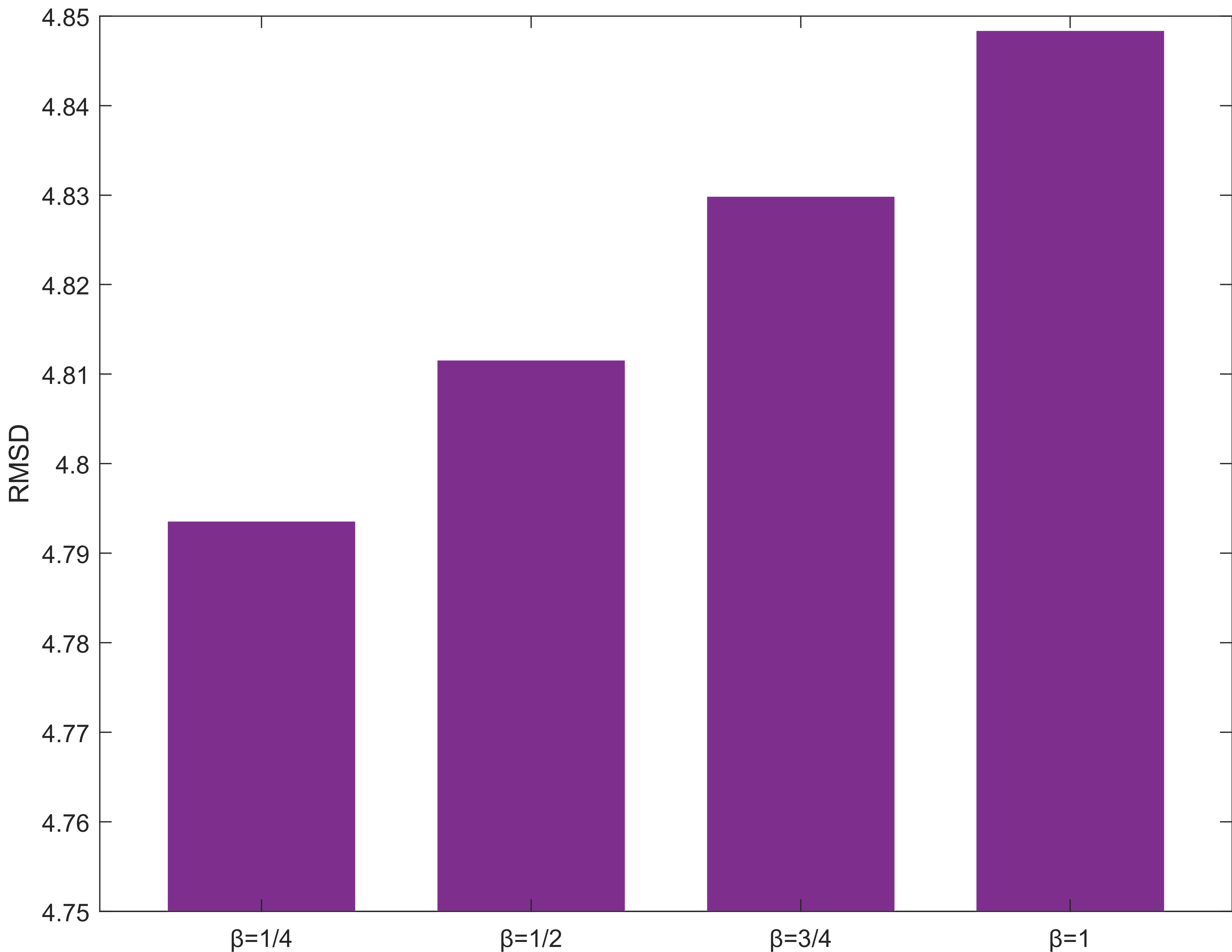}
		\end{minipage}
	}
	\subfloat[RealNoise]{
		\begin{minipage}[b]{0.24\textwidth}
			\includegraphics[width=4.0cm,height=2.8cm]{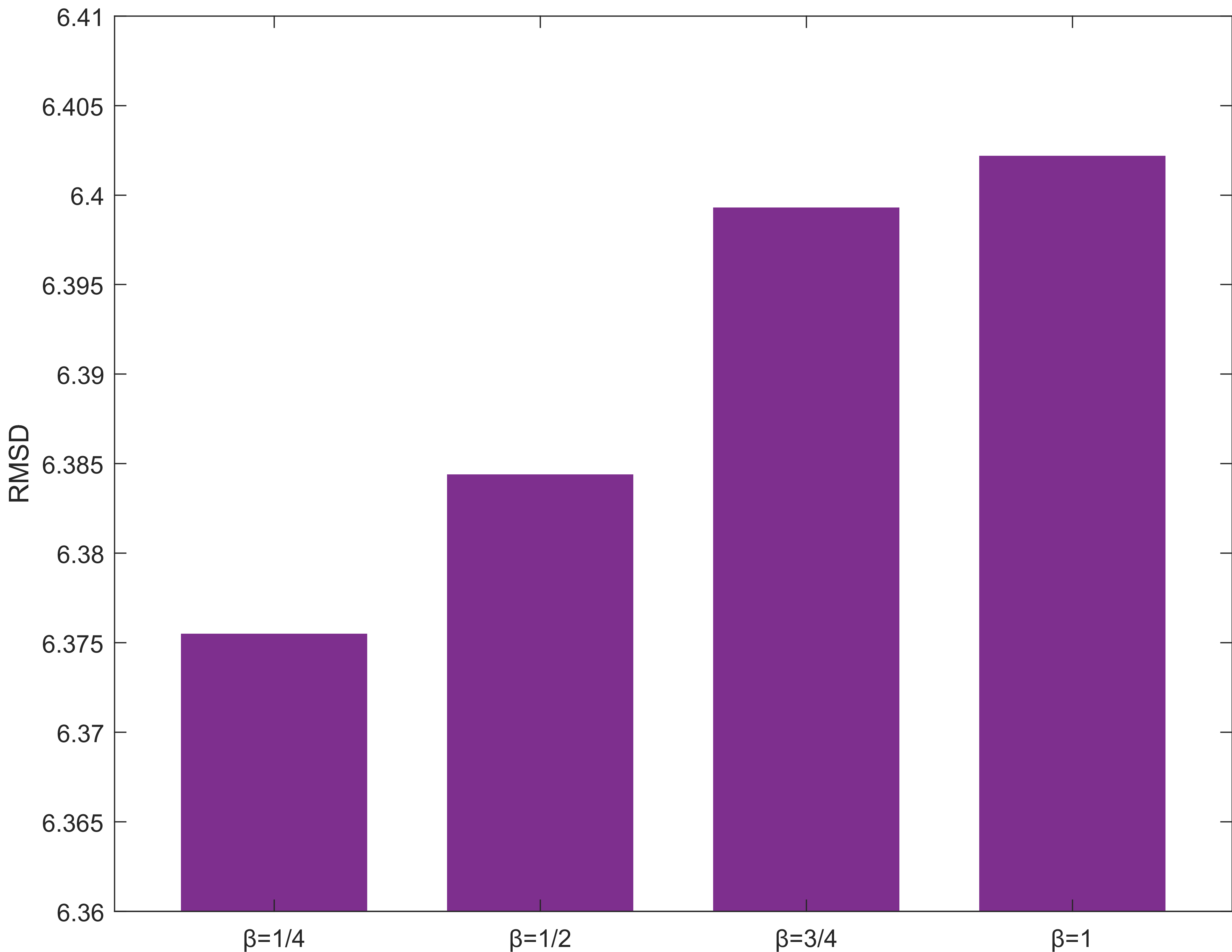}
		\end{minipage}
	}
		%\vspace{0.1cm}
	\caption{Performance of the RMSE for synthetic images and the RMSD for real images with the different value of $\beta$.}
	\label{fig:beta_selection}
	\vspace*{-0.4cm}
\end{figure*}

Similar to the WGIF, the edge-aware weighting $\Gamma^{G}_{p'}$ is defined  by using local variances of $3\times 3$ windows of all pixels from the guidance image $G$ as follows:
\begin{flalign}
\label{edgeaware_weighing}
&\Gamma^{G}_{p'}=\frac{1}{M}\sum_{p=1}^{M}\frac{\sigma^2_{G,1}(p')+\varepsilon}{\sigma^2_{G,1}(p)+\varepsilon},&
\end{flalign}
where $M$ is the total number of pixels in the image $G$.  $\varepsilon$ is a small constant and it is 1 for an 8-bit image in the range [0, 255].
$\lambda$ is adaptive to the image $G$ and it is defined as $\lambda_0(\frac{1}{M}\sum_{p=1}^{M}\sigma^2_{G,\zeta}(p))^{0.5}$, where $\lambda_0$ is a regularization parameter which penalizes a large $a_{p'}$. This is different from the $\lambda$'s in  \cite{he2013guided,li2015weighted,chen2020weighted}.

The optimal $a_{p'}$ and $b_{p'}$ are computed as
\begin{flalign}
\label{optimal_ap&bp}
&\begin{aligned}
a_{p'}=\frac{\Gamma^{G}_{p'}\mbox{cov}_{Z,G,\zeta}(p')}{\Gamma^{G}_{p'}\sigma^2_{G, \zeta}(p')+\lambda}\; ;\;
b_{p'}=\mu_{Z,\zeta}(p')-a_{p'}\mu_{G,\zeta}(p'),
\end{aligned}&
\end{flalign}
where $\mbox{cov}_{Z,G,\zeta}(p')$ is
\begin{flalign}
\label{cov}
&\begin{aligned}
\mbox{cov}_{Z,G,\zeta}(p')=\mu_{G\odot Z, \zeta}(p')-\mu_{G,\zeta}(p')\mu_{Z,\zeta}(p'),
\end{aligned}&
\end{flalign}
and $\mu_{G, \zeta}(p')$, $\mu_{Z, \zeta}(p')$, and  $\mu_{G\odot Z, \zeta}(p')$ are the mean values of $G$, $Z$ and $G\odot Z$ in the window $\Omega_{\zeta}(p')$,  respectively. The operation $\odot$ is the Hadamard product.

When the images $Z$ and $G$ are the same, the optimal values of $a_{p'}$ and $b_{p'}$ are computed as
\begin{flalign}
\label{G_Z_same}
&\begin{aligned}
a_{p'}=\frac{\Gamma^{Z}_{p'}\sigma^2_{Z, \zeta}(p')}{\Gamma^{Z}_{p'}\sigma^2_{Z, \zeta}(p')+\lambda}\; ;\;
b_{p'}=(1-a_{p'})\mu_{Z,\zeta}(p').
\end{aligned}&
\end{flalign}

It can be easily shown that $a_{p'}$ approaches 1 closer if $p'$ is at a sharp edge and approaches 0 closer if $p'$ is in a flat area. Therefore, the proposed AWGIF preserves sharp edges better than the GIF, while suppressing the halo artifacts.

Instead of using the averaging method in \cite{he2013guided,li2015weighted}, a weighted averaging method is adopted to compute $\bar{a}_p$ and $\bar{b}_p$ as
\begin{flalign}
\label{average_ap&bp}
&\begin{aligned}
\{\bar{a}_p, \bar{b}_p\}=\frac{1}{W^{sum}_p}\sum_{p'\in \Omega_{\zeta}(p)}W_{p'}\{a_{p'}, b_{p'}\},
\end{aligned}&
\end{flalign}
where $W_p^{sum}$ is $=\sum_{p'\in \Omega_{\zeta}(p)}W_{p'}$, and $W_{p'}$ is given as \cite{chen2020weighted}
\begin{flalign}
\label{weighting_chen}
&\begin{aligned}
W_{p'}=\exp^{-\frac{1}{|\Omega_{\zeta}(p')|}{\displaystyle \sum_{p\in \Omega_{\zeta}(p')}}\frac{(a_{p'}G(p)+b_{p'}-Z(p))^2}{\eta}}+0.001,
\end{aligned}&
\end{flalign}
$\eta$ is a small positive constant and its value is $1/200^2$ if not specified rather than 1/200 in \cite{chen2020weighted}, and $|\Omega_{\zeta}(p')|$ is the cardinality of the set $\Omega_{\zeta}(p')$.

The output image is represented as $(\bar{a}_pG(p)+\bar{b}_p)$. Same as the GIF \cite{he2013guided}, WGIF \cite{li2015weighted} and WAGIF \cite{chen2020weighted}, the complexity of the proposed AWGIF is $O(M)$ for an image with $M$ pixels. The proposed AWGIF can be applied to decompose the image $Z$ into a base layer $Z_b$ and a detail layer $Z_d$. An output depth map $Z_f$ is generated by adding an adjusted detail layer to the base layer as
\begin{flalign}
\label{detailenhancement}
&Z_f=Z_b+(\alpha+\beta\bar{a})\odot Z_d,&
\end{flalign}
where $(\alpha+\beta\bar{a})$ is the amplification factor for the detail layers,  which is adaptive to the content of the image Z. $\alpha$ and $\beta$ are two constants. Some special cases of the equation (\ref{detailenhancement}) are discussed as follows:

{\it Case 1}: If both $\alpha$ and $\beta$ are zeros, the equation (\ref{detailenhancement}) is an edge-preserving smoothing algorithm. $\zeta = 15$, $\lambda_0 = 10^3$ are set for all five filters  \cite{he2013guided,li2015weighted,chen2020weighted,lu2018effictive} and the results are show in Fig.\ref{fig:case1_comparison}. All five filters reduce noise well, but the proposed AWGIF preserves edges best while smoothing the noise image among all the GIF and its variants.

{\it Case 2}: If $\alpha$ is larger than 1 and $\beta$ is zero, the equation (\ref{detailenhancement}) becomes a conventional detail enhancement algorithm. Set as $\alpha = 3$, $\beta = 0$, $\zeta = 15$, $\lambda_0 = 10^3$ for all five filters \cite{he2013guided,li2015weighted,chen2020weighted,lu2018effictive}. The results of image enhancement in Fig.\ref{fig:case2_comparison}(b-f) show that the proposed AWGIF algorithm maintains better details than other algorithms and reduces halo artifacts appropriately. 

{\it Case 3}: If $\alpha$ is  1 and $\beta$ is positive, the equation (\ref{detailenhancement}) is a selective detail enhancement algorithm. The parameters are set as $\alpha = 1$, $\beta = 255^2/48$, $\zeta = 15$, $\lambda_0 = 10^3$, and the enhanced image is described in Fig.\ref{fig:case2_comparison}(g). The edges of the flower are well preserved, the details of the flower center are enhanced, and the halo artifacts are suppressed.

{\it Case 4}: If $\alpha$ is  0 and $\beta$ is positive, the equation (\ref{detailenhancement}) becomes a hybrid smoothing and detail enhancement algorithm. Set the parameters as $\alpha = 0$, $\beta = 255^2/20$, $\zeta = 15$, $\lambda_0 = 10^3$, and the processed image is demonstrated in Fig.\ref{fig:case2_comparison}(h). The details of flower edges and centers are well enhanced while the background is well smoothed, indicating that the algorithm can better smooth useless information such as noise while enhancing details.

From the analysis of the above four special cases, it can be seen that the proposed AWGIF algorithm has outstanding performance in edge preserving and noise smoothing, and is suitable for all kinds of situations requiring structure preserving or smoothing filtering. Especially in case 4, the adaptive adjustment algorithm smooths the noise and reduces the halo artifacts while maintaining the structural details, thus, has better image enhancement capability than the existing algorithms \cite{he2013guided,li2015weighted,lu2018effictive,chen2020weighted}. Due to the inherent properties of SFF, namely, the depth image has a lot of noise, the enhancement of the depth image needs to maintain the scene structure and smooth the noise. Therefore, the proposed AWGIF can solve the problem of depth enhancement well.

\begin{figure*}[htp]
	\centering
	%\vspace*{-0.5cm}
	\subfloat[$\beta = 1/4$]{
		\begin{minipage}[b]{0.24\textwidth}
			\includegraphics[width=4.0cm,height=2.8cm]{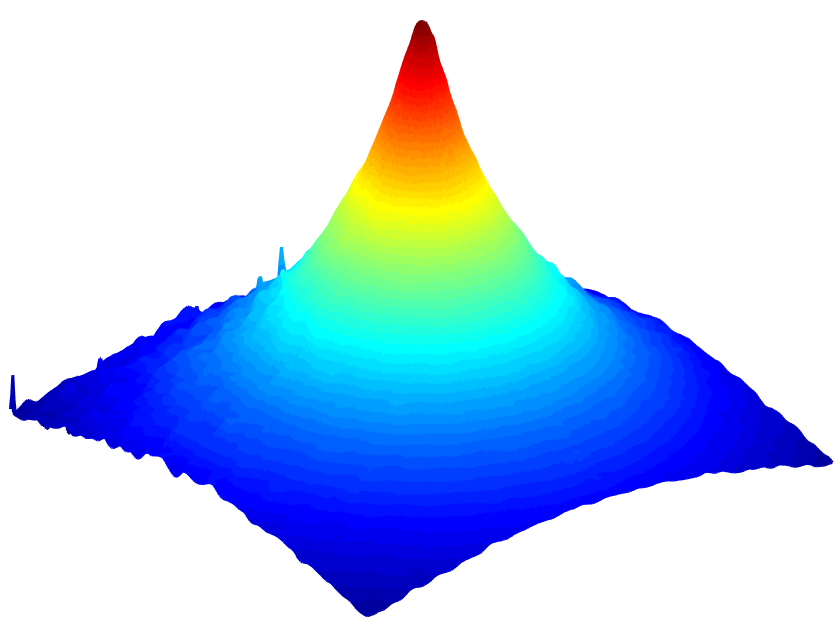}
		\end{minipage}
	}
	\subfloat[$\beta = 1/2$]{
		\begin{minipage}[b]{0.24\textwidth}
			\includegraphics[width=4.0cm,height=2.8cm]{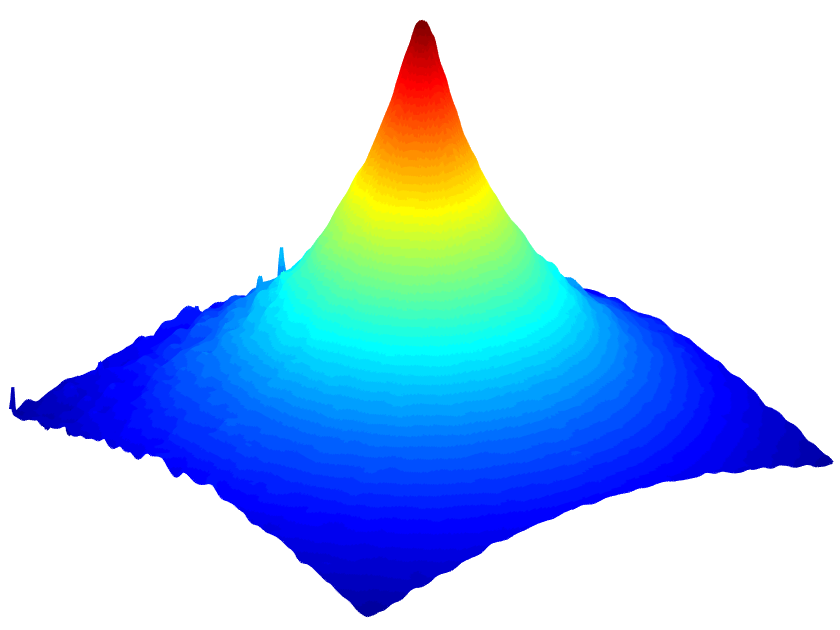}
		\end{minipage}
	}
	\subfloat[$\beta = 3/4$]{
		\begin{minipage}[b]{0.24\textwidth}
			\includegraphics[width=4.0cm,height=2.8cm]{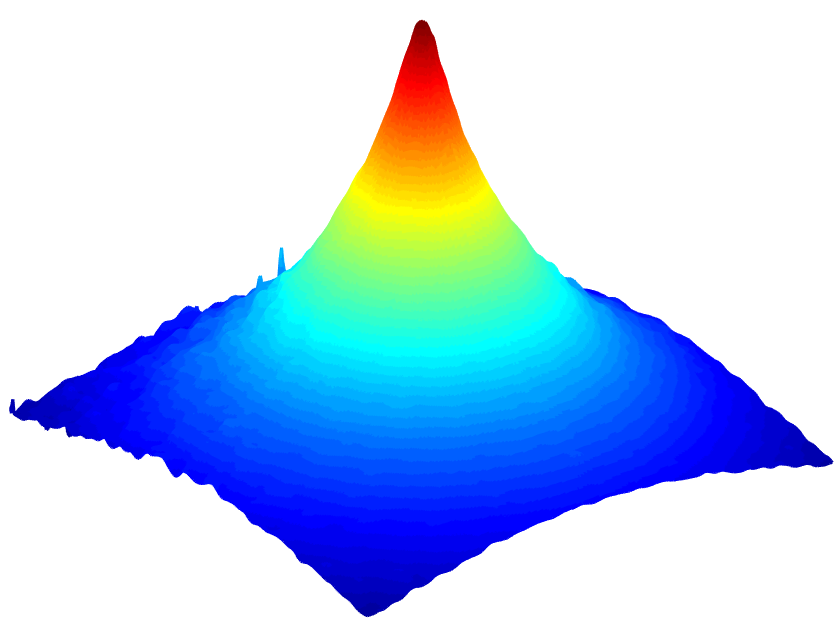}
		\end{minipage}
	}
	\subfloat[$\beta = 1$]{
		\begin{minipage}[b]{0.24\textwidth}
			\includegraphics[width=4.0cm,height=2.8cm]{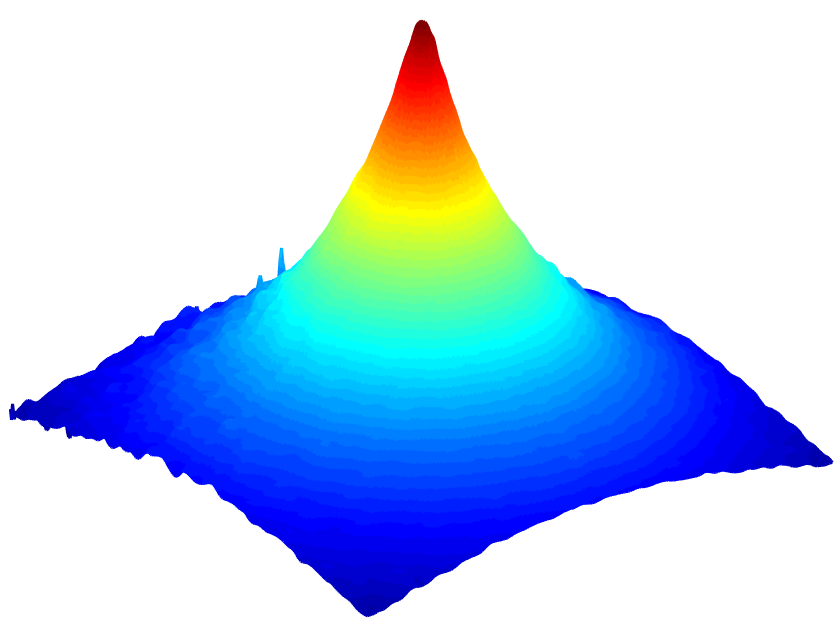}
		\end{minipage}
	}\hspace{1in}
	\subfloat[$\beta = 5/4$]{
		\begin{minipage}[b]{0.24\textwidth}
			\includegraphics[width=4.0cm,height=2.8cm]{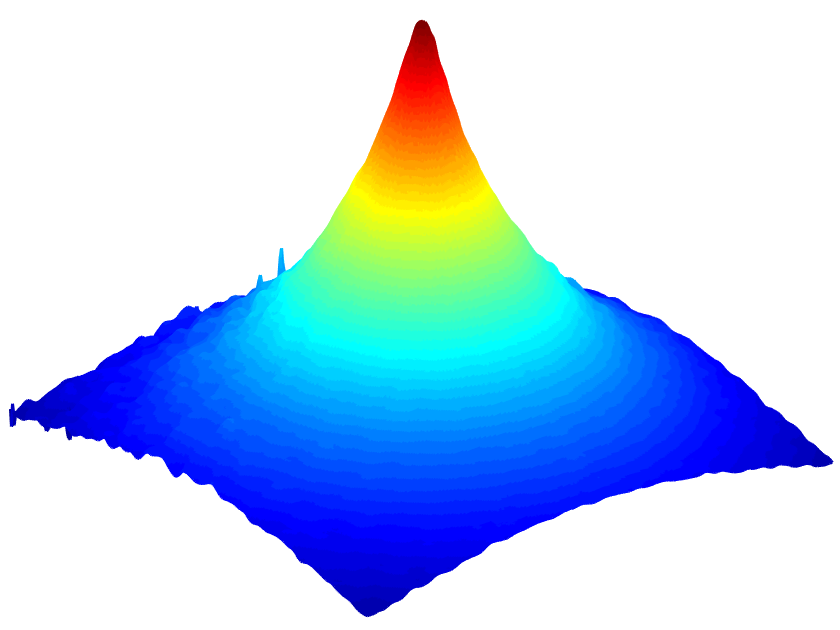}
		\end{minipage}
	}
	\subfloat[$\beta = 3/2$]{
		\begin{minipage}[b]{0.24\textwidth}
			\includegraphics[width=4.0cm,height=2.8cm]{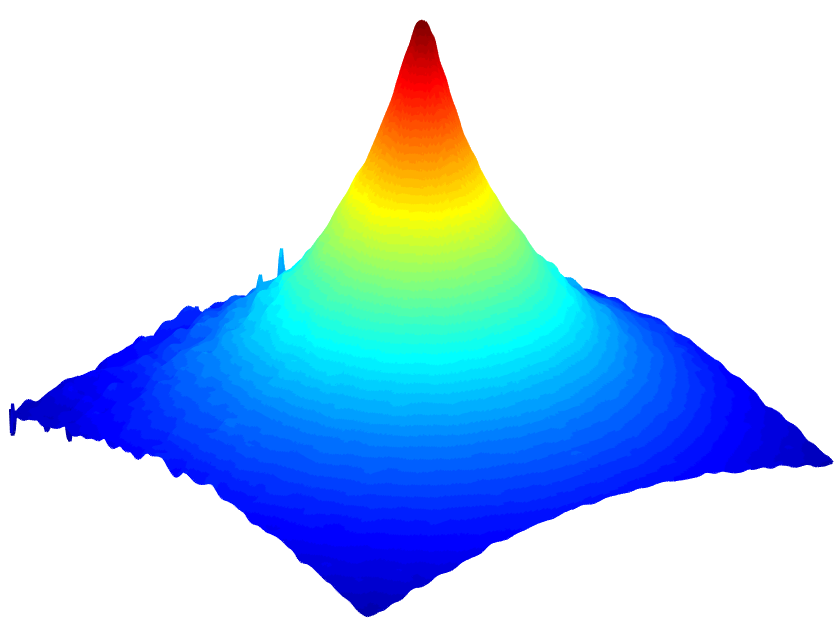}
		\end{minipage}
	}
	\subfloat[$\beta = 7/4$]{
		\begin{minipage}[b]{0.24\textwidth}
			\includegraphics[width=4.0cm,height=2.8cm]{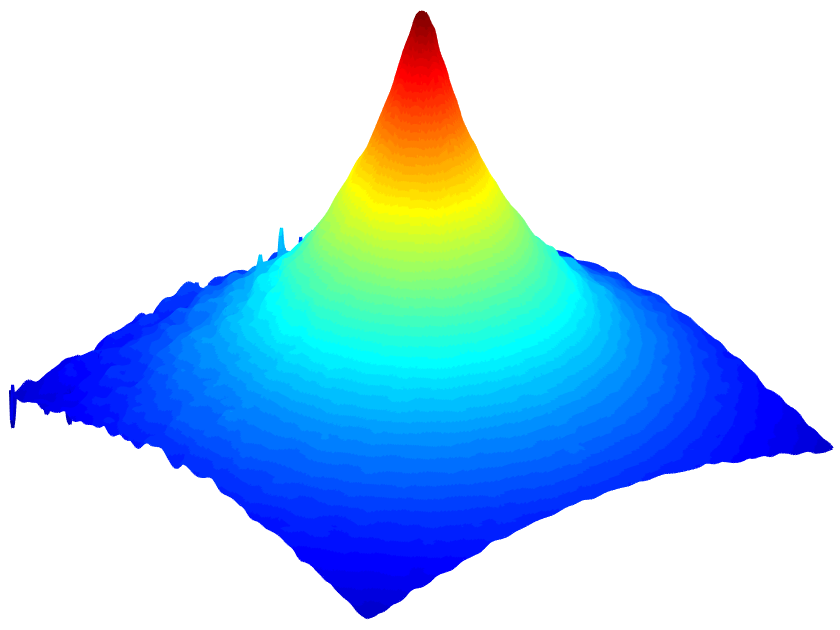}
		\end{minipage}
	}
	\subfloat[$\beta = 2$]{
		\begin{minipage}[b]{0.24\textwidth}
			\includegraphics[width=4.0cm,height=2.8cm]{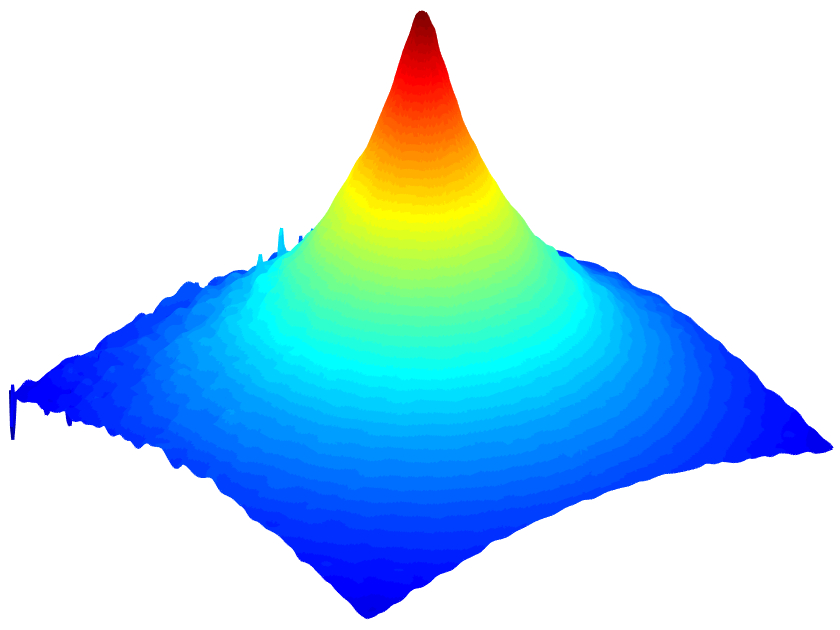}
		\end{minipage}
	}
	%	\vspace{0.1cm}
	\caption{Depth maps of N-Cone in different $\beta$ values. $\zeta = 3$, $\lambda_0 = 50$ in all above the proposed filters by fix different adaptive amplification factors (a) $\beta = 1/4$, (b) $\beta = 1/2$, (c) $\beta = 3/4$, (d) $\beta = 1$, (e) $\beta = 5/4$, and (f) $\beta = 3/2$.}
	\label{fig:beta_noise}
	\vspace*{-0.4cm}
\end{figure*}

\subsection{AWGIF for Depth Enhancement in SFF}\label{Enhancement}
In the following, the proposed AWGIF algorithm is introduced to decompose the initial depth map $Z$ into a base layer $Z_b$ and a detail layer $Z_d$ including noise. This process is shown in Fig. \ref{fig:framework}. Let $G$ be a structure guidance image which is selected by comparing the quality of the depth maps attained from different guidance maps \cite{ali2021guided}. The least root mean square error and highest correlation for the resultant depth maps have been obtained by mean image intensity along optical dimension.Therefore, the mean image intensity in the multi-focus image sequence $S$ along k-direction is the optimal one which is selected as the structure guidance map $G$ for our proposed strategy, and is described as
\begin{flalign}
\label{eq:SFF_guidance}
&G(p) = \frac{1}{K}\sum\nolimits_{k = 1}^K {{S_k}(p)},&
\end{flalign}		
Then, the base layer image $Z_b$ is calculated by the linear transformation of the guidance image $G$, which is expressed as
\begin{flalign}
\label{detail_Zb}
&Z_b = {\bar{a} _p}G(p) + {\bar{b}_p},&
\end{flalign}
where $\bar{a}_p$ and $\bar{b}_p$ are two constants which can be derived from equation (\ref{average_ap&bp}).

Thus, the erroneous depth estimation is compensated by incorporating the guidance information about the structure of the scene. Meanwhile, due to the advantages of the proposed AWGIF algorithm, the final depth map can not only suppress the halo artifact, but also preserve clearer and sharper depth edges.

For the detail layer, when the amplification factor $\alpha+\beta\bar{a}$ in equation (\ref{detailenhancement}) is small, the depth details are suppressed. If the amplification factor is large, the noise is amplified. In order to balance between retaining fine details and suppressing the noise, the values of two constants $\alpha$ and $\beta$ in the amplification factor are discussed through the four cases mentioned above. The results of the discussion are that 1) the output depth map $Z_f$ lacks details in the case 1; 2) the noise is amplified when the value of $\alpha$ increases in the case 2; 3) the noise is still amplified because the value of $\alpha$ is large in the case 3; 4) the output depth map $Z_f$ in the case 4 is better in suppressing noise and preserving details than the other three cases because of the adaptive amplification factor $\beta\bar{a}$. Therefore, the final value of $Z_f$ is given as follows,
\begin{flalign}
\label{detailenhancement2}
&Z_f=Z_b+\beta\bar{a}\odot Z_d,&
\end{flalign}
where the selection of $\beta$ value will be discussed in the experimental results and analysis section. Finally, the improved depth map can be enhanced by the equation (\ref{detailenhancement2}). Clearly, the coefficient of the proposed AWGIF are used smartly by the proposed depth enhancement algorithm.

\begin{figure*}[htp]
	\centering
	%\vspace*{-0.5cm}
	\subfloat[GT]{
		\begin{minipage}[b]{0.138\textwidth}
			\begin{overpic}[width=2.4cm,height=1.8cm]{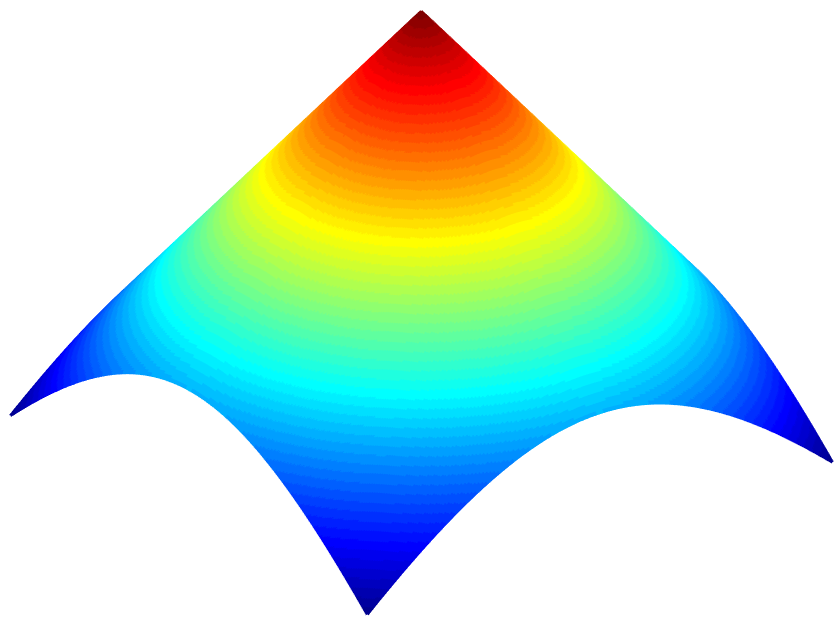}\vspace{.01in}
			\end{overpic}\vspace{.01in}
			
			\begin{overpic}[width=2.4cm,height=1.8cm]{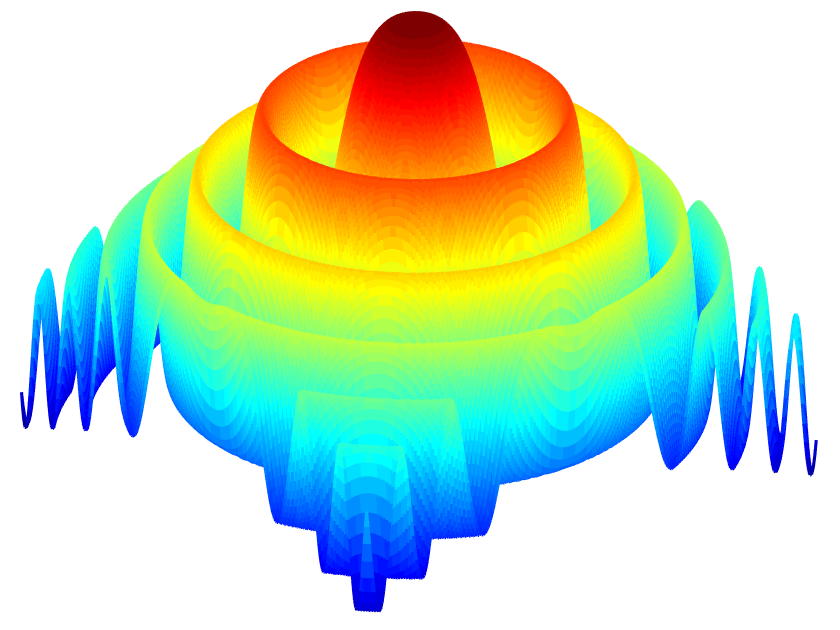}\vspace{.01in}
			\end{overpic}\vspace{.01in}
			
			\begin{overpic}[width=2.4cm,height=1.8cm]{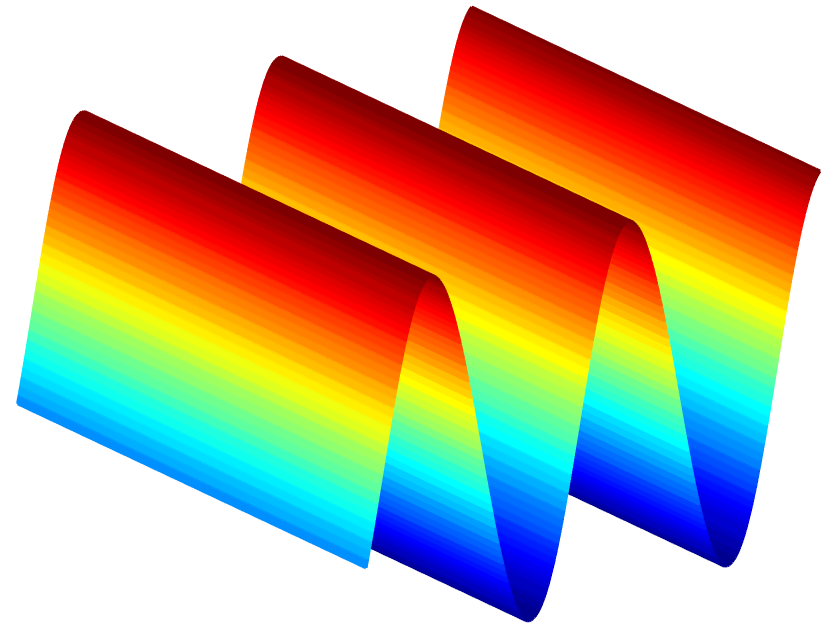}
			\end{overpic}\vspace{.01in}
		\end{minipage}
	}\hspace{-.1in}
	\subfloat[Initial]{
		\begin{minipage}[b]{0.138\textwidth}
			\begin{overpic}[width=2.4cm,height=1.8cm]{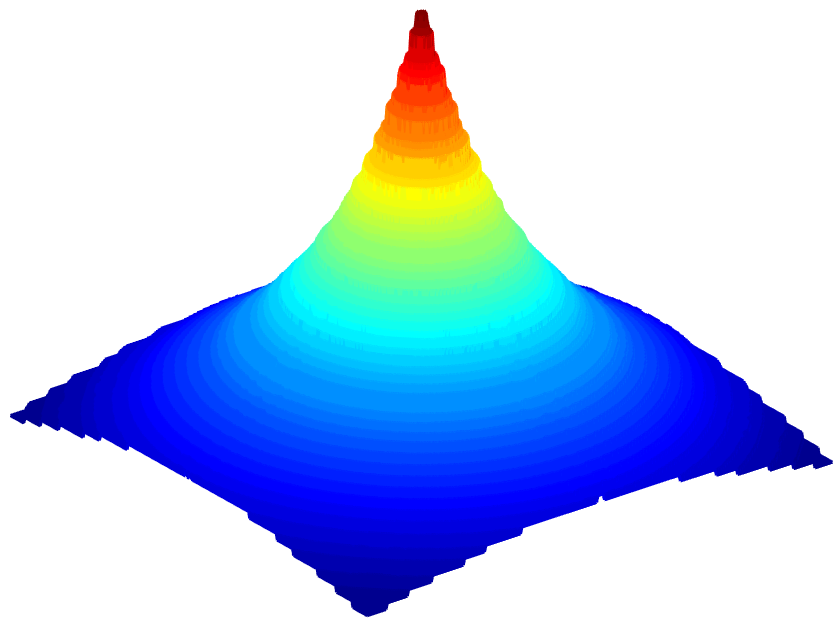}\vspace{.01in}
			\end{overpic}\vspace{.01in}
			
			\begin{overpic}[width=2.4cm,height=1.8cm]{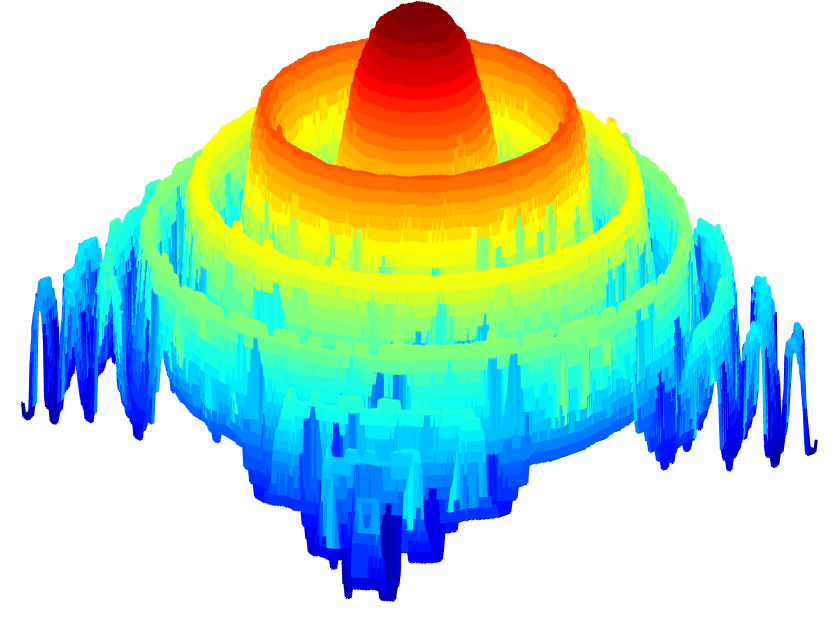}\vspace{.01in}
			\end{overpic}\vspace{.01in}
			
			\begin{overpic}[width=2.4cm,height=1.8cm]{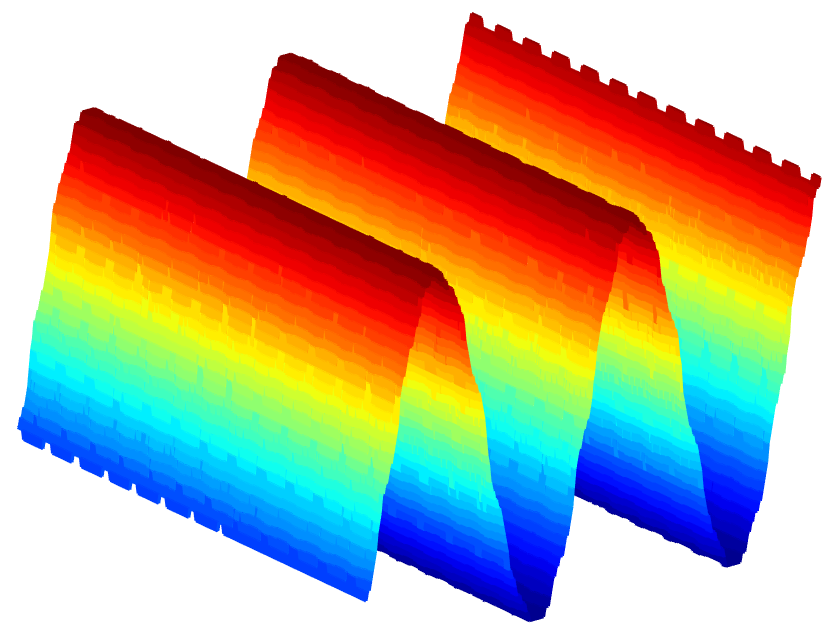}
			\end{overpic}\vspace{.01in}
		\end{minipage}
	} \hspace{-.1in}
	\subfloat[GIF]{
		\begin{minipage}[b]{0.138\textwidth}
			\begin{overpic}[width=2.4cm,height=1.8cm]{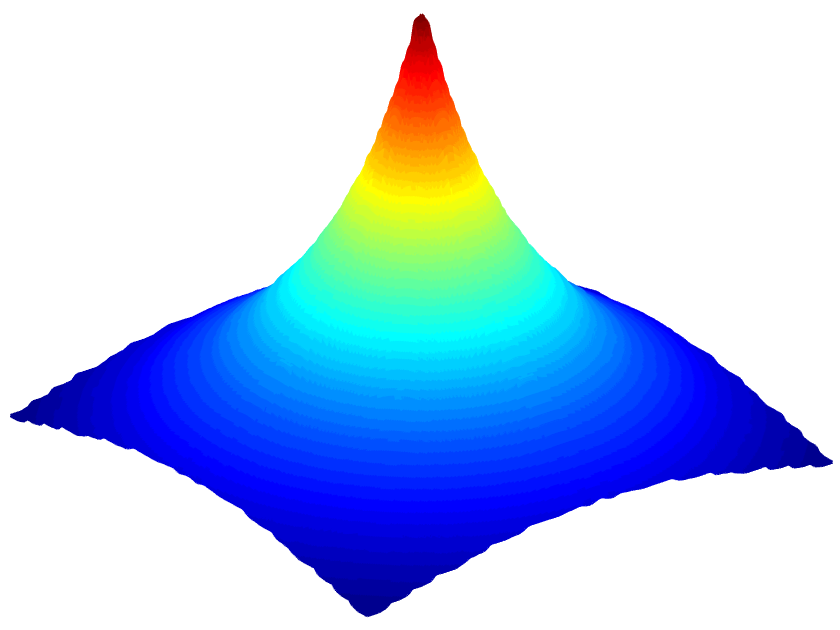}\vspace{.01in}
			\end{overpic}\vspace{.01in}
			
			\begin{overpic}[width=2.4cm,height=1.8cm]{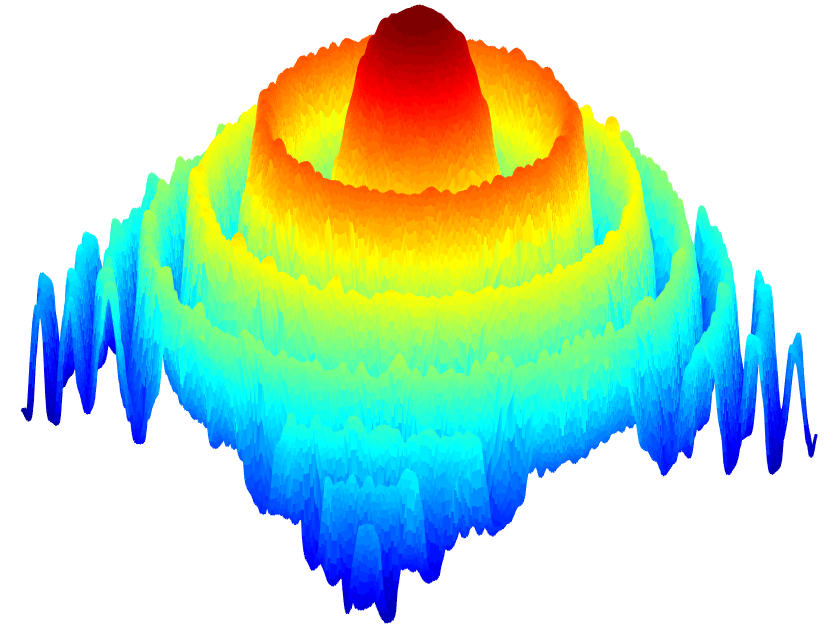}\vspace{.01in}
			\end{overpic}\vspace{.01in}
			
			\begin{overpic}[width=2.4cm,height=1.8cm]{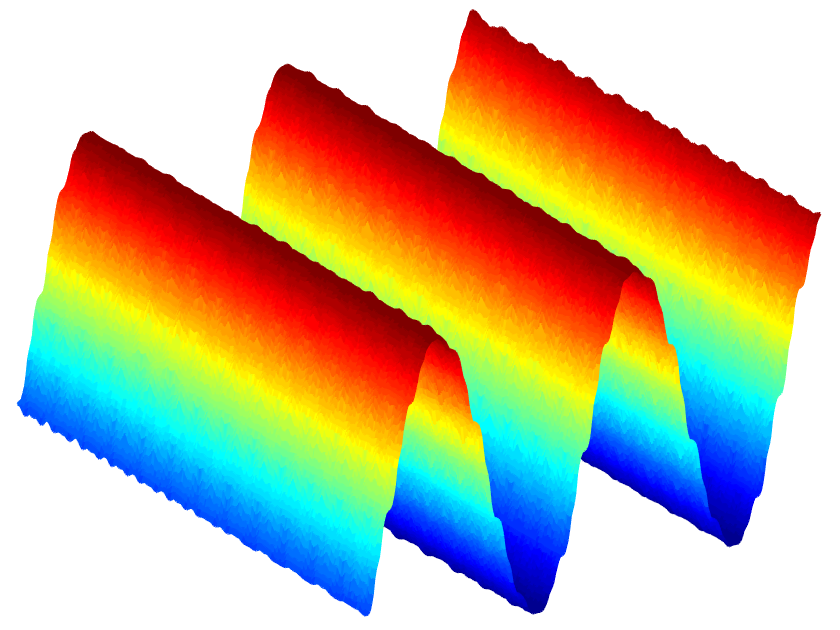}
			\end{overpic}\vspace{.01in}
		\end{minipage}
	}\hspace{-.1in}
	\subfloat[WGIF]{
		\begin{minipage}[b]{0.138\textwidth}
			\begin{overpic}[width=2.4cm,height=1.8cm]{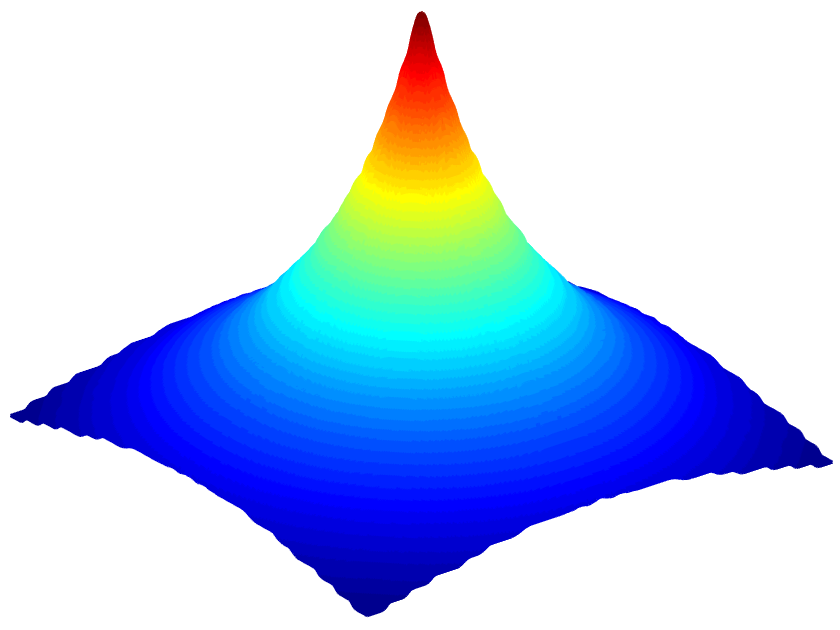}\vspace{.01in}
			\end{overpic}\vspace{.01in}
			
			\begin{overpic}[width=2.4cm,height=1.8cm]{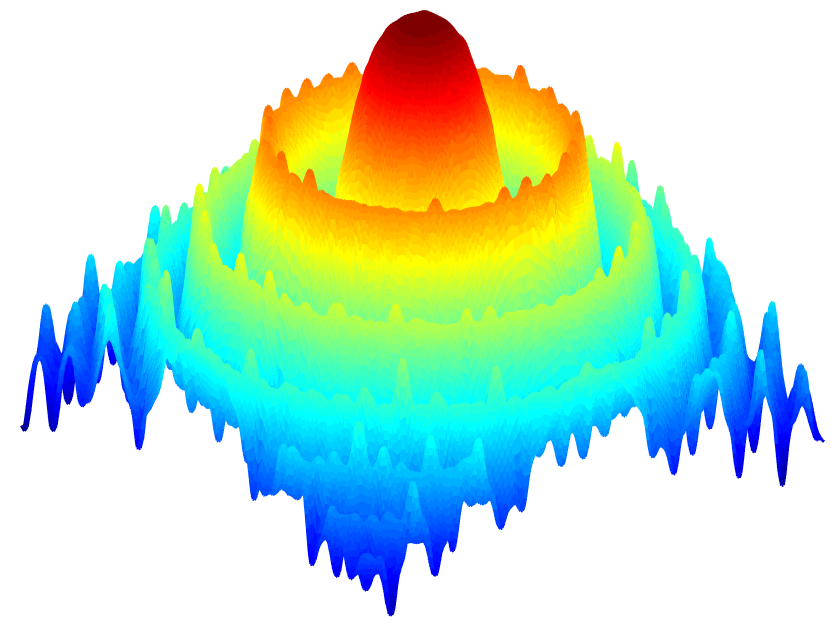}\vspace{.01in}
			\end{overpic}\vspace{.01in}
			
			\begin{overpic}[width=2.4cm,height=1.8cm]{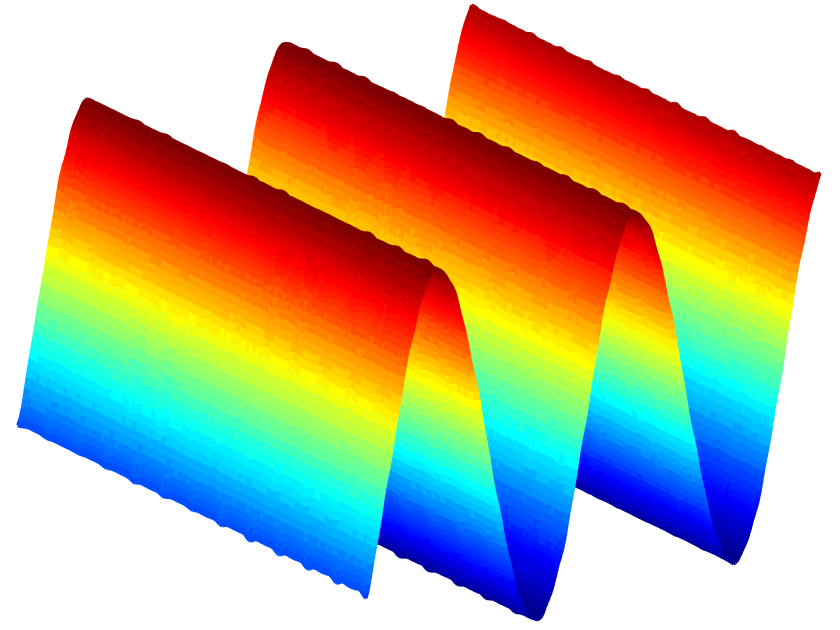}
			\end{overpic}\vspace{.01in}
		\end{minipage}
	}\hspace{-.1in}
	\subfloat[EGIF]{
		\begin{minipage}[b]{0.138\textwidth}
			\begin{overpic}[width=2.4cm,height=1.8cm]{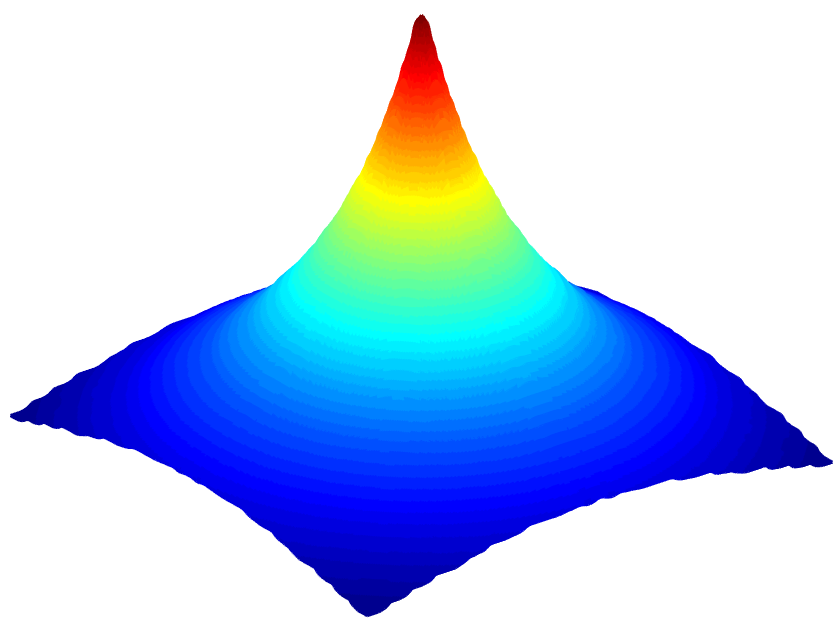}\vspace{.01in}
			\end{overpic}\vspace{.01in}
			
			\begin{overpic}[width=2.4cm,height=1.8cm]{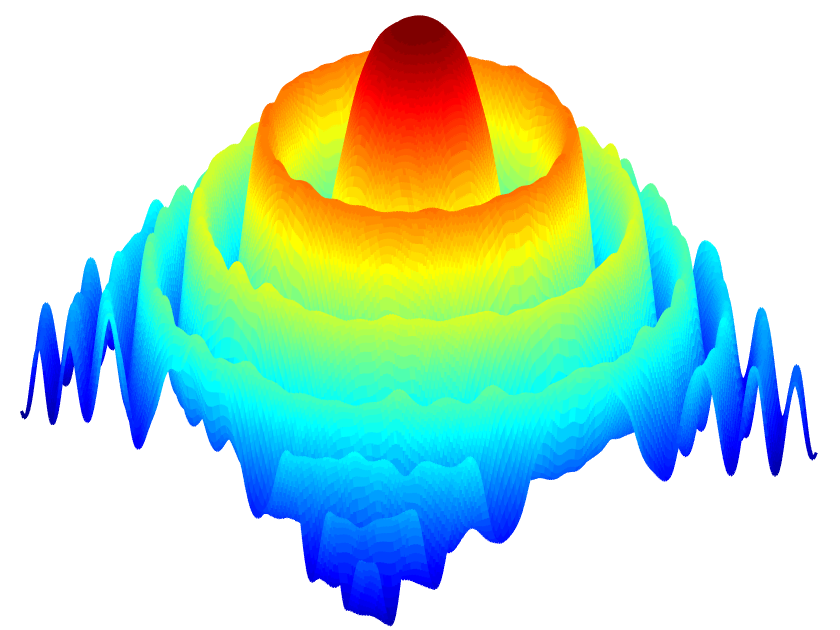}\vspace{.01in}
			\end{overpic}\vspace{.01in}
			
			\begin{overpic}[width=2.4cm,height=1.8cm]{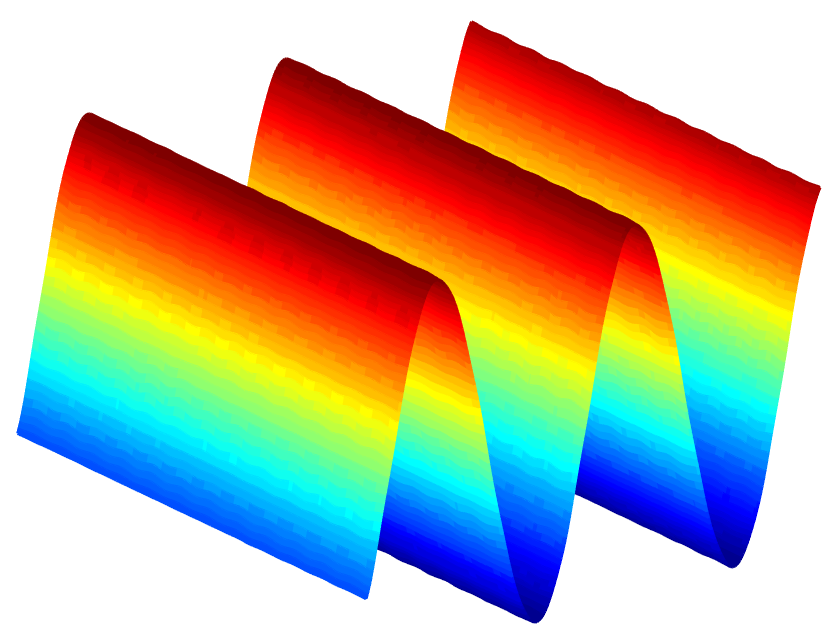}
			\end{overpic}\vspace{.01in}
		\end{minipage}
	}\hspace{-.1in}
	\subfloat[WAGIF]{
		\begin{minipage}[b]{0.138\textwidth}
			\begin{overpic}[width=2.4cm,height=1.8cm]{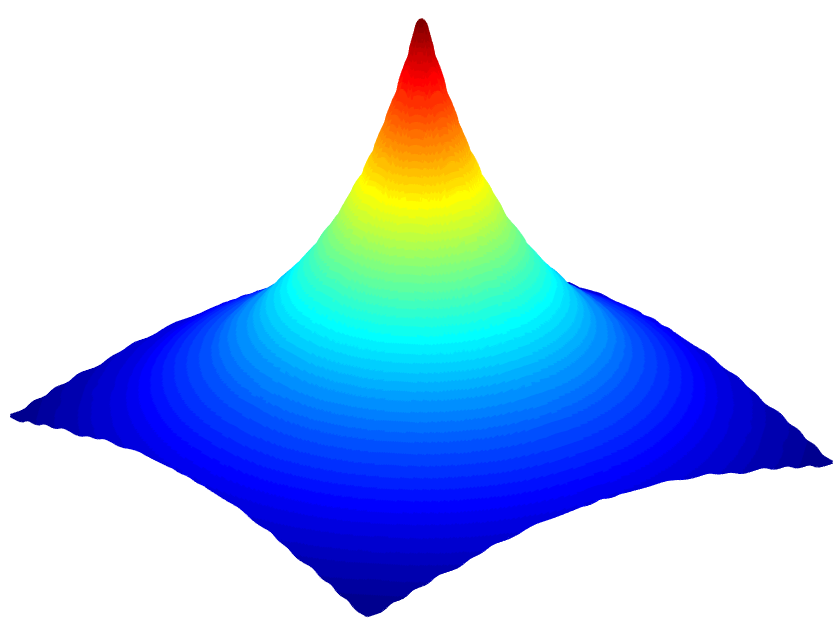}\vspace{.01in}	
			\end{overpic}\vspace{.01in}
			
			\begin{overpic}[width=2.4cm,height=1.8cm]{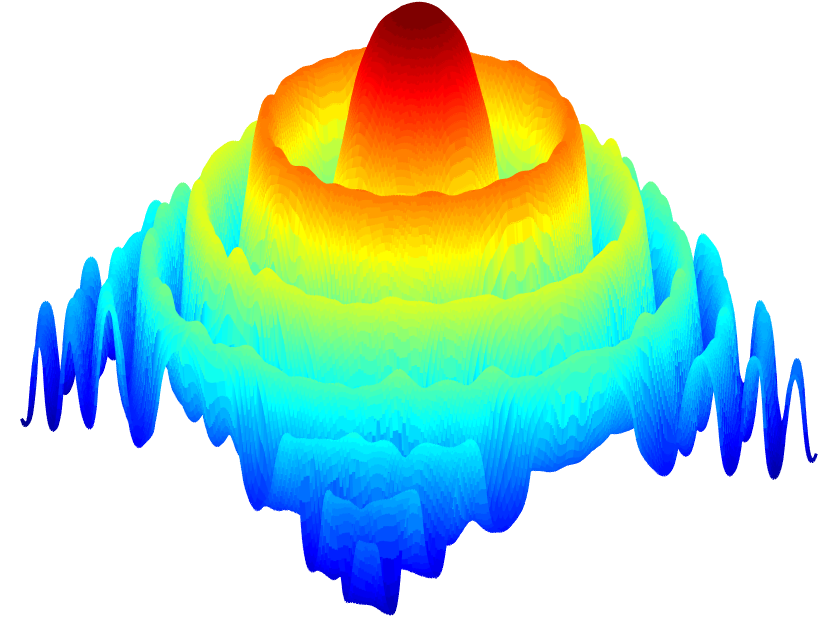}\vspace{.01in}
			\end{overpic}\vspace{.01in}
			
			\begin{overpic}[width=2.4cm,height=1.8cm]{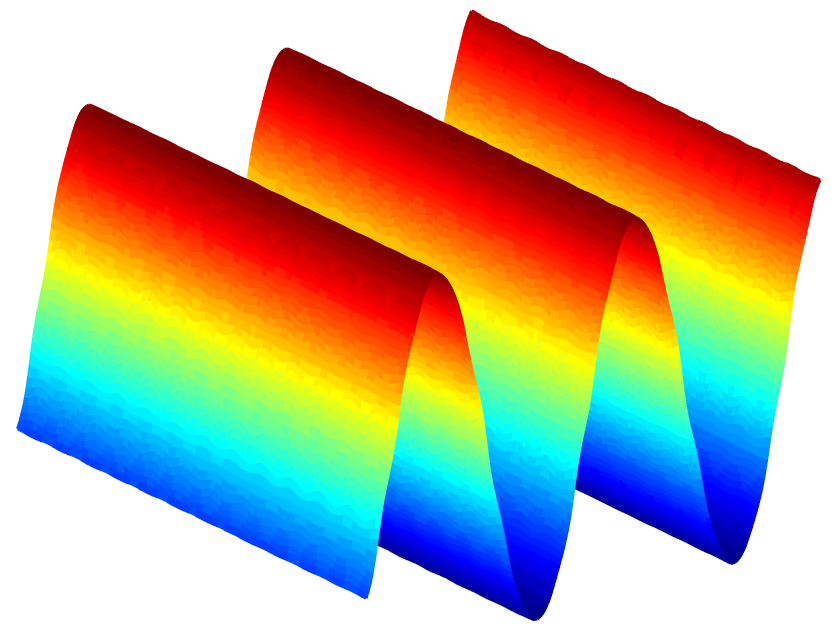}
			\end{overpic}\vspace{.01in}
		\end{minipage}
	}\hspace{-.1in}
	\subfloat[Ours]{
		\begin{minipage}[b]{0.138\textwidth}
			\begin{overpic}[width=2.4cm,height=1.8cm]{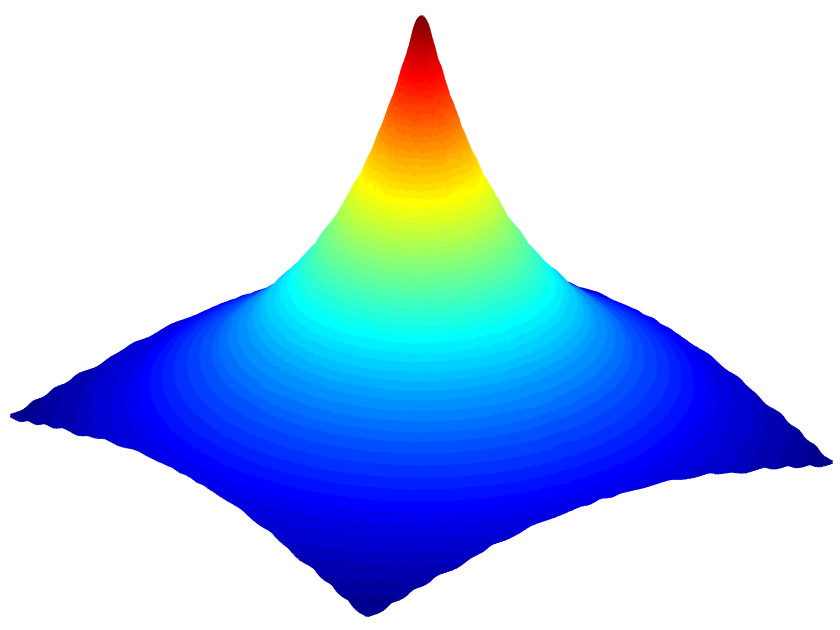}\vspace{.01in}	
			\end{overpic}\vspace{.01in}
			
			\begin{overpic}[width=2.4cm,height=1.8cm]{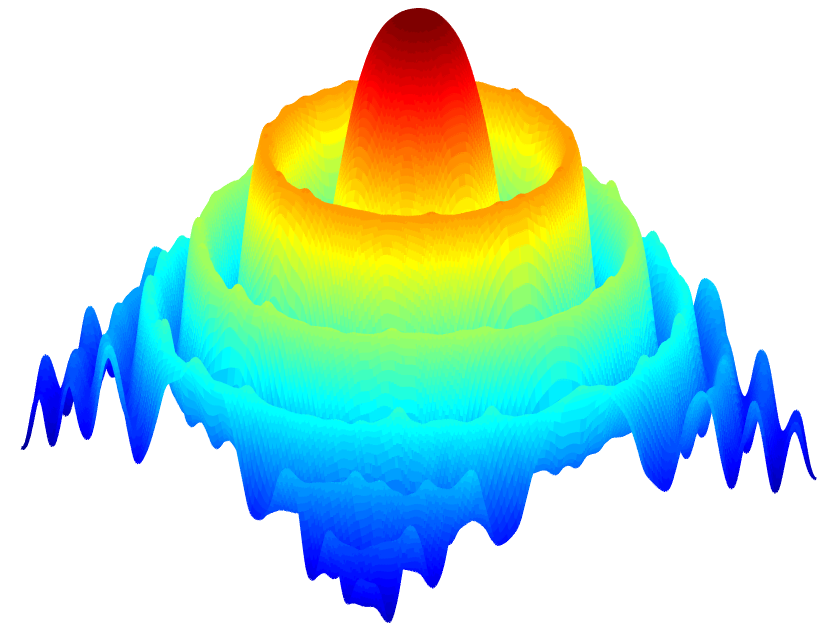}\vspace{.01in}
			\end{overpic}\vspace{.01in}
			
			\begin{overpic}[width=2.4cm,height=1.8cm]{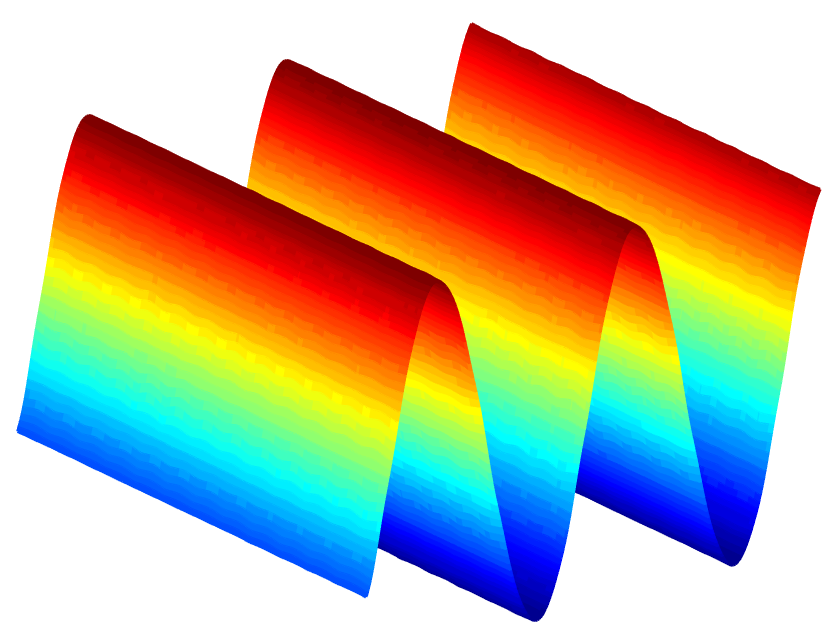}
			\end{overpic}\vspace{.01in}
		\end{minipage}
	}
	%\vspace{0.1cm}
	\caption{Depth maps of synthetic image sequences. The cone, cos and sine objects correspond from the first row to the third row respectively. (a) Ground truth images. (b) initial depth maps. The depth maps of cone, cos and sine by (c) GIF, (d) WGIF, (e) EGIF, (f) WAGIF, and (g) Ours.}
	\label{fig:SyntheticClean_Filter}
	\vspace*{-0.4cm}
\end{figure*}

\begin{figure*}[htp]
	\centering
	%\vspace*{-0.5cm}
	\subfloat[GD]{
		\begin{minipage}[b]{0.138\textwidth}
			\begin{overpic}[width=2.42cm,height=1.8cm]{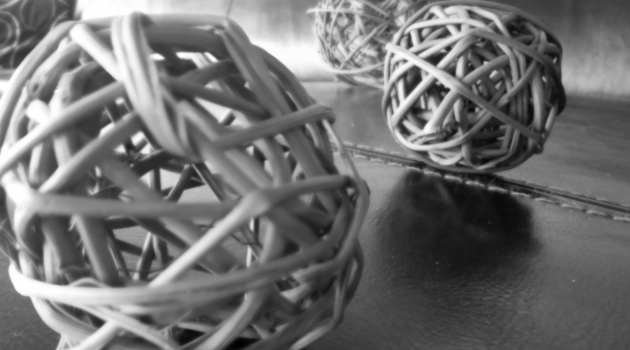}\vspace{.01in}
			\end{overpic}\vspace{.01in}
			
			\begin{overpic}[width=2.42cm,height=1.8cm]{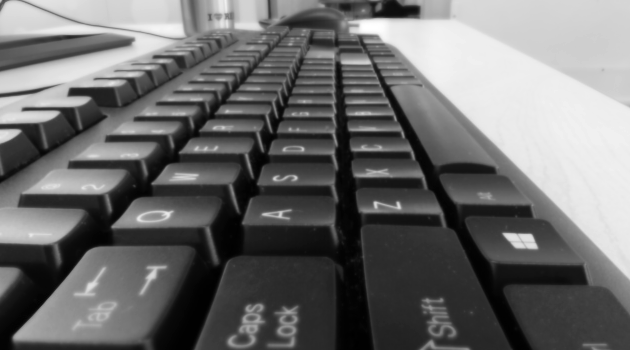}\vspace{.01in}
			\end{overpic}\vspace{.01in}
			
			\begin{overpic}[width=2.42cm,height=1.8cm]{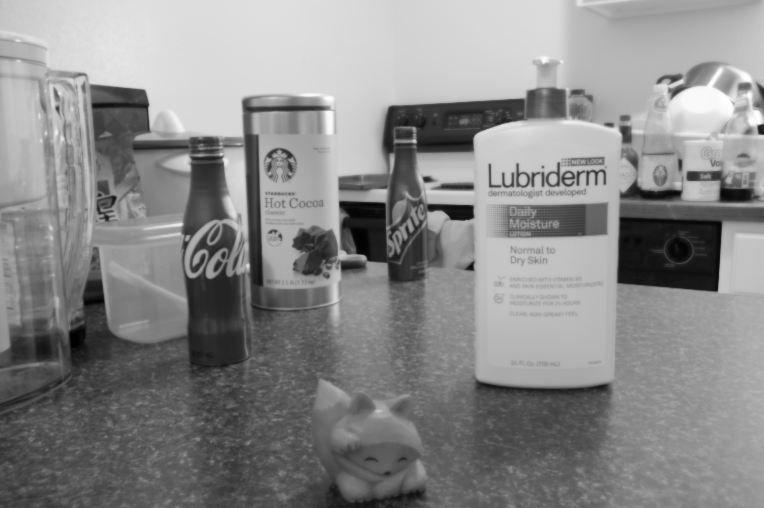}
			\end{overpic}\vspace{.01in}
		\end{minipage}
	}\hspace{-.1in}
	\subfloat[Initial]{
		\begin{minipage}[b]{0.138\textwidth}
			\begin{overpic}[width=2.42cm,height=1.8cm]{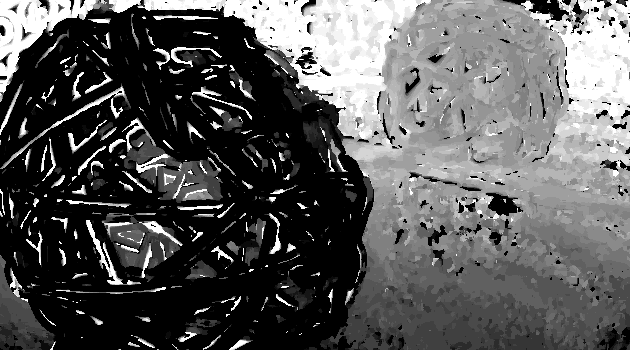}\vspace{.01in}
			\end{overpic}\vspace{.01in}
			
			\begin{overpic}[width=2.42cm,height=1.8cm]{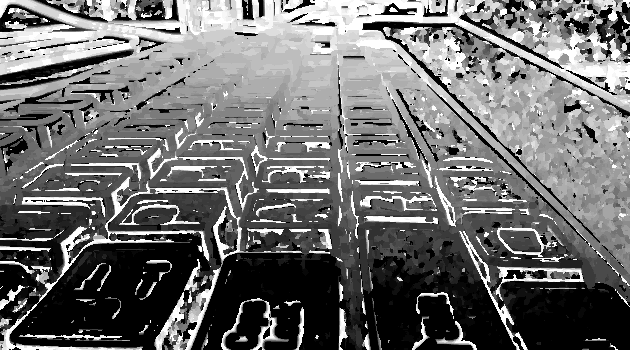}\vspace{.01in}
			\end{overpic}\vspace{.01in}
			
			\begin{overpic}[width=2.42cm,height=1.8cm]{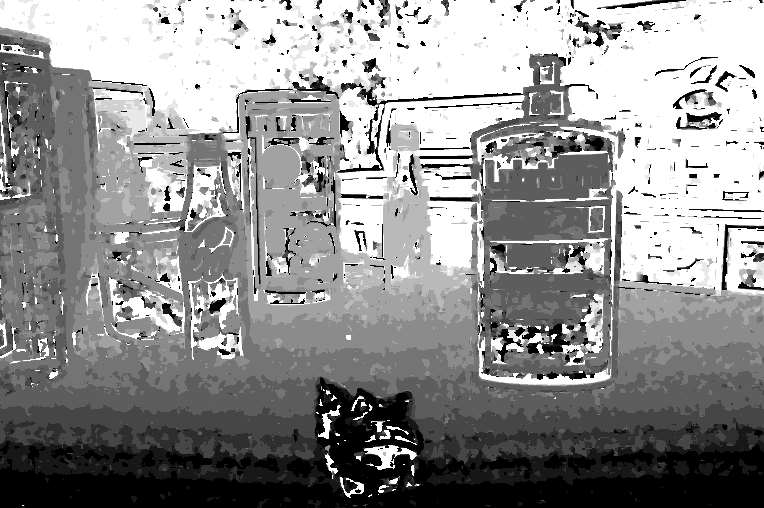}
			\end{overpic}\vspace{.01in}
		\end{minipage}
	} \hspace{-.1in}
	\subfloat[GIF]{
		\begin{minipage}[b]{0.138\textwidth}
			\begin{overpic}[width=2.42cm,height=1.8cm]{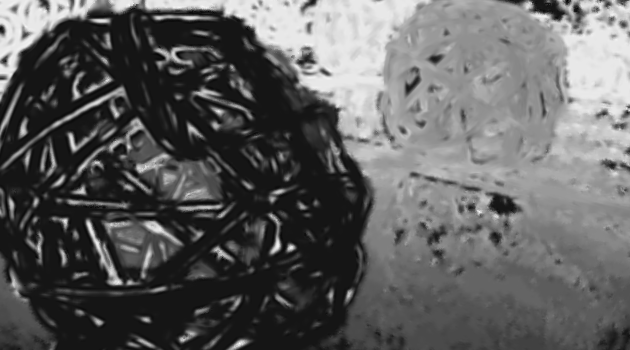}\vspace{.01in}
			\end{overpic}\vspace{.01in}
			
			\begin{overpic}[width=2.42cm,height=1.8cm]{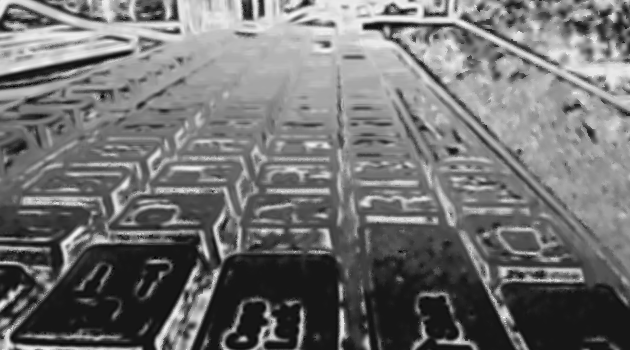}\vspace{.01in}
			\end{overpic}\vspace{.01in}
			
			\begin{overpic}[width=2.42cm,height=1.8cm]{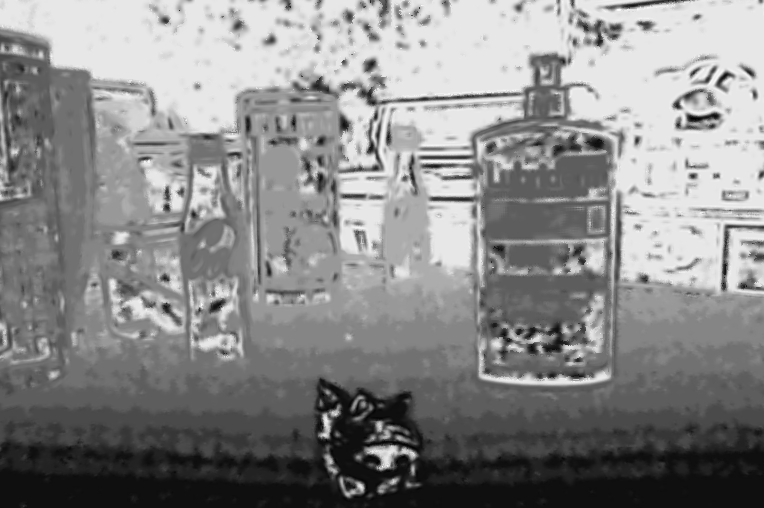}
			\end{overpic}\vspace{.01in}
		\end{minipage}
	}\hspace{-.1in}
	\subfloat[WGIF]{
		\begin{minipage}[b]{0.138\textwidth}
			\begin{overpic}[width=2.42cm,height=1.8cm]{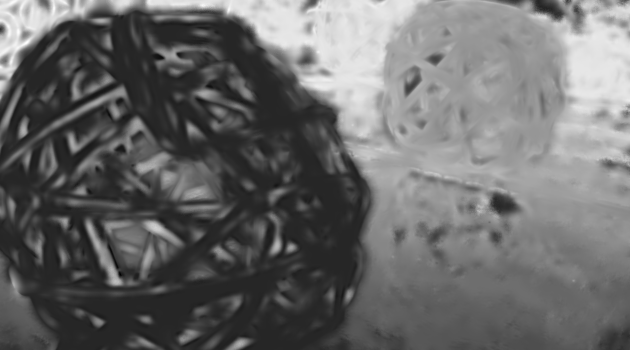}\vspace{.01in}
			\end{overpic}\vspace{.01in}
			
			\begin{overpic}[width=2.42cm,height=1.8cm]{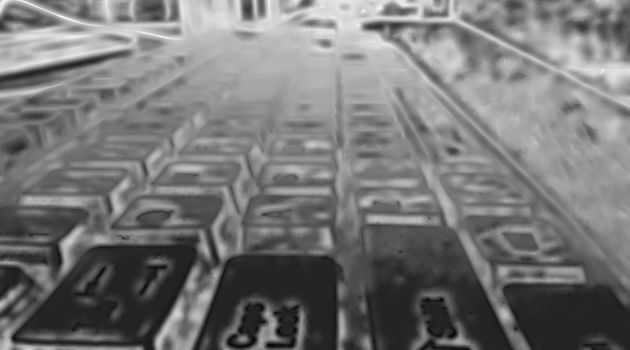}\vspace{.01in}
			\end{overpic}\vspace{.01in}
			
			\begin{overpic}[width=2.42cm,height=1.8cm]{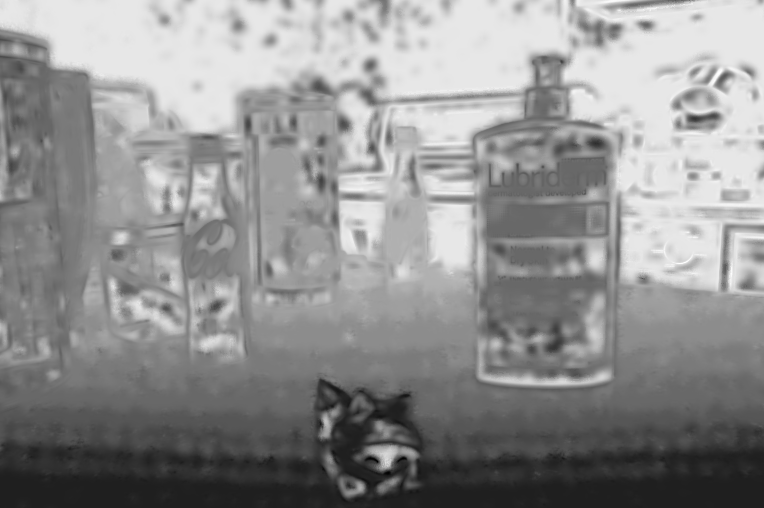}
			\end{overpic}\vspace{.01in}
		\end{minipage}
	}\hspace{-.1in}
	\subfloat[EGIF]{
		\begin{minipage}[b]{0.138\textwidth}
			\begin{overpic}[width=2.42cm,height=1.8cm]{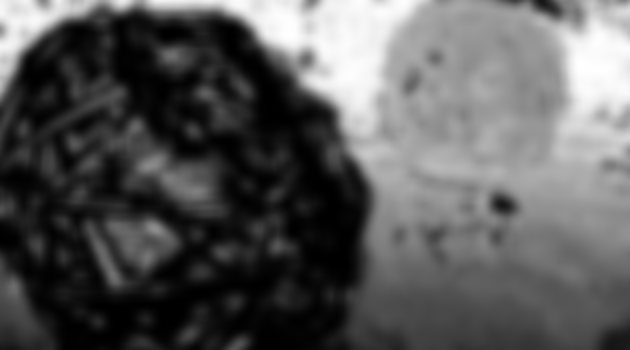}\vspace{.01in}
			\end{overpic}\vspace{.01in}
			
			\begin{overpic}[width=2.42cm,height=1.8cm]{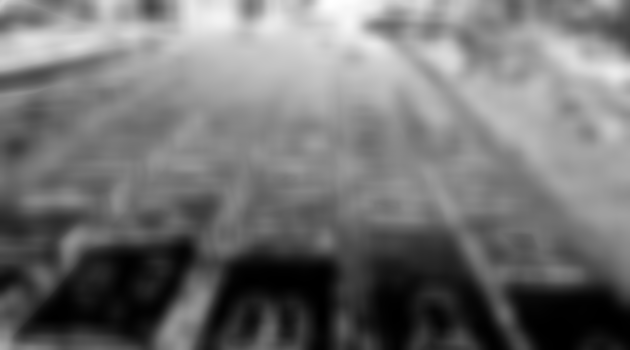}\vspace{.01in}
			\end{overpic}\vspace{.01in}
			
			\begin{overpic}[width=2.42cm,height=1.8cm]{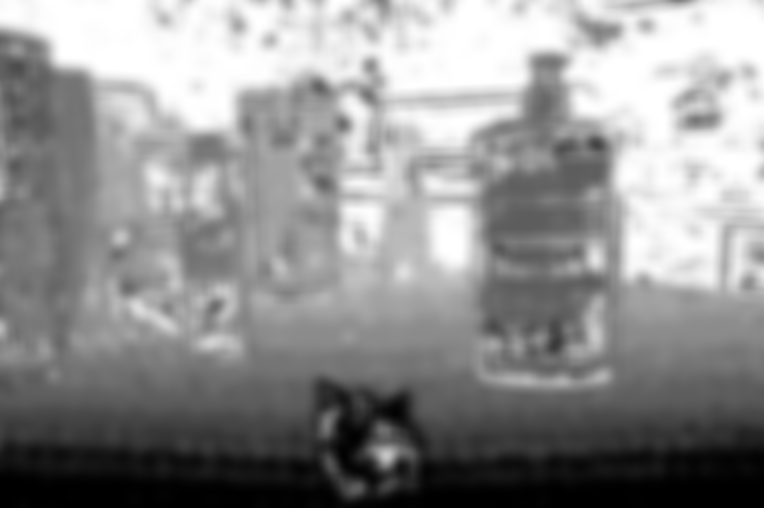}
			\end{overpic}\vspace{.01in}
		\end{minipage}
	}\hspace{-.1in}
	\subfloat[WAGIF]{
		\begin{minipage}[b]{0.138\textwidth}
			\begin{overpic}[width=2.42cm,height=1.8cm]{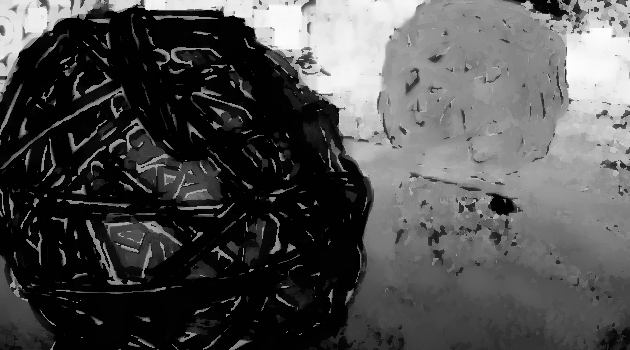}\vspace{.01in}	
			\end{overpic}\vspace{.01in}
			
			\begin{overpic}[width=2.42cm,height=1.8cm]{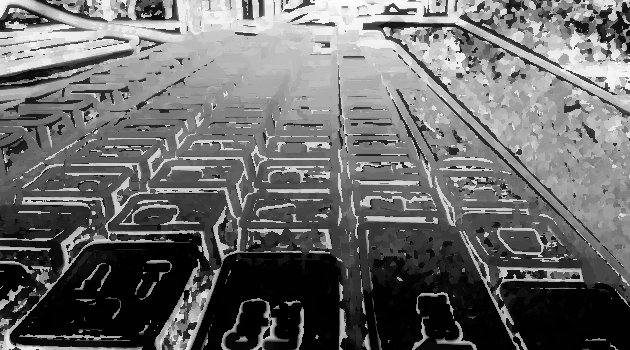}\vspace{.01in}
			\end{overpic}\vspace{.01in}
			
			\begin{overpic}[width=2.42cm,height=1.8cm]{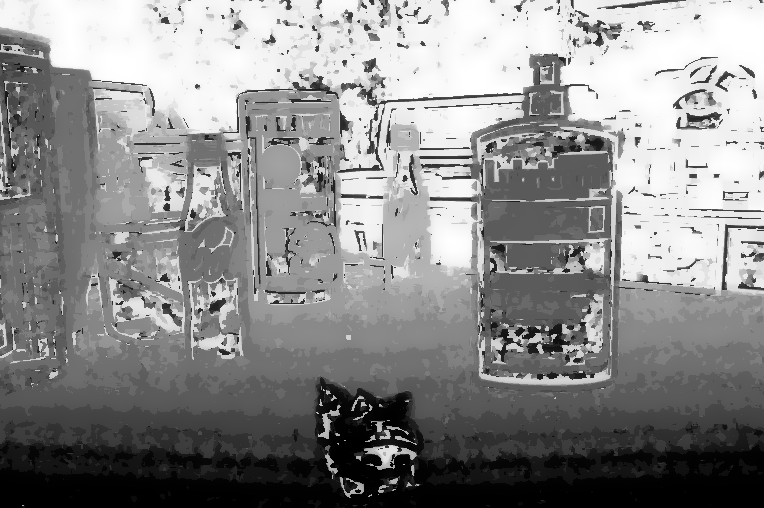}
			\end{overpic}\vspace{.01in}
		\end{minipage}
	}\hspace{-.1in}
	\subfloat[Ours]{
		\begin{minipage}[b]{0.138\textwidth}
			\begin{overpic}[width=2.42cm,height=1.8cm]{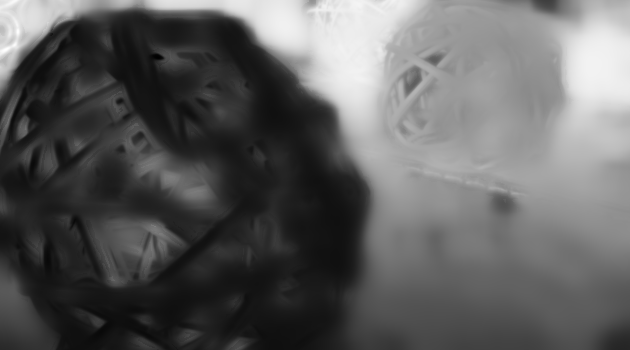}\vspace{.01in}	
			\end{overpic}\vspace{.01in}
			
			\begin{overpic}[width=2.42cm,height=1.8cm]{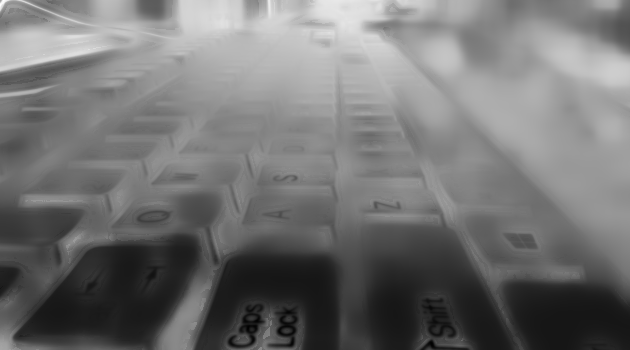}\vspace{.01in}
			\end{overpic}\vspace{.01in}
			
			\begin{overpic}[width=2.42cm,height=1.8cm]{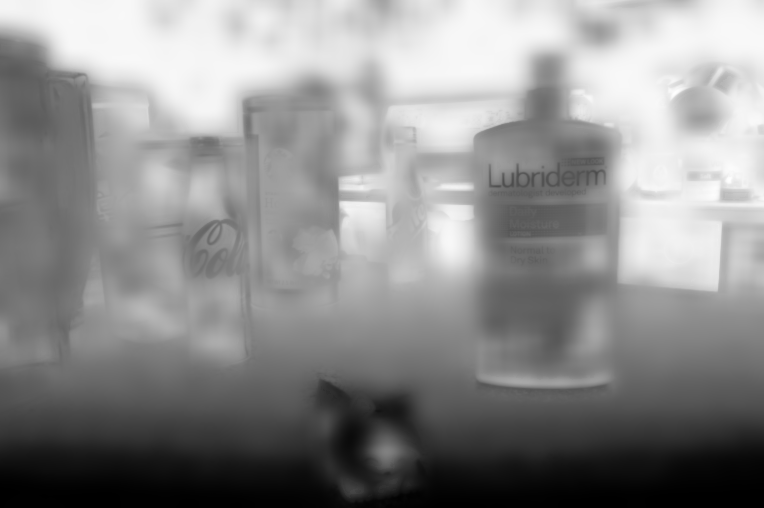}
			\end{overpic}\vspace{.01in}
		\end{minipage}
	}
	%\vspace{0.1cm}
	\caption{Depth maps of real image sequences. The BA, KB and KC images correspond from the first row to the third row respectively. (a) The guidance maps. (b) Initial depth maps. The depth maps of BA, KB and KC by (c) GIF, (d) WGIF, (e) EGIF, (f) WAGIF, and (g) Ours.}
	\label{fig:RealClean_Filter}
	\vspace*{-0.4cm}
\end{figure*}

\section{Experimental Results and Analysis}\label{Experiment}
In this section, the performance of the proposed AWGIF algorithm for depth enhancement in SFF is evaluated experimentally. First of all, the datasets of image sequences and the quality evaluation indicators of depth map are setup for the experiment in Section \ref{Setup}. Then, the effects of different adaptive amplification factors on the proposed algorithm are discussed through extensive experiments in Section \ref{Effect}. The adaptive amplification factor that produced the best results is selected and applied to all experiments later. In Section \ref{Comparative}, the proposed algorithm is compared with various advanced guided filtering algorithms, and qualitative and quantitative analysis is carried out for synthetic and real images. In order to verify the robustness of the proposed algorithm, the performances of various guided filters in noisy image sequences are compared in Section \ref{Robustness}. Finally, the computational times of various filters are compared in Section \ref{Computation}.

\subsection{Experimental Setup}\label{Setup}
Experimental datasets consist of three synthetic image sequences and three real image sequences from \cite{mahmood2012nonlinear,suwajanakorn2015depth}. Synthetic image sequences are shaped like cone, cos and sine wave, corresponding to contain 97, 60 and 60 images of dimension 360$\times$360, 300$\times$300 and 300$\times$300 pixels each, while real image sequences are named as ball (BA), keyboard (KB) and kitchen (KC) which have 25, 32 and 12 images respectively with dimensions of 640$\times$360, 640$\times$360 and 774$\times$518 pixels. In order to evaluate the robustness of the proposed algorithm, sine and cos wave image sequences have been garbled with the Gaussian white noise of 0.005 variance, while cone image sequences and all real images with 0.02 and 0.003 variance, respectively. For the convenience of later description, these noisy images are correspondingly called N-Sine, N-Cos, N-Cone, N-BA, N-KB and N-KC.   

For verifying the performance of the proposed algorithm, root mean square error (RMSE) and correlation (CORR) are computed as indicators of quantitative measurement (QM) in the case of the synthetic images, bacause they have ground truth depth map. While root mean square difference (RMSD) has been computed between the final depth map and initial depth map in the case of the real images \cite{ali2021guided}. The three quantitative analysis indicators are expressed as follows,
\begin{flalign}
\label{RMSE}
&RMSE = \sqrt {\frac{1}{{UV}}\sum\limits_{u = 1}^U {\sum\limits_{v = 1}^V {{{\left( {{Z_f}(u,v) - {Z_g}(u,v)} \right)}^2}} } },&
\end{flalign}
where ${Z_f}(u,v)$ represents the improved depth map, ${Z_g}(u,v)$ denotes the ground truth depth maps, amd UV is the total number of pixels in the image. Small value of the root mean
square error indicates high quality of the improved depth map.

\begin{figure*}[htp]
	\centering
	%\vspace*{-0.5cm}
	\subfloat[Cone and N-Cone]{
		\begin{minipage}[b]{0.5\textwidth}
			\includegraphics[width=8cm,height=4.5cm]{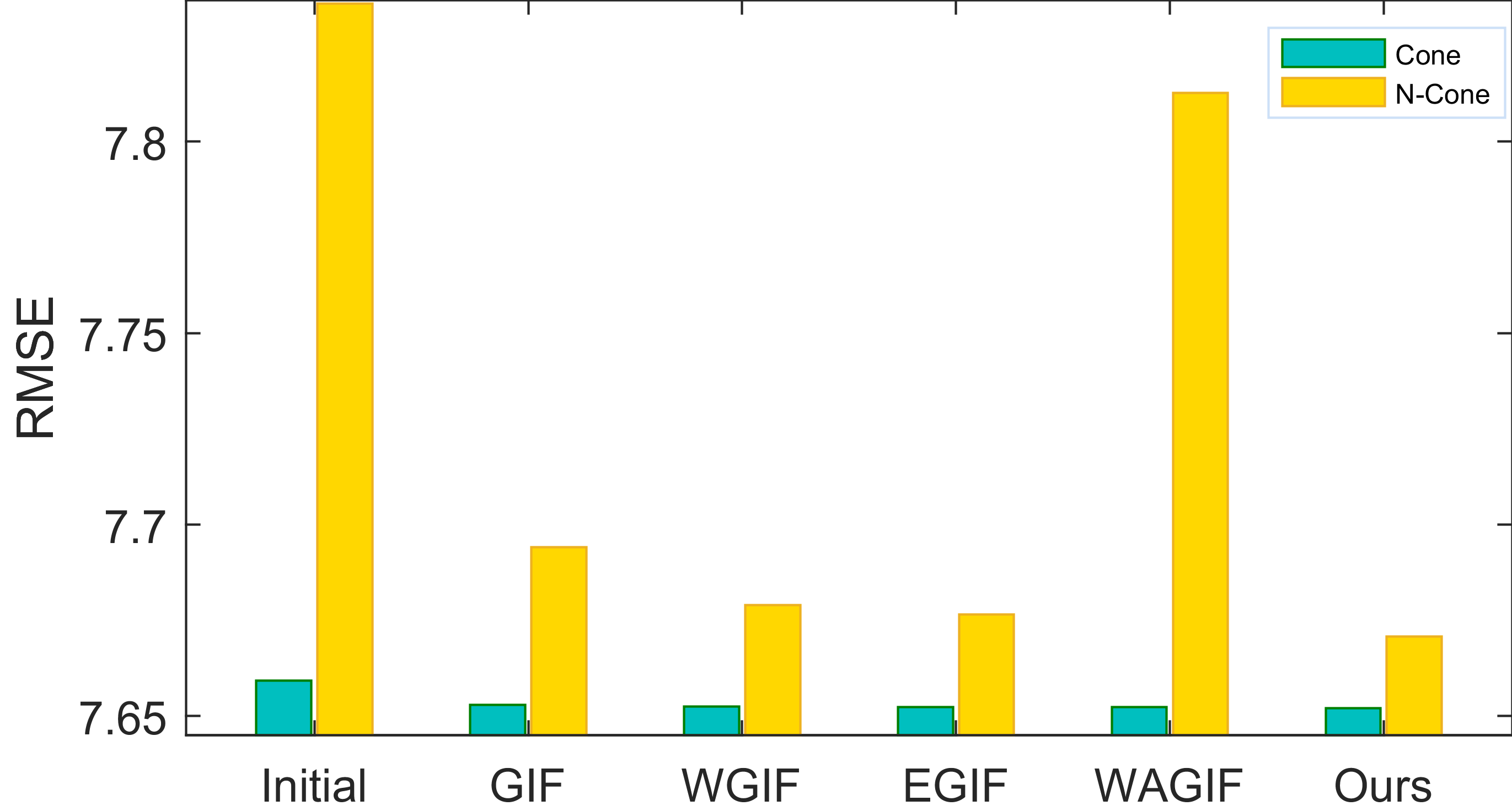}
		\end{minipage}
	}
	\subfloat[Cone and N-Cone]{
		\begin{minipage}[b]{0.5\textwidth}
			\includegraphics[width=8cm,height=4.5cm]{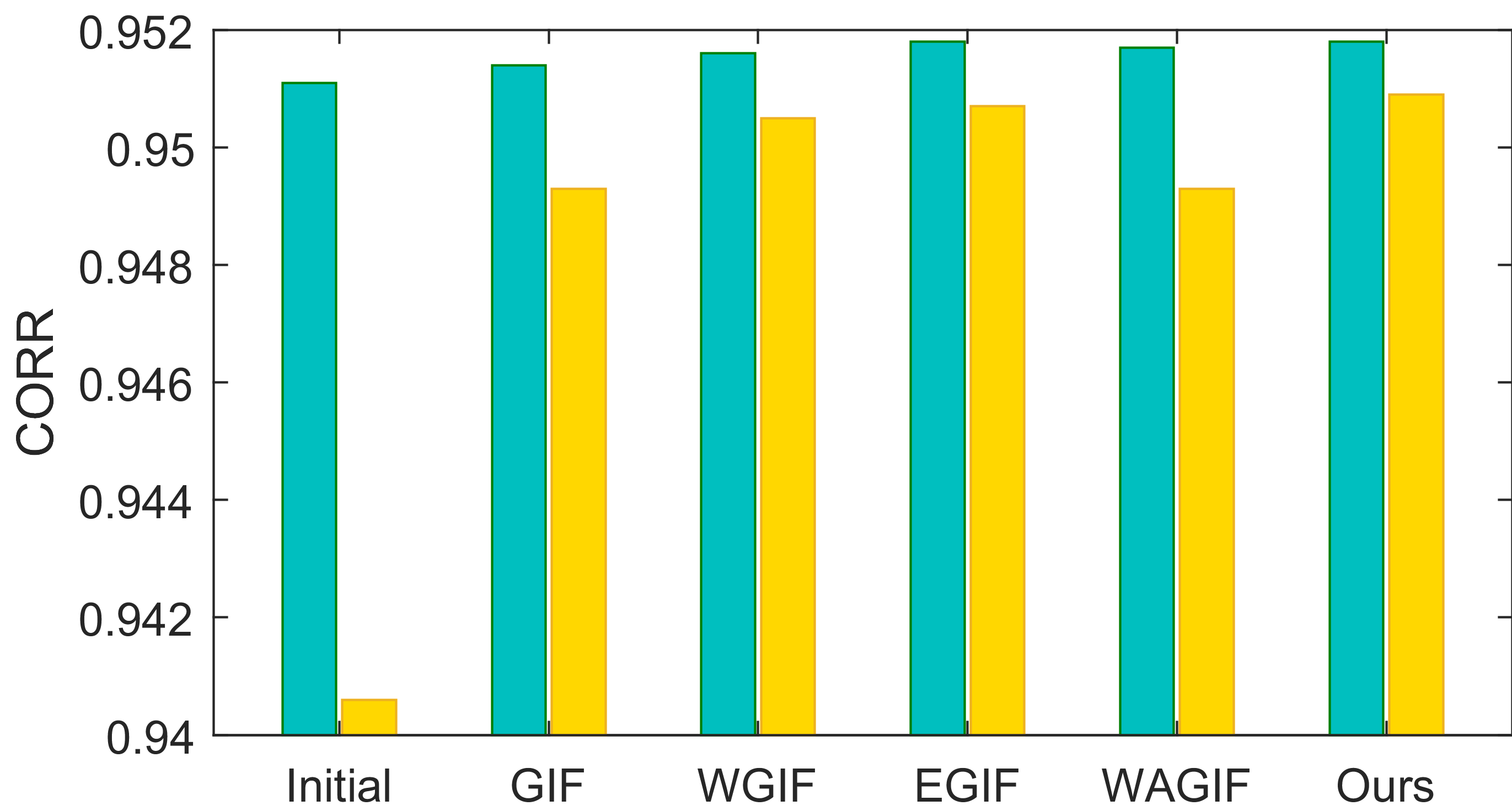}
		\end{minipage}
	}\hspace{1in}
	\subfloat[Cos and N-Cos]{
		\begin{minipage}[b]{0.5\textwidth}
			\includegraphics[width=8cm,height=4.5cm]{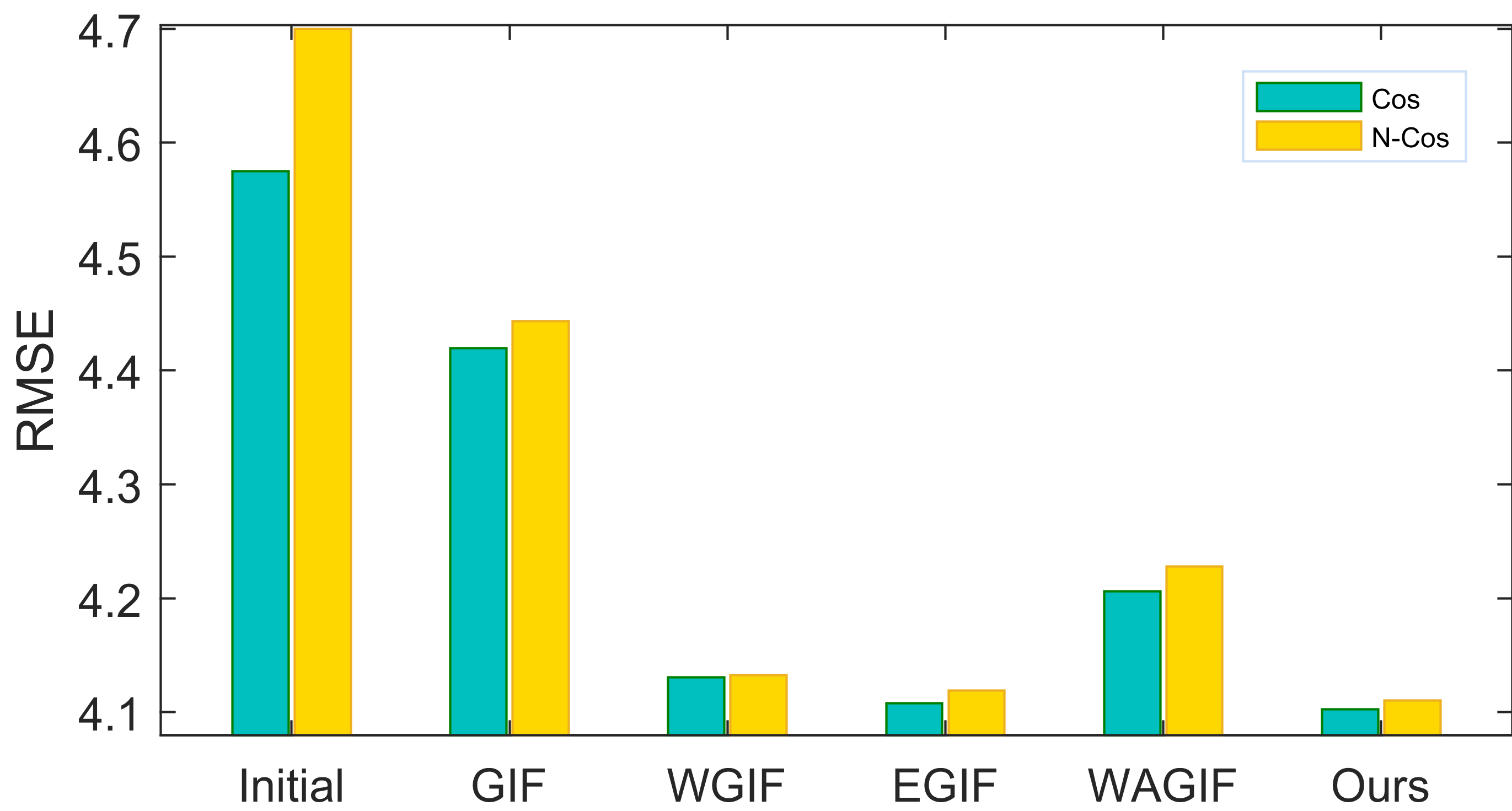}
		\end{minipage}
	}
	\subfloat[Cos and N-Cos]{
		\begin{minipage}[b]{0.5\textwidth}
			\includegraphics[width=8cm,height=4.5cm]{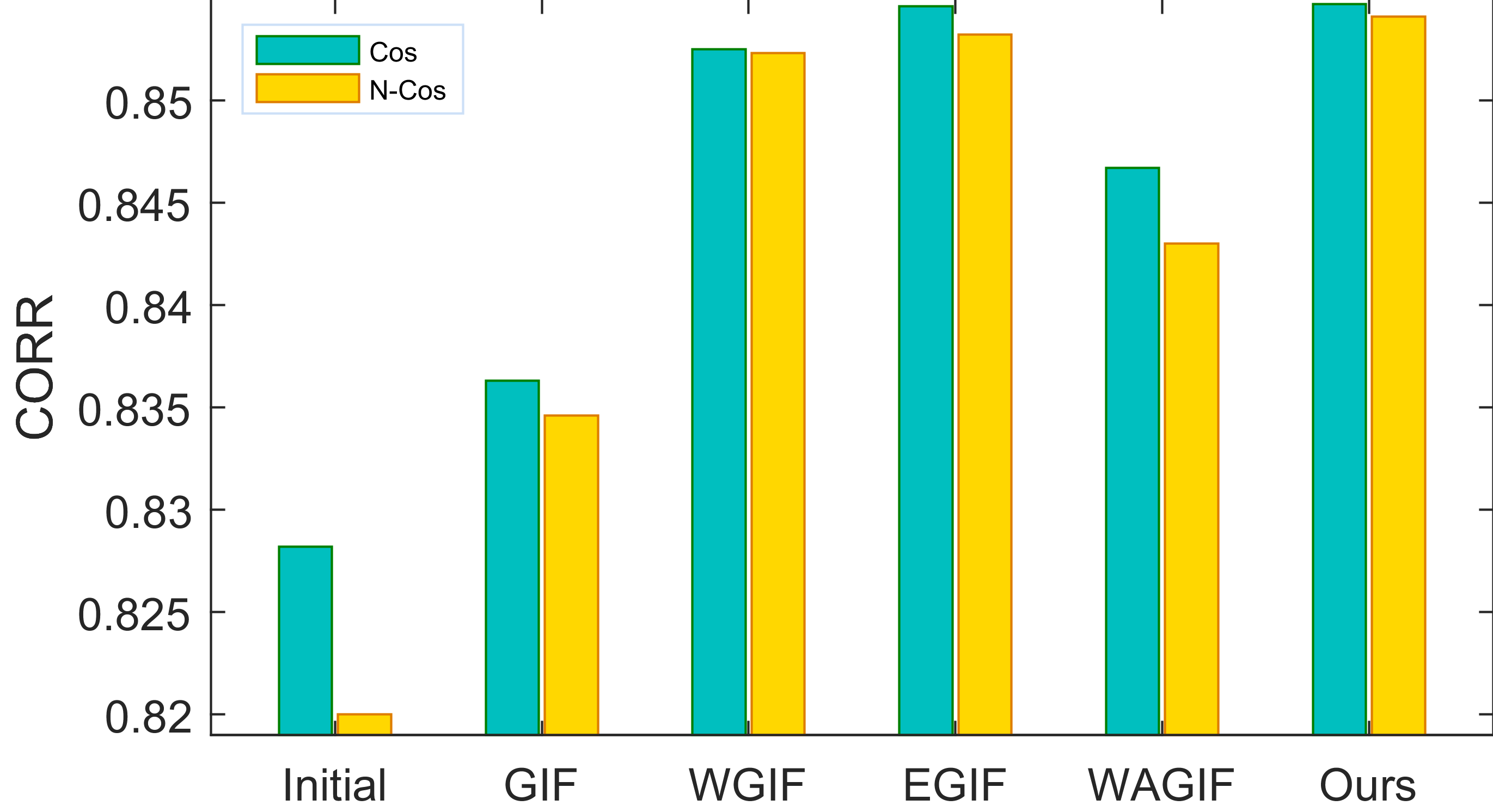}
		\end{minipage}
	}\hspace{1in}
	\subfloat[Sine and N-Sine]{
		\begin{minipage}[b]{0.5\textwidth}
			\includegraphics[width=8cm,height=4.5cm]{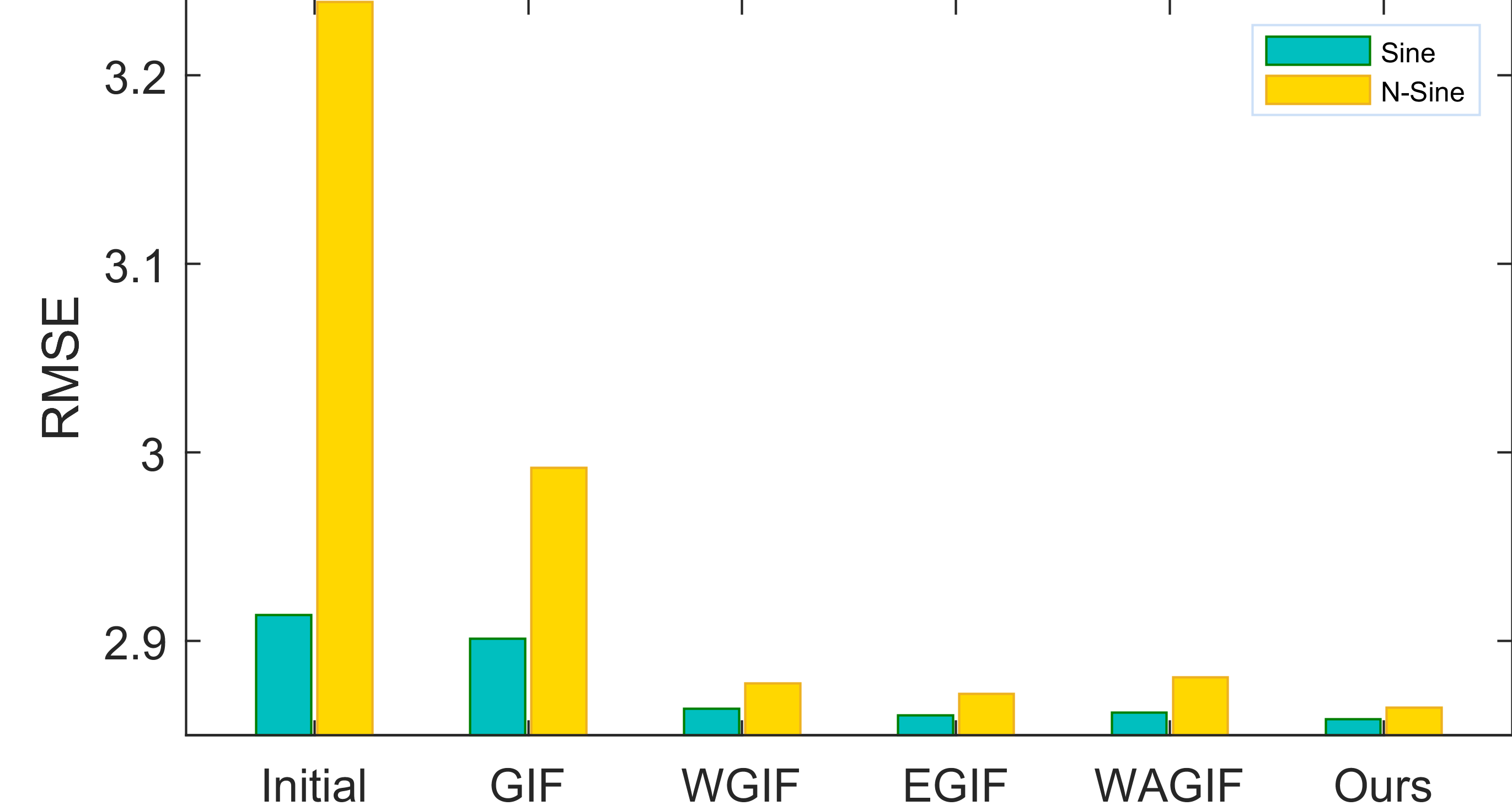}
		\end{minipage}
	}
	\subfloat[Sine and N-Sine]{
		\begin{minipage}[b]{0.5\textwidth}
			\includegraphics[width=8cm,height=4.5cm]{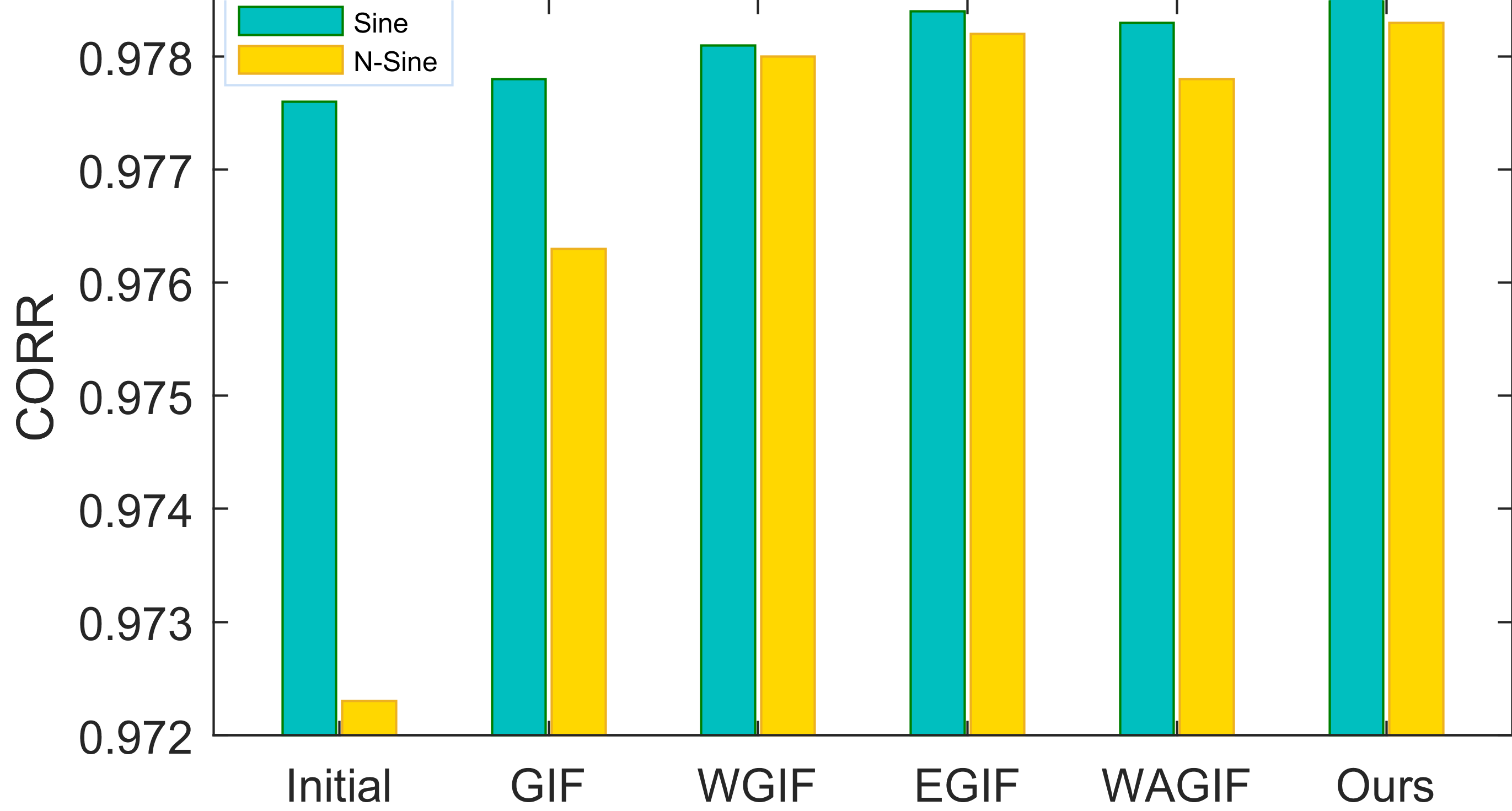}
		\end{minipage}
	}
	%	\vspace{0.1cm}
	\caption{Quantitative measures (RMSE and CORR) for the depth maps of clean and noisy synthetic image sequences.}
	\label{fig:QM_synthetic}
	\vspace*{-0.4cm}
\end{figure*}

\begin{flalign}
\label{CORR}
&\begin{array}{l}
CORR = \\
\displaystyle\frac{{\sum\limits_{u = 1}^U {\sum\limits_{v = 1}^V {\left( {{Z_f}(u,v) - {{\bar Z}_f}} \right)} } \left( {{Z_g}(u,v) - {{\bar Z}_g}} \right)}}{{\sqrt {\sum\limits_{u = 1}^U {\sum\limits_{v = 1}^V {{{\left( {{Z_f}(u,v) - {{\bar Z}_f}} \right)}^2}\sum\limits_{u = 1}^U {\sum\limits_{v = 1}^V {{{\left( {{Z_g}(u,v) - {{\bar Z}_g}} \right)}^2}} } } } } }}
\end{array},&
\end{flalign}
where $\bar{Z_f}$ represents the mean of the improved depth map, $\bar{Z_g}$ denotes the mean of the ground truth depth map. The correlation ranges from 0 to 1, and the closer the value is to 1, the better the quality of the improved depth map.
\begin{flalign}
\label{RMSD}
&RMSD = \sqrt {\frac{1}{{UV}}\sum\limits_{u = 1}^U {\sum\limits_{v = 1}^V {{{\left( {{Z_f}(u,v) - Z(u,v)} \right)}^2}} } },&
\end{flalign}
where ${Z_f}(u,v)$ represents the improved depth map, ${Z}(u,v)$ denotes the initial depth map. The value of the root mean square difference is greater, the depth enhancement is better.

In a word, smaller RMSE, larger CORR, and higher RMSD imply more improvement.

\begin{figure*}[htp]
	\centering
	%\vspace*{-0.5cm}
	\subfloat[GT]{
		\begin{minipage}[b]{0.138\textwidth}
			\begin{overpic}[width=2.3cm,height=1.8cm]{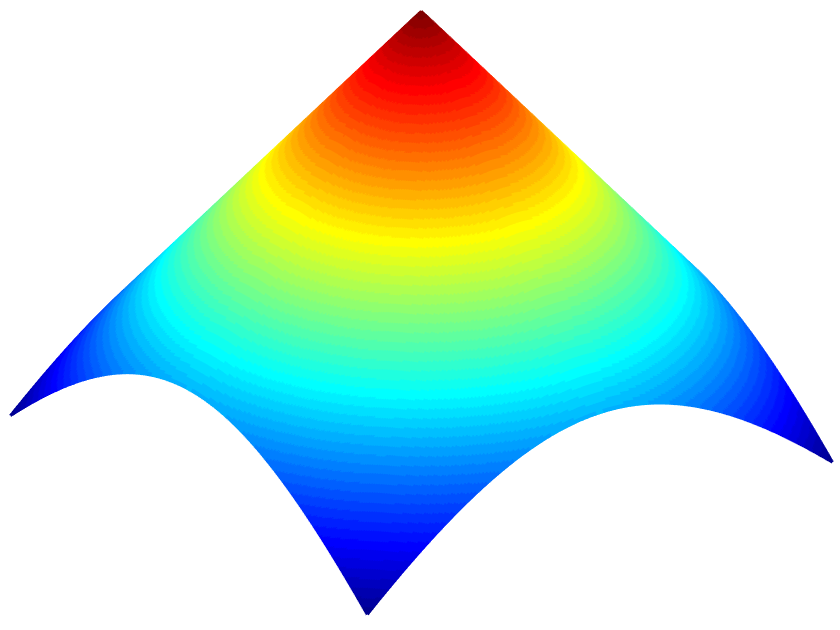}\vspace{.01in}
			\end{overpic}\vspace{.01in}
			
			\begin{overpic}[width=2.3cm,height=1.8cm]{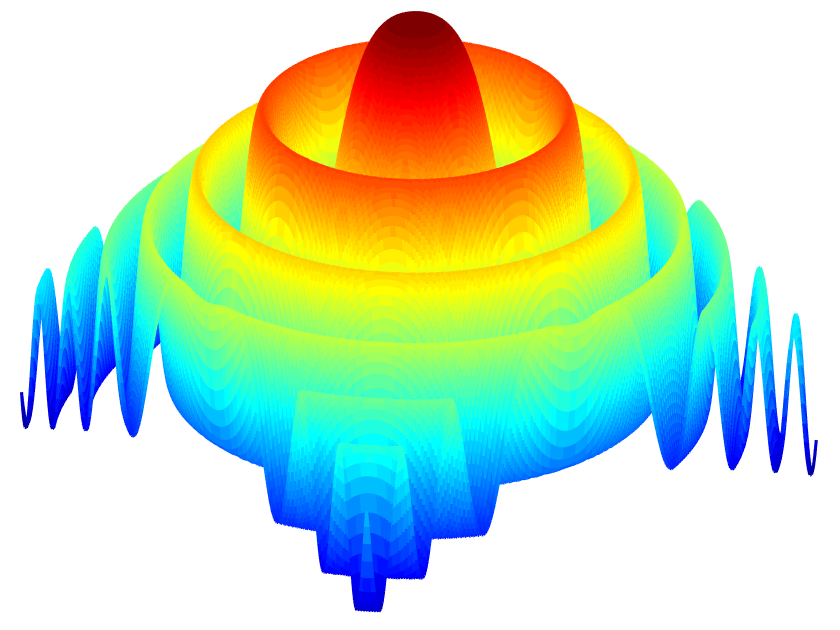}\vspace{.01in}
			\end{overpic}\vspace{.01in}
			
			\begin{overpic}[width=2.3cm,height=1.8cm]{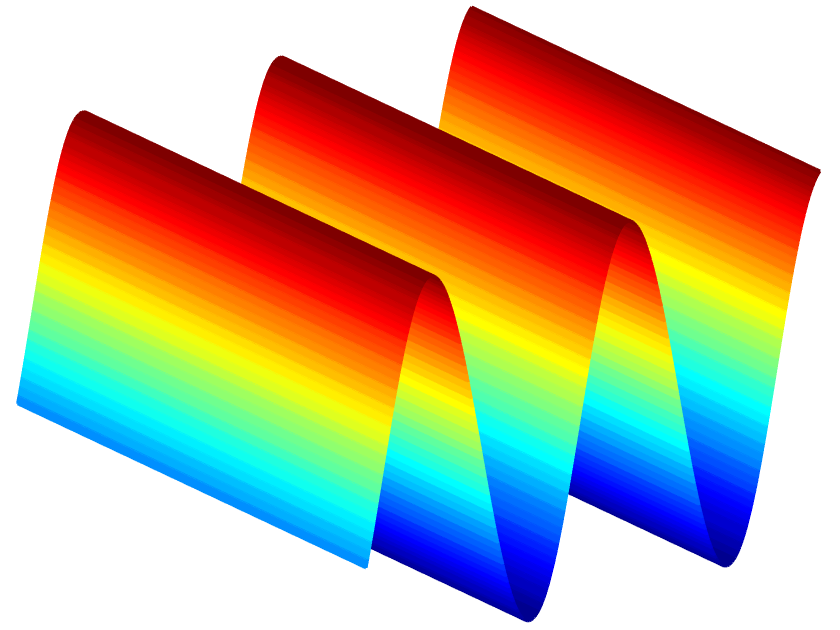}
			\end{overpic}\vspace{.01in}
		\end{minipage}
	}\hspace{-.1in}
	\subfloat[Initial]{
		\begin{minipage}[b]{0.138\textwidth}
			\begin{overpic}[width=2.4cm,height=1.8cm]{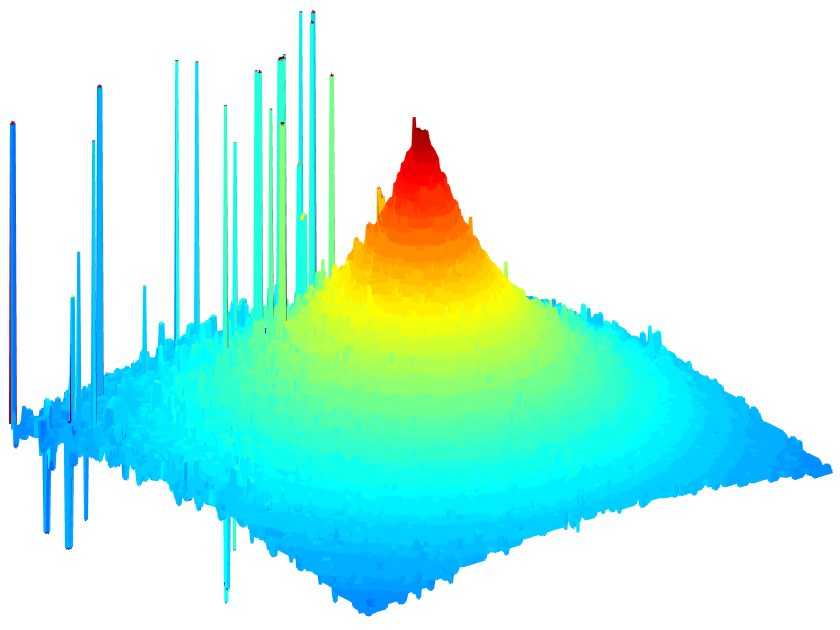}\vspace{.01in}
			\end{overpic}\vspace{.01in}
			
			\begin{overpic}[width=2.4cm,height=1.8cm]{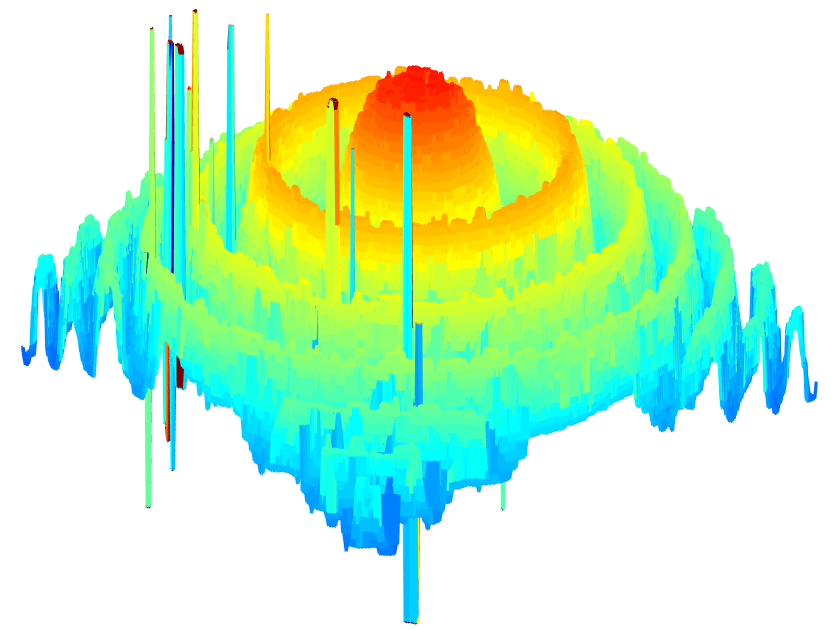}\vspace{.01in}
			\end{overpic}\vspace{.01in}
			
			\begin{overpic}[width=2.4cm,height=1.8cm]{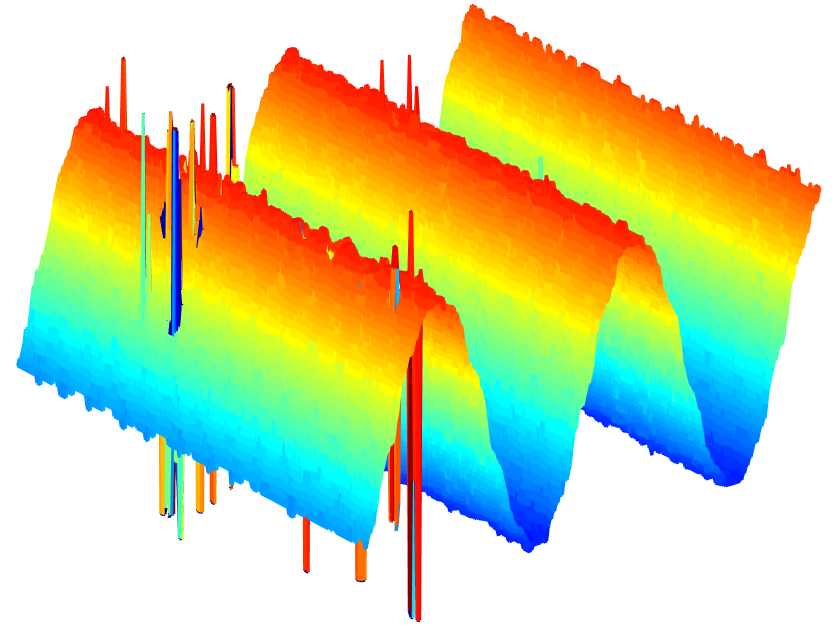}
			\end{overpic}\vspace{.01in}
		\end{minipage}
	} \hspace{-.1in}
	\subfloat[GIF]{
		\begin{minipage}[b]{0.138\textwidth}
			\begin{overpic}[width=2.4cm,height=1.8cm]{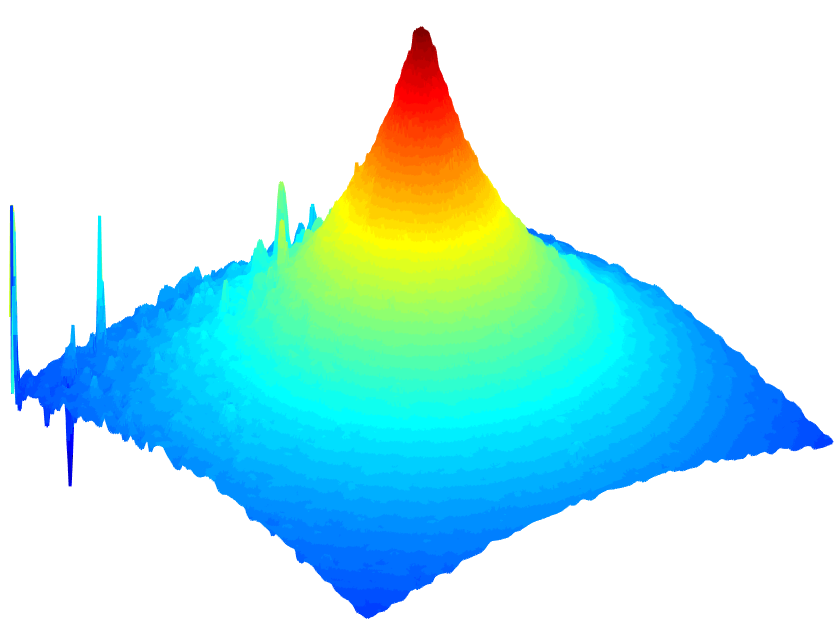}\vspace{.01in}
			\end{overpic}\vspace{.01in}
			
			\begin{overpic}[width=2.4cm,height=1.8cm]{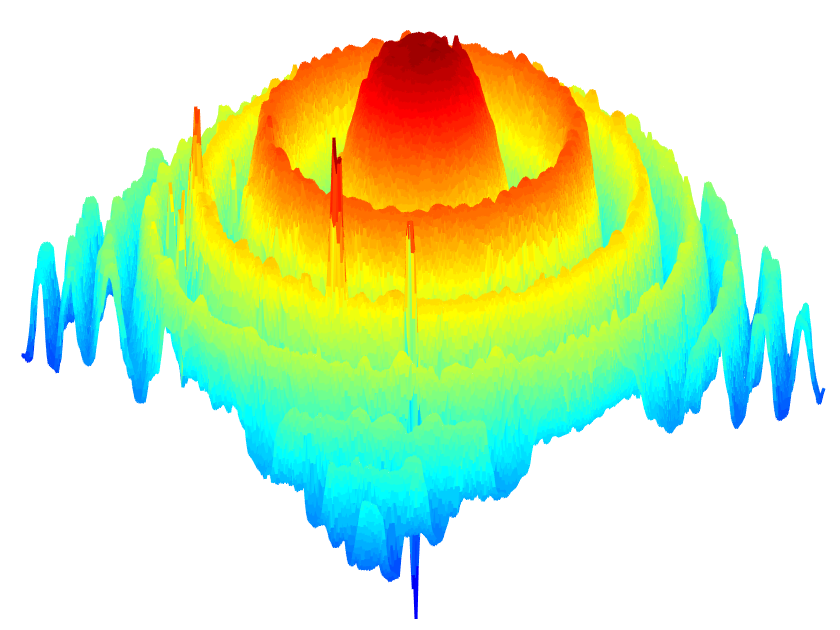}\vspace{.01in}
			\end{overpic}\vspace{.01in}
			
			\begin{overpic}[width=2.4cm,height=1.8cm]{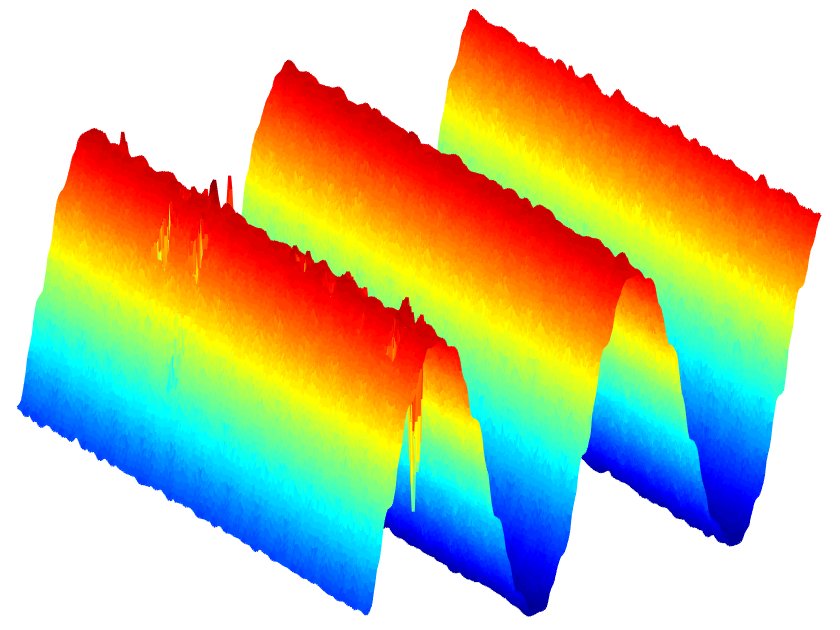}
			\end{overpic}\vspace{.01in}
		\end{minipage}
	}\hspace{-.1in}
	\subfloat[WGIF]{
		\begin{minipage}[b]{0.138\textwidth}
			\begin{overpic}[width=2.4cm,height=1.8cm]{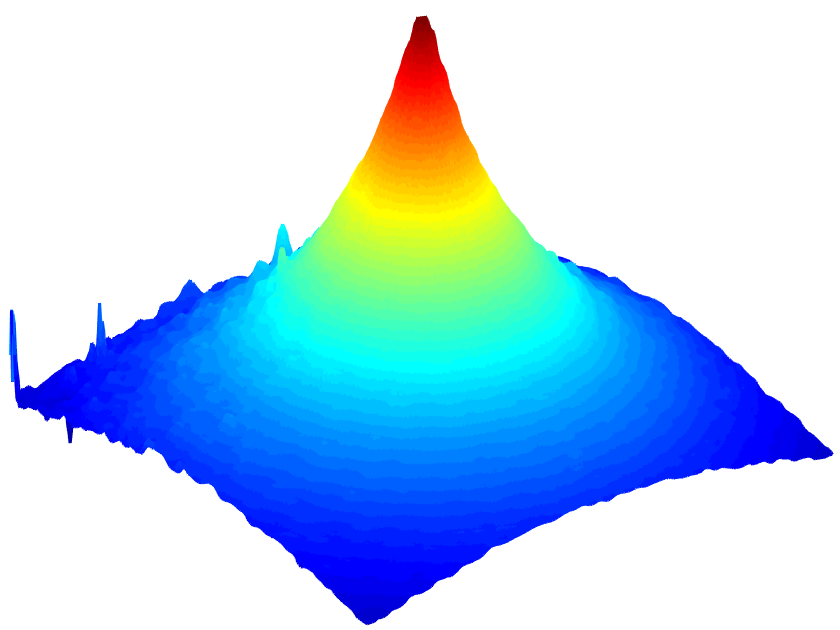}\vspace{.01in}
			\end{overpic}\vspace{.01in}
			
			\begin{overpic}[width=2.4cm,height=1.8cm]{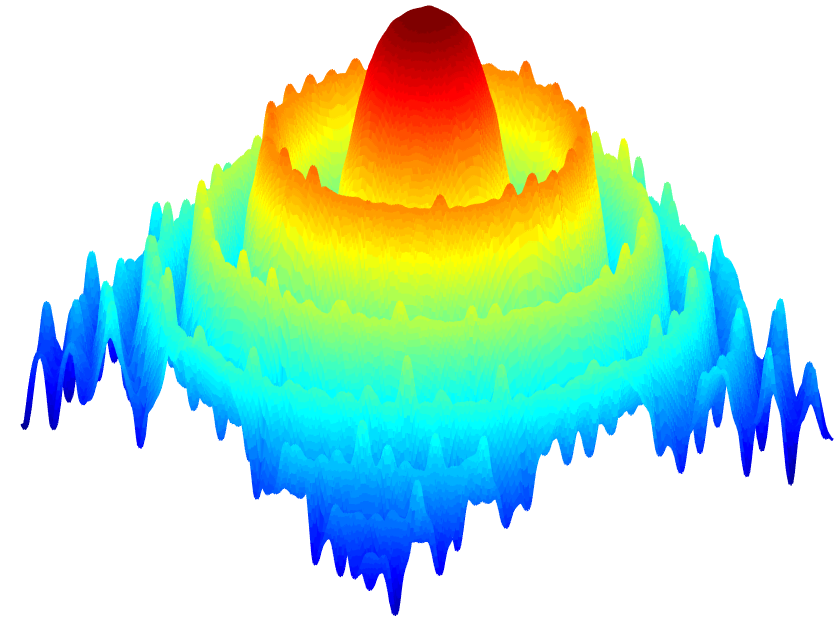}\vspace{.01in}
			\end{overpic}\vspace{.01in}
			
			\begin{overpic}[width=2.4cm,height=1.8cm]{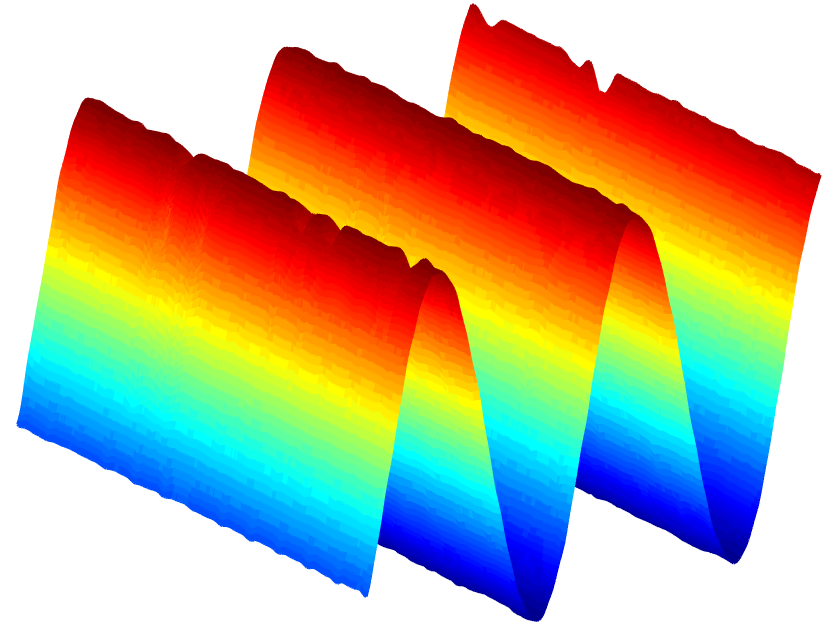}
			\end{overpic}\vspace{.01in}
		\end{minipage}
	}\hspace{-.1in}
	\subfloat[EGIF]{
		\begin{minipage}[b]{0.138\textwidth}
			\begin{overpic}[width=2.4cm,height=1.8cm]{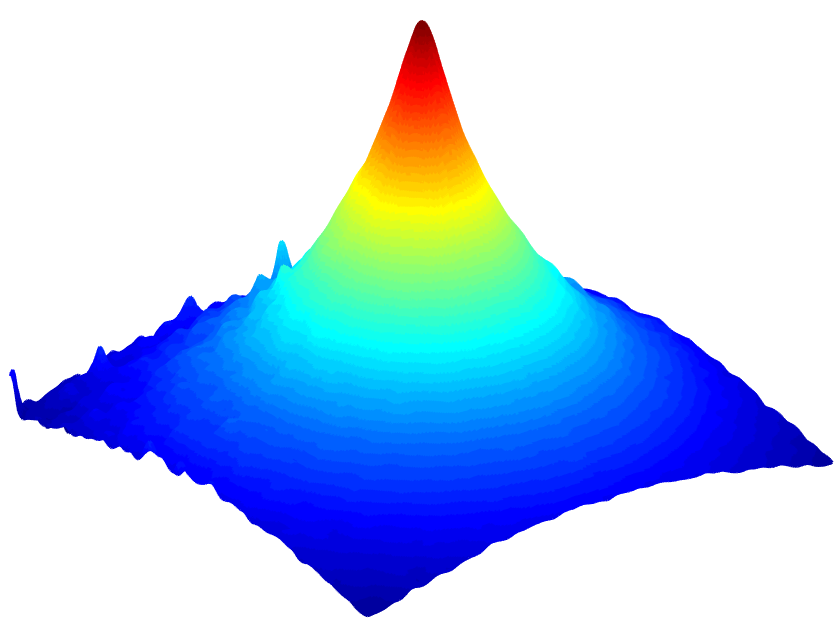}\vspace{.01in}
			\end{overpic}\vspace{.01in}
			
			\begin{overpic}[width=2.4cm,height=1.8cm]{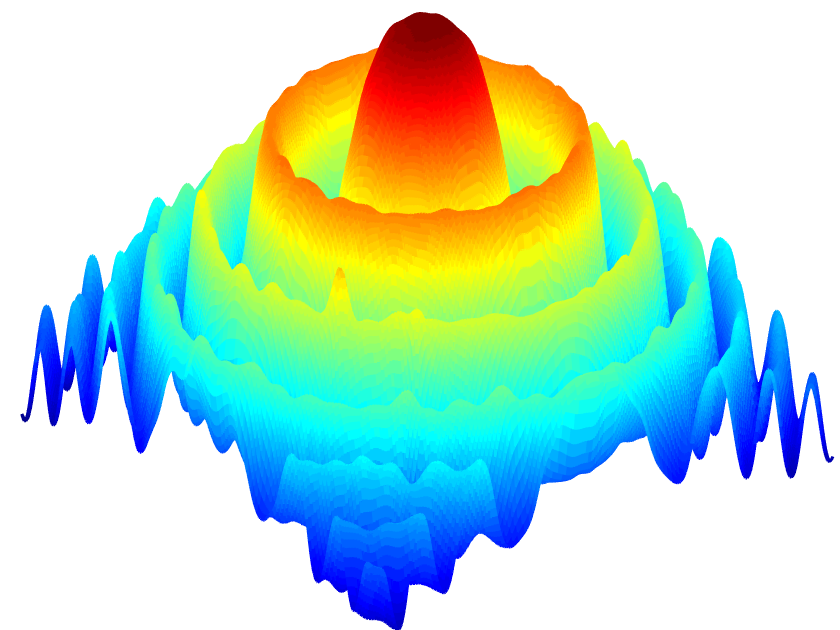}\vspace{.01in}
			\end{overpic}\vspace{.01in}
			
			\begin{overpic}[width=2.4cm,height=1.8cm]{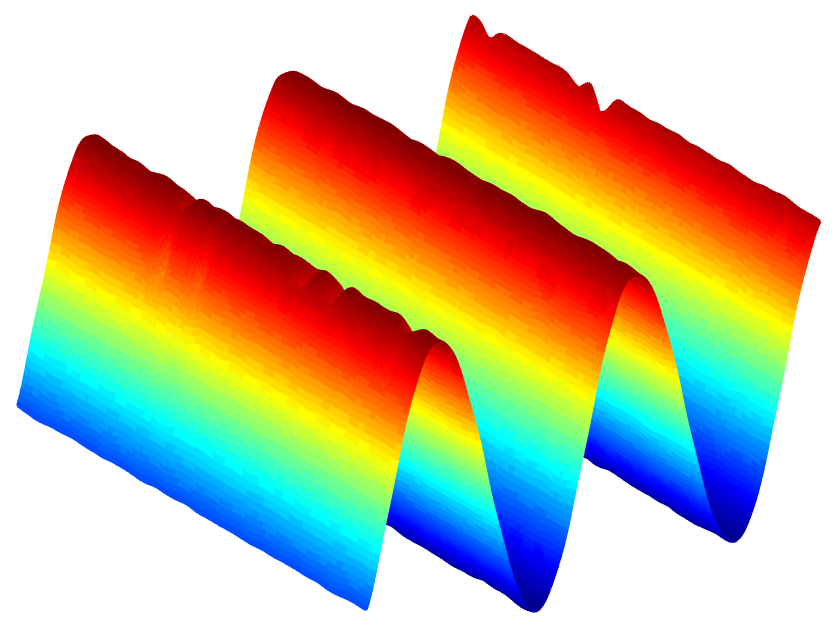}
			\end{overpic}\vspace{.01in}
		\end{minipage}
	}\hspace{-.1in}
	\subfloat[WAGIF]{
		\begin{minipage}[b]{0.138\textwidth}
			\begin{overpic}[width=2.4cm,height=1.8cm]{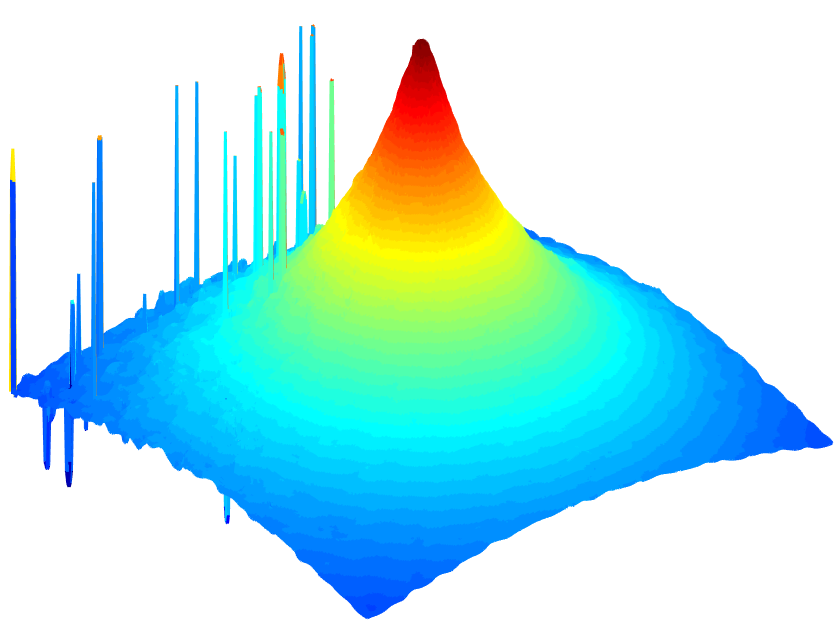}\vspace{.01in}	
			\end{overpic}\vspace{.01in}
			
			\begin{overpic}[width=2.4cm,height=1.8cm]{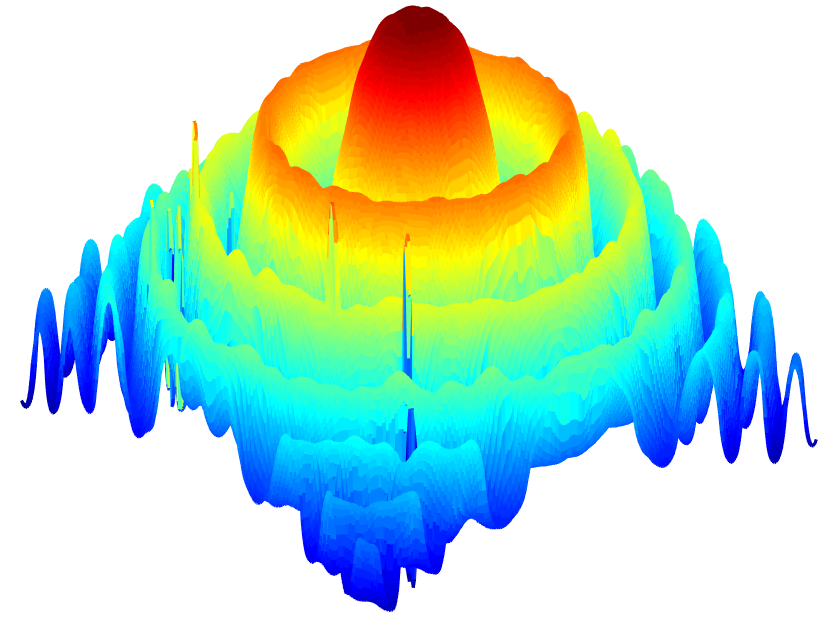}\vspace{.01in}
			\end{overpic}\vspace{.01in}
			
			\begin{overpic}[width=2.4cm,height=1.8cm]{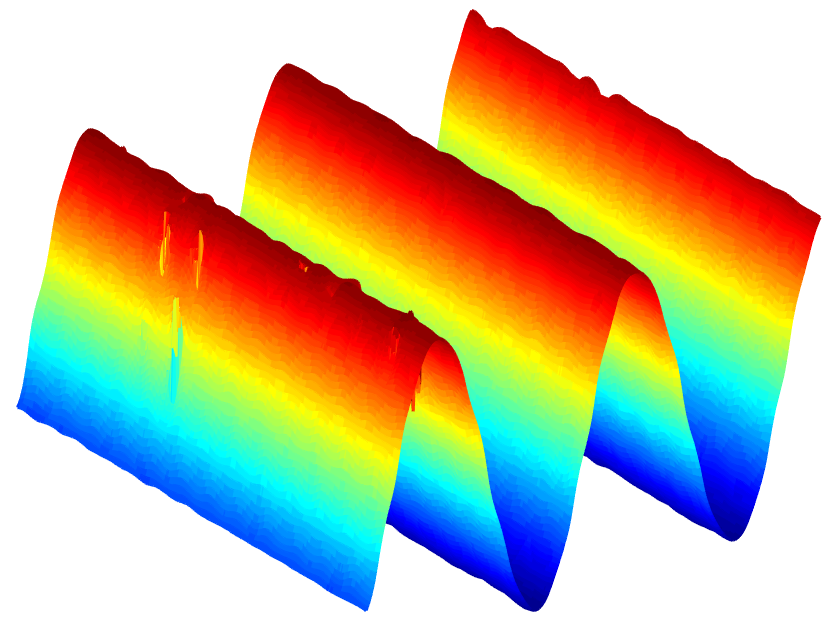}
			\end{overpic}\vspace{.01in}
		\end{minipage}
	}\hspace{-.1in}
	\subfloat[Ours]{
		\begin{minipage}[b]{0.138\textwidth}
			\begin{overpic}[width=2.4cm,height=1.8cm]{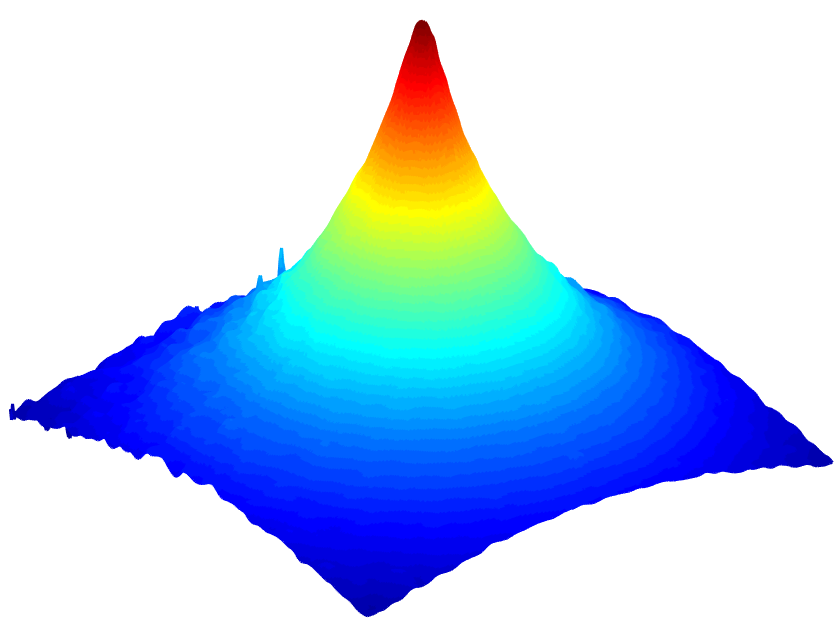}\vspace{.01in}	
			\end{overpic}\vspace{.01in}
			
			\begin{overpic}[width=2.4cm,height=1.8cm]{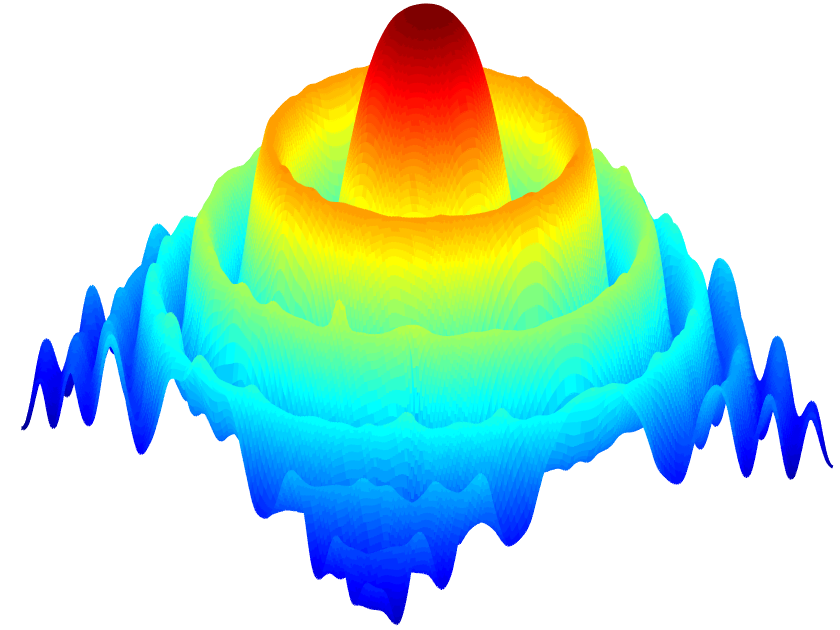}\vspace{.01in}
			\end{overpic}\vspace{.01in}
			
			\begin{overpic}[width=2.4cm,height=1.8cm]{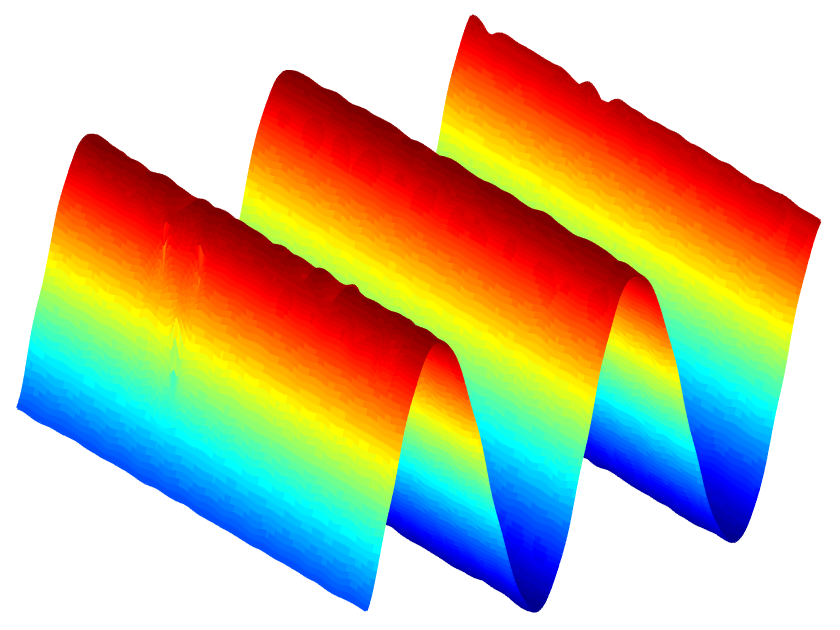}
			\end{overpic}\vspace{.01in}
		\end{minipage}
	}
	%\vspace{0.1cm}
	\caption{Depth maps of noisy synthetic image sequences. The N-Cone, N-Cos and N-Sine objects correspond from the first row to the third row respectively. (a) Ground truth images. (b) initial depth maps. The depth maps of N-Cone, N-Cos and N-Sine by (c) GIF, (d) WGIF, (e) EGIF, (f) WAGIF, and (g) Ours.}
	\label{fig:SyntheticNoise_Filter}
	\vspace*{-0.4cm}
\end{figure*}

\begin{figure*}[htp]
	\centering
	%\vspace*{-0.5cm}
	\subfloat[GD]{
		\begin{minipage}[b]{0.138\textwidth}
			\begin{overpic}[width=2.42cm,height=1.8cm]{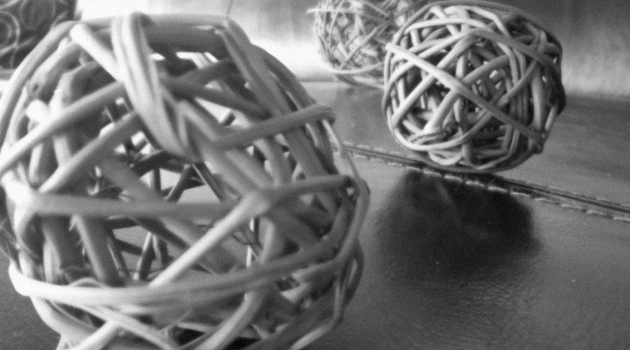}\vspace{.01in}
			\end{overpic}\vspace{.01in}
			
			\begin{overpic}[width=2.42cm,height=1.8cm]{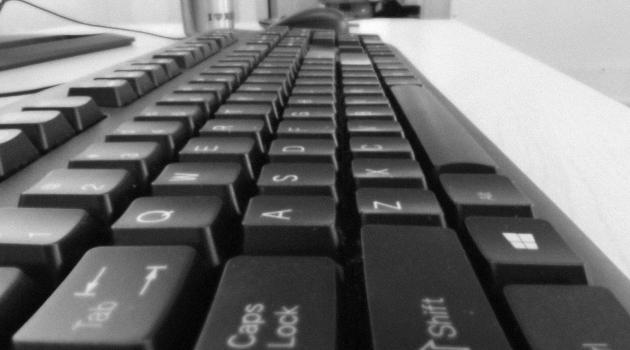}\vspace{.01in}
			\end{overpic}\vspace{.01in}
			
			\begin{overpic}[width=2.42cm,height=1.8cm]{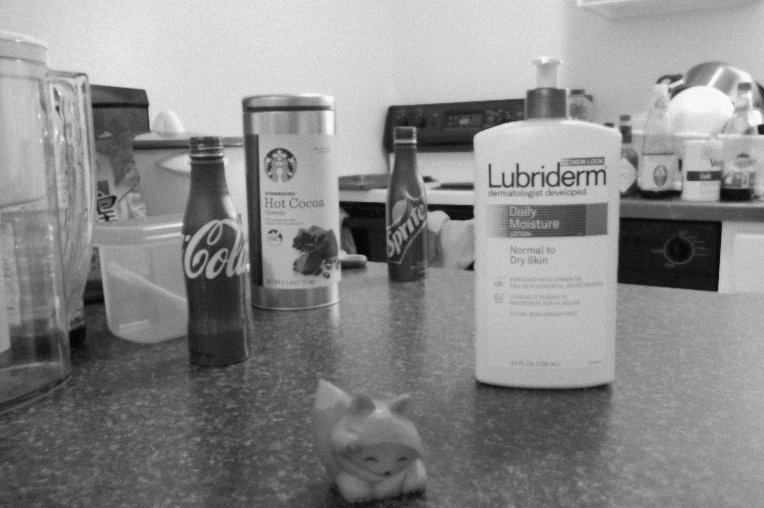}
			\end{overpic}\vspace{.01in}
		\end{minipage}
	}\hspace{-.1in}
	\subfloat[Initial]{
		\begin{minipage}[b]{0.138\textwidth}
			\begin{overpic}[width=2.42cm,height=1.8cm]{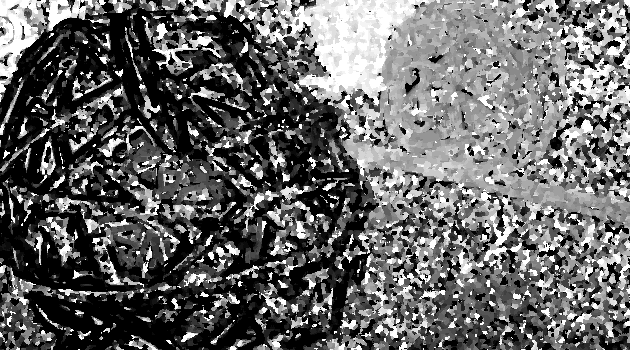}\vspace{.01in}
			\end{overpic}\vspace{.01in}
			
			\begin{overpic}[width=2.42cm,height=1.8cm]{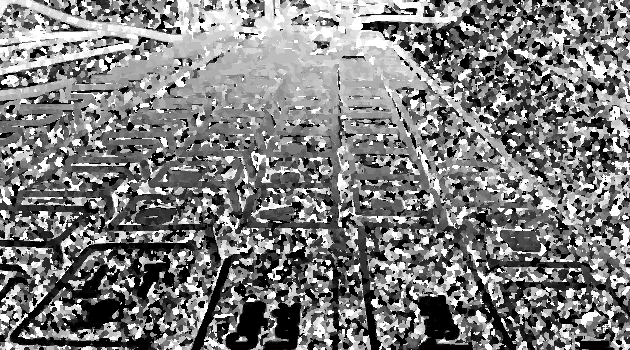}\vspace{.01in}
			\end{overpic}\vspace{.01in}
			
			\begin{overpic}[width=2.42cm,height=1.8cm]{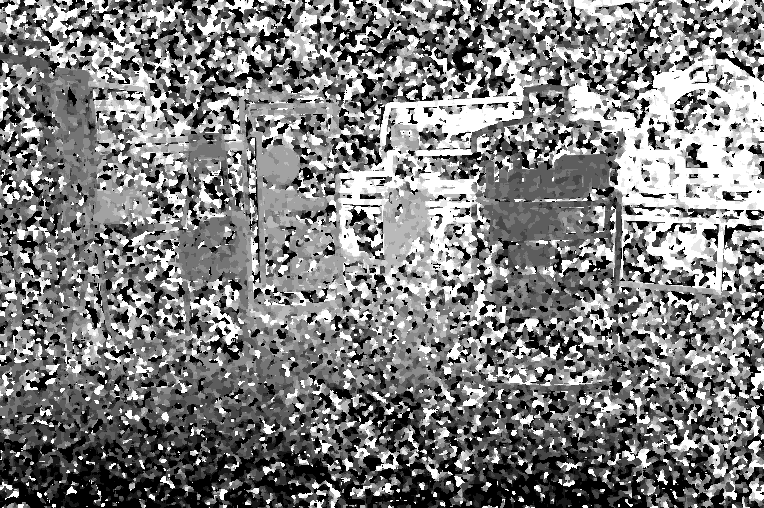}
			\end{overpic}\vspace{.01in}
		\end{minipage}
	} \hspace{-.1in}
	\subfloat[GIF]{
		\begin{minipage}[b]{0.138\textwidth}
			\begin{overpic}[width=2.42cm,height=1.8cm]{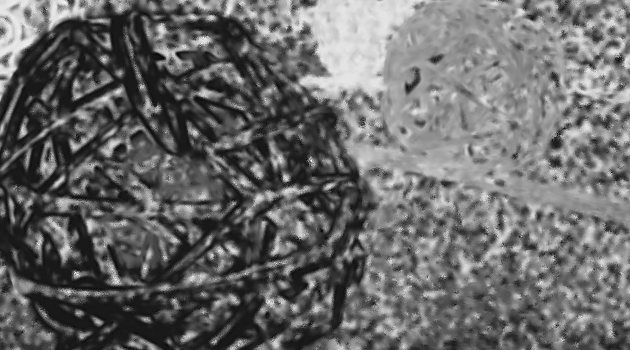}\vspace{.01in}
			\end{overpic}\vspace{.01in}
			
			\begin{overpic}[width=2.42cm,height=1.8cm]{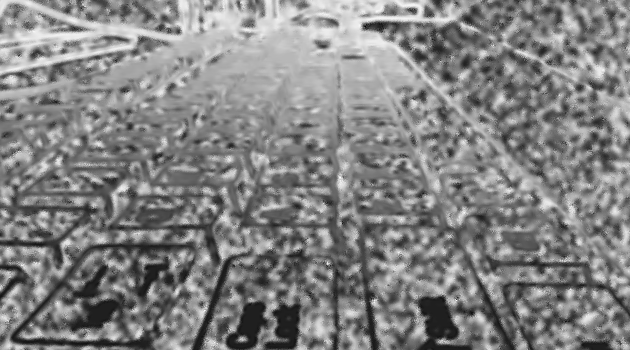}\vspace{.01in}
			\end{overpic}\vspace{.01in}
			
			\begin{overpic}[width=2.42cm,height=1.8cm]{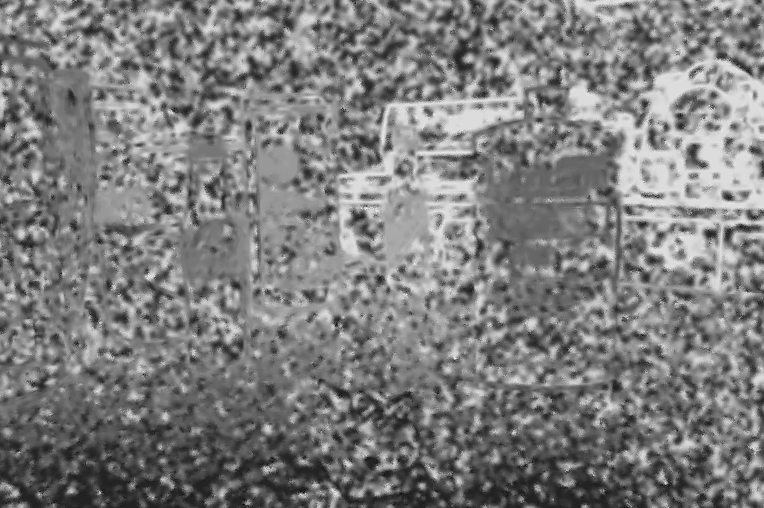}
			\end{overpic}\vspace{.01in}
		\end{minipage}
	}\hspace{-.1in}
	\subfloat[WGIF]{
		\begin{minipage}[b]{0.138\textwidth}
			\begin{overpic}[width=2.42cm,height=1.8cm]{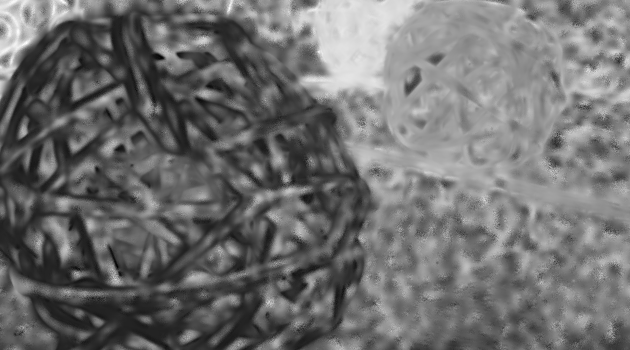}\vspace{.01in}
			\end{overpic}\vspace{.01in}
			
			\begin{overpic}[width=2.42cm,height=1.8cm]{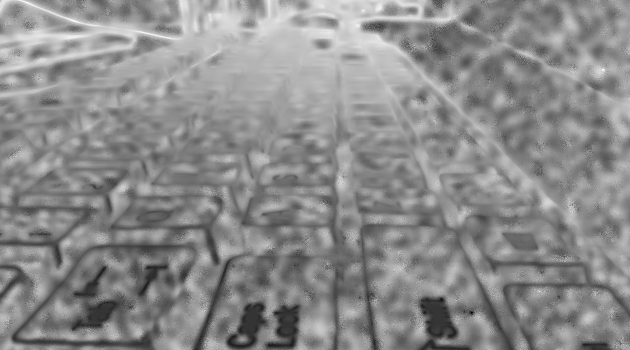}\vspace{.01in}
			\end{overpic}\vspace{.01in}
			
			\begin{overpic}[width=2.42cm,height=1.8cm]{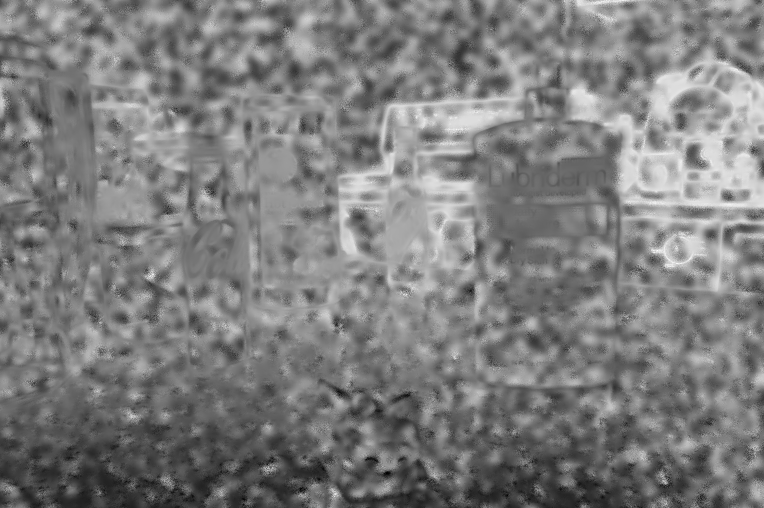}
			\end{overpic}\vspace{.01in}
		\end{minipage}
	}\hspace{-.1in}
	\subfloat[EGIF]{
		\begin{minipage}[b]{0.138\textwidth}
			\begin{overpic}[width=2.42cm,height=1.8cm]{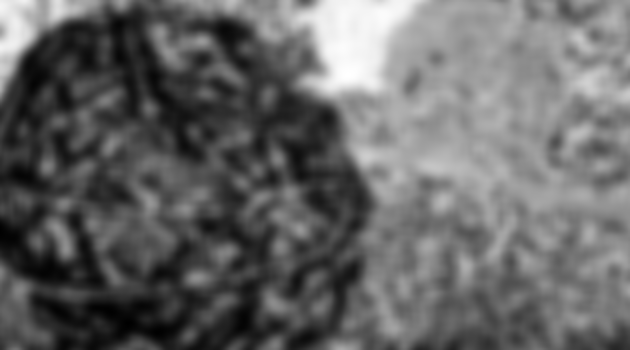}\vspace{.01in}
			\end{overpic}\vspace{.01in}
			
			\begin{overpic}[width=2.42cm,height=1.8cm]{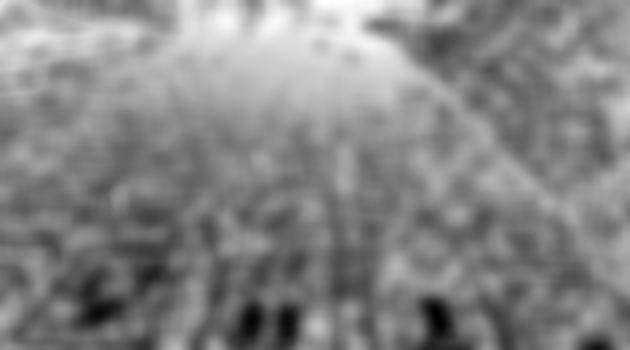}\vspace{.01in}
			\end{overpic}\vspace{.01in}
			
			\begin{overpic}[width=2.42cm,height=1.8cm]{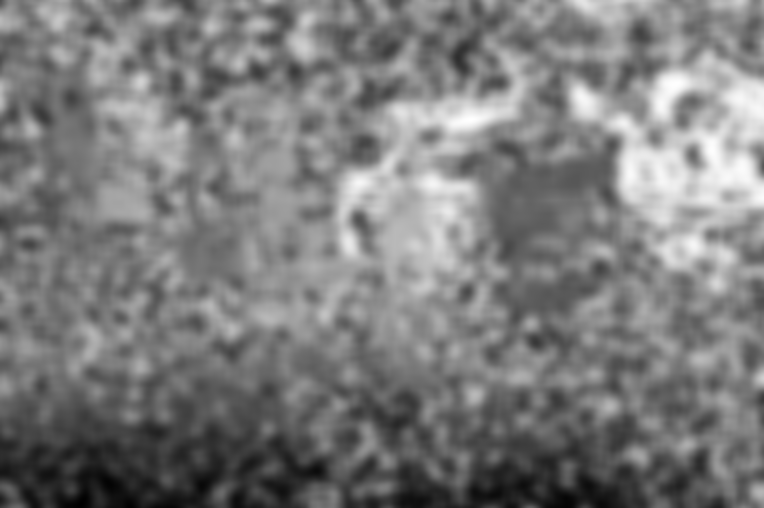}
			\end{overpic}\vspace{.01in}
		\end{minipage}
	}\hspace{-.1in}
	\subfloat[WAGIF]{
		\begin{minipage}[b]{0.138\textwidth}
			\begin{overpic}[width=2.42cm,height=1.8cm]{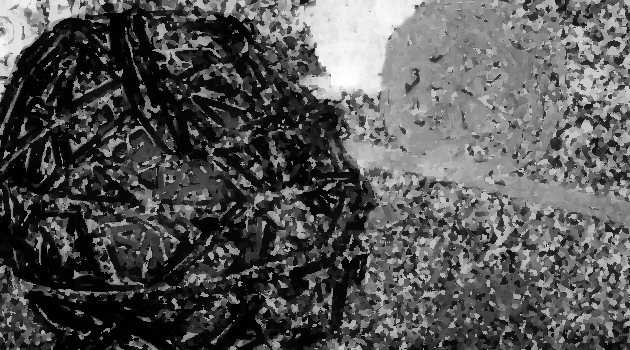}\vspace{.01in}	
			\end{overpic}\vspace{.01in}
			
			\begin{overpic}[width=2.42cm,height=1.8cm]{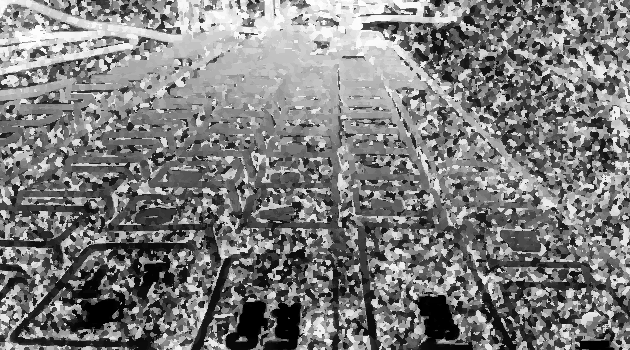}\vspace{.01in}
			\end{overpic}\vspace{.01in}
			
			\begin{overpic}[width=2.42cm,height=1.8cm]{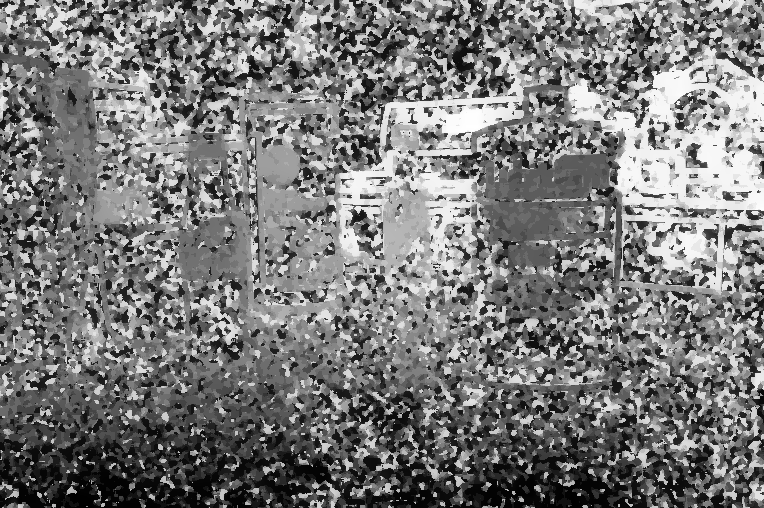}
			\end{overpic}\vspace{.01in}
		\end{minipage}
	}\hspace{-.1in}
	\subfloat[Ours]{
		\begin{minipage}[b]{0.138\textwidth}
			\begin{overpic}[width=2.42cm,height=1.8cm]{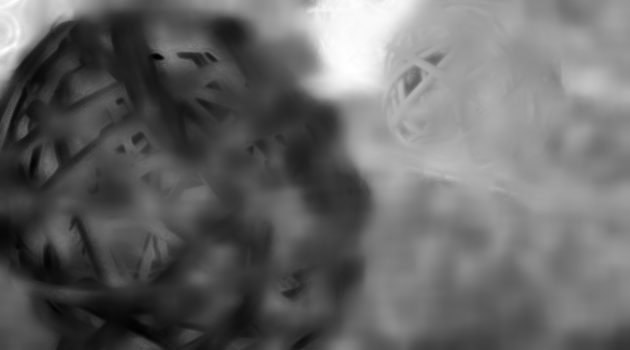}\vspace{.01in}	
			\end{overpic}\vspace{.01in}
			
			\begin{overpic}[width=2.42cm,height=1.8cm]{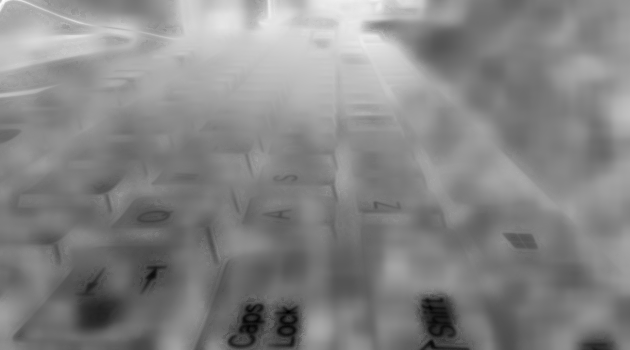}\vspace{.01in}
			\end{overpic}\vspace{.01in}
			
			\begin{overpic}[width=2.42cm,height=1.8cm]{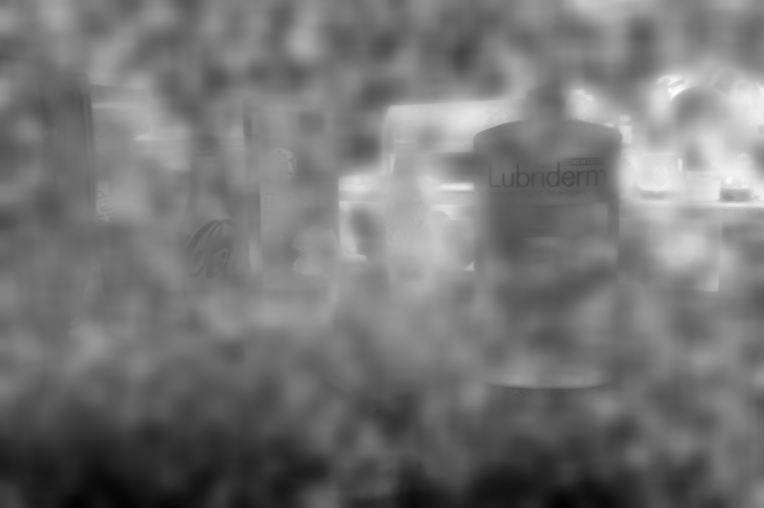}
			\end{overpic}\vspace{.01in}
		\end{minipage}
	}
	%\vspace{0.1cm}
	\caption{Depth maps of noisy real image sequences. The N-BA, N-KB and N-KC images correspond from the first row to the third row respectively. (a) The guidance maps. (b) Initial depth maps. The depth maps of N-BA, N-KB and N-KC by (c) GIF, (d) WGIF, (e) EGIF, (f) WAGIF, and (g) Ours.}
	\label{fig:RealNoise_Filter}
	\vspace*{-0.4cm}
\end{figure*}

\subsection{Effect of different adaptive amplification factor}\label{Effect}
In order to analyze the influence of the proposed adaptive amplification factor on the improved depth map, the value of $\beta$ is empirically selected to be 1/4,1/2, 3/4, 1, for experiments respectively. Thus, the optimal adaptive amplification factor can be selected by comparing the quality of the updated depth maps and used in all the subsequent experiments. Firstly, the proposed algorithm is updated with different adaptive amplification factors to obtain the relevant depth maps. Then, the quantitative measurement RMSE and RMSD are calculated for the resultant depth maps of synthesize images Cone, Sine, Cos, N-Cone, N-Sine, N-Cos and real images BA, KB, KC, N-BA, N-KB, N-KC, respectively. Finally, for each different adaptive amplification factor, the mean of RMSEs are computed separately for three clean synthesize images and three noise synthesize images, while the means of RMSDs are computed separately for three clean real images and three noise real images.

The experimental results are shown in Fig. \ref{fig:beta_selection}. In the bar figure, with the increase of $\beta$ value, the RMSEs of the synthetic images slowly decrease, while the RMSDs of the real images gradually increase, which indicates that the depth map of the synthetic image is closer to the ground truth, and the depth map of the real image is much better with the increase of $\beta$. However, when the $\beta$ value is too large in the synthetic images, the quality of the depth map is not improved and the noise will be amplified, that is displayed in Fig. \ref{fig:beta_noise}. In order to intuitively observe the effect of larger $\beta$ value on the quality of depth map, two data 5/4 and 3/2 with $\beta$ greater than 1 are added for the experiment. The experimental parameters are set as $\zeta = 3$, $\lambda_0 = 50$. By comparing the depth maps of N-cones with six different adaptive amplification coefficients ($\beta$ values are 1/4,1/2, 3/4, 1, 5/4, and 3/2), it can be known that the depth noise decreases gradually with the increase of $\beta$. When $\beta = 1$, the depth map is the best one. When $\beta$ is greater than 1, the noise increases in the opposite direction. Therefore, $\beta = 1$ is selected as the optimal parameter in this paper to preserve details while better suppressing noise and increasing the robustness of the algorithm.

\begin{figure*}[htp]
	\centering
	%\vspace*{-0.5cm}
	\subfloat[RealClean]{
		\begin{minipage}[b]{0.5\textwidth}
			\includegraphics[width=8.0cm,height=4.5cm]{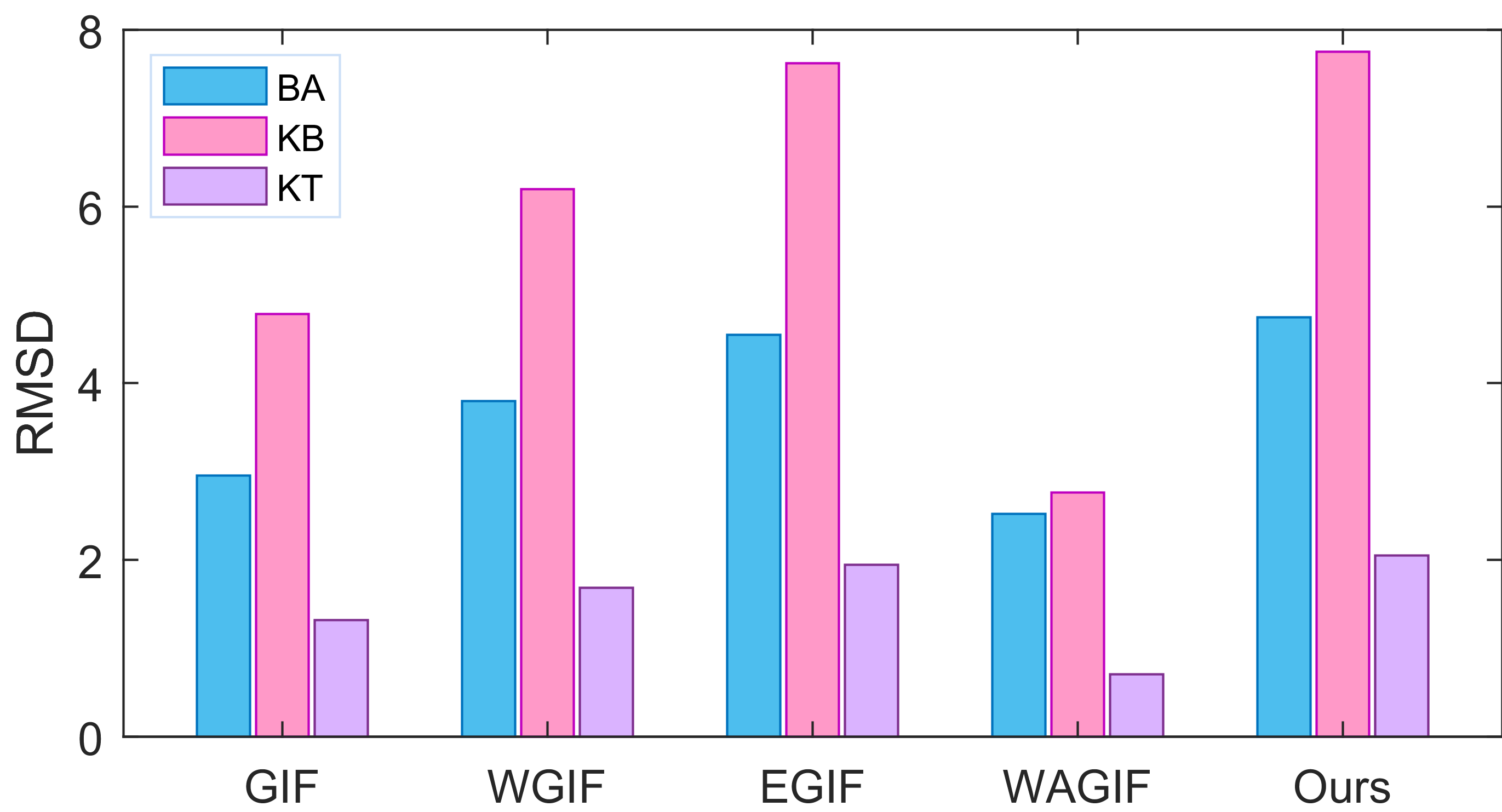}
		\end{minipage}
	}
	\subfloat[RealNoise]{
		\begin{minipage}[b]{0.5\textwidth}
			\includegraphics[width=8cm,height=4.5cm]{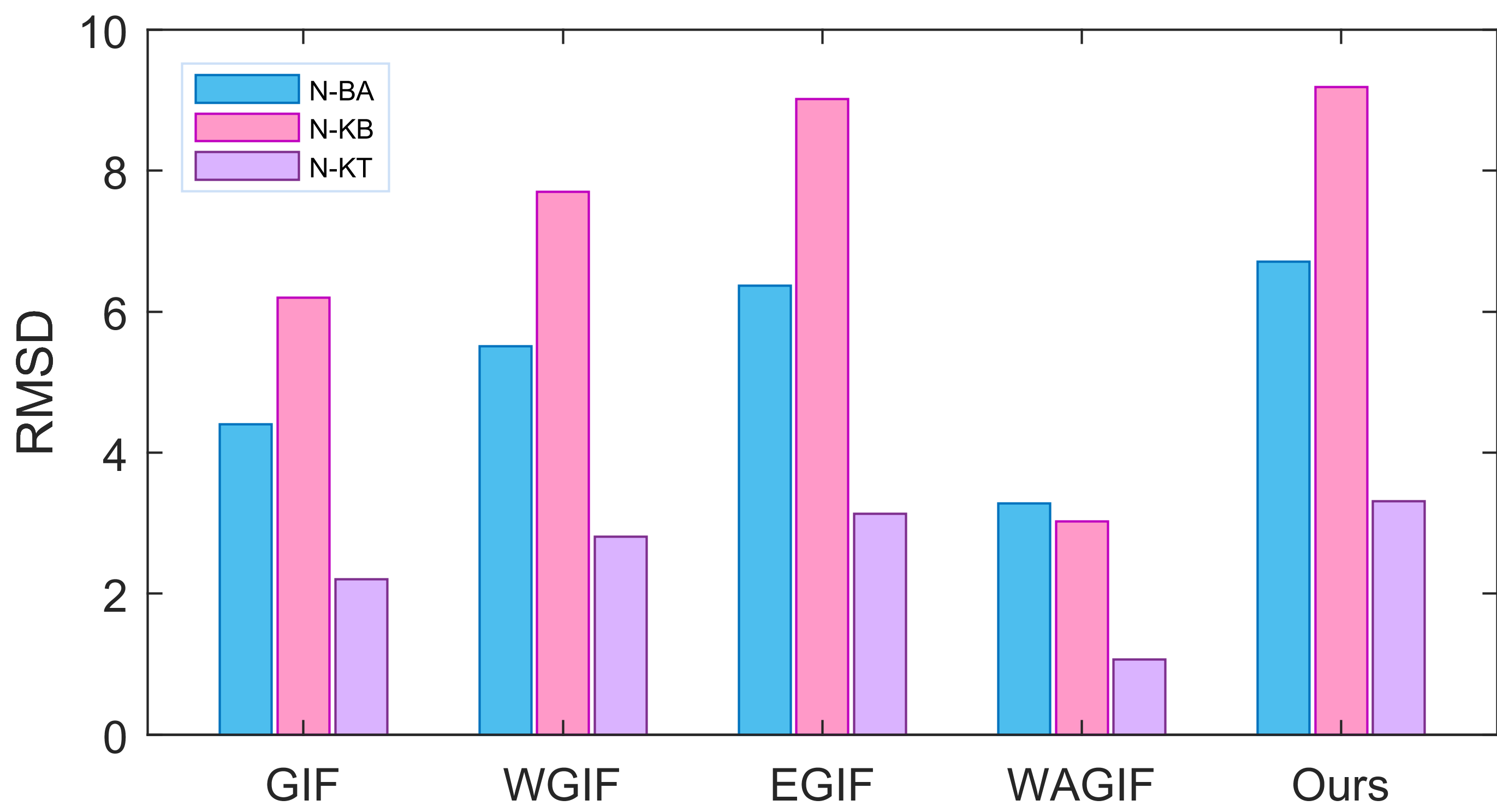}
		\end{minipage}
	}
	\caption{Quantitative measures (RMSD) for the depth maps of clean and noisy real image sequences.}
	\label{fig:QM_real}
	\vspace*{-0.4cm}
\end{figure*}

\subsection{Comparative analysis}\label{Comparative}
\subsubsection{Qualitative analysis}\label{Qualitative}
Firstly, the quality of depth maps for synthetic images obtained by different guided filtering are compared, and the performance of the proposed AWGIF algorithm is analyzed qualitatively. The experimental results have been shown in Fig. \ref{fig:SyntheticClean_Filter}, depth maps of cone, cos and sine are displayed from the first row to third row respectively. The first column includes the ground truth (GT) depth maps, the second column is the initial depth maps which suffer from severe distortions. It can be observed that the initial depth map can be moderately improved by the GIFs (GIF, WGIF, EGIF and WAGIF) algorithms, but there is still a certain degree of edge distortion, the surface is not enough smooth, and the detail structure is not maintained well. For the depth maps of cone object, the parameters of the guided filter are set as $\zeta = 2$, $\lambda_0 = 100$ better comparative analysis. The experimental results show that the existing GIFs algorithms still has a rough edge at the top of the depth map, and the proposed AWGIF algorithm smooths the edge better. For the depth maps of cos object, $\zeta = 5$, $\lambda_0 = 700$ are set for the guided filter. Due to the complex geometry and high contrast texture of the cos image sequences, there is a false prediction of depth discontinuity. Our proposed algorithm solves this problem better than the other GIFs because of its adaptive properties. For the depth maps of sine object, set as $\zeta = 5$, $\lambda_0 = 3000$ for the guided filter. The results of the experiment clearly show that our proposed AWGIF algorithm does a better job of removing distortion, preserving edges, smoothing surfaces, and enhancing structural details.

Then, the performance of each guide filters is qualitatively analyzed by comparing the depth maps of real images. The parameters are set as $\zeta = 11$, $\lambda_0 = 400$ for the guided filter. A similar observation can be made in Fig. \ref{fig:RealClean_Filter}. The first row to the third row display depth maps of BA, KB and KC. The first column displays the guidance maps which contain structural information about the real scene. The second column is the initial depth maps which are obtained by the traditional SFF techniques. The quality of the depth maps is evaluated by distinguishing near and far end points. The depth map gradually becomes lighter as the distance increases. However, the initial depth maps do not smooth gradually from near end points to far end points and there are obvious noise. The experimental results show that GIFs algorithms improve the initial depth map to some extent, but the GIF and WAGIF lack in providing substantial improvement, the WGIF improves a little better, the EGIF makes the depth map more obscure, the AWGIF has clear distinction between near and far end points. Although there are some errors, the proposed AWGIF algorithm provides better improvements than the other guided filters.

\subsubsection{Quantitative analysis}\label{Quantitative}
To analyze the performance of the proposed algorithm objectively, quantitative measures RMSE and CORR for the synthetic image sequences have been shown in Fig. \ref{fig:QM_synthetic}. The green bars in Fig. \ref{fig:QM_synthetic} (a) and (b) display the RMSE and CORR values of depth maps for Cone images which are obtained by the different filters. It can be observed that the RMSE of all the GIFs algorithms are particularly close. They differ by 0.0001 in RMSE values. Readers are invited to enlarge the full-size image of the electronic version to better compare the differences of GIFs. Nevertheless, the RMSE value of the proposed AWGIF algorithm is minimal. While the maximum CORR value is by the AWGIF and EGIF. Combined with qualitative analysis, it can be concluded that the AWGIF has the best improvement on the depth map of cone. The RMSE and CORR values of the Cos depth maps have been demonstrated in the green bars of Fig. \ref{fig:QM_synthetic} (c) and (d). Obviously, there is a great improvement on the initial depth map by all the GIFs, especially the WGIF, EGIF and AWGIF. Comparatively speaking, the RMSE values are lower and the CORR values are higher of these three filters. In particularly, our proposed AWGIF algorithm has the best RMSE and CORR values, and provides the top-quality depth map for the Cos images. For the Sine depth maps, the RMSE and CORR have been exhibited in the green bar of the Fig. \ref{fig:QM_synthetic} (e) and (f). The smaller RMSE and larger CORR values have been obtained by all the GIFs. The WGIF, EGIF, AWGIF and WAGIF provide a greater improvement on the depth map, and their improved depth maps are very close to the ground truth image. Among them, the proposed WAGIF algorithm has the least RMSE and maximum CORR. Therefore, the WAGIF has the excellent performance in improving the sine depth maps.

\begin{table*}[]
	\centering
	%\vspace*{-0.2cm}
	\caption{Comparison of computational time in seconds for various guided filters}
	%\vspace*{-0.2cm}
	\label{ComputationalTime}
	\begin{tabular}{p{.8cm}<{\raggedleft}|ccc|ccc|ccc|ccc}
		\hline
		GIFs & Cone  & Cos  & Sine   & N-Cone  & N-Cos  & N-Sine & BA  & KB  & KC  & N-BA  & N-KB  & N-KC  \\ \hline
		GIF & 0.0182  & 0.0099  & 0.0096  & 0.0175  & 0.009 & 0.0079  & 0.0252  & 0.0274  & 0.0439  & 0.0258  & 0.0262 & 0.0446 \\
		WGIF & 0.0340  & 0.0185  & 0.0183  & 0.0344  & 0.0180 & 0.0185 & 0.0899  & 0.0934  & 0.1524  & 0.0888  & 0.0949 & 0.1515 \\
		WAGIF & 0.0242  & 0.0118  & 0.0117  & 0.0210  & 0.0123 & 0.0121 & 0.0627  & 0.0658  & 0.1019  & 0.0622  & 0.0624 & 0.1062 \\
		EGIF & 0.6428  & 0.1647  & 0.2599  & 0.7121  & 0.1699 & 0.2625 & 2.1077  & 1.9012  & 2.1772  & 2.1183  & 1.9932  & 2.1727  \\
		Ours & 0.0435  & 0.0277  & 0.0275  & 0.0420  & 0.0278 & 0.0283 & 0.1158  & 0.1010  & 0.1965  & 0.1180  & 0.1033 & 0.1984 \\ \hline
	\end{tabular}
	\vspace*{-0.4cm}
\end{table*}

For the depth maps of the real image sequences, the RMSD has been displayed in Fig. \ref{fig:QM_real}. Since there are no the ground truth depth maps available for the real images, it is difficult to quantitatively evaluate the performance of the guided filters for the real images. According to literature \cite{ali2021guided}, calculating RMSD between the initial depth maps and the improved depth maps can perform quantitative analysis of experimental results to a certain extent. The value of RMSD is larger, the guided filters offer greater improvements for the depth maps. However, this RMSD value is only for reference and is not 100$\%$ accurate. For the depth maps of the real image sequences, the RMSD has been displayed in Fig. \ref{fig:QM_real} (a). The blue, pink, and purple bars correspond to the RMSD values of BA, KB, and KT images respectively. As can be seen from these bars, the RMSD value of the proposed AWGIF algorithm is the largest, and the RMSD values of EGIF, WGIF, GIF, and WAGIF decreases successively. This means that the AWGIF produces the best quality depth maps, followed by the EGIF, WGIF, GIF, and WAGIF. This is consistent with the results of qualitative analysis.

\subsection{Robustness analysis}\label{Robustness}

\subsubsection{Qualitative analysis}\label{QualitativeRo}
Depth maps of noisy synthetic image sequences have been shown in Fig. \ref{fig:SyntheticNoise_Filter}. From the first row to the third row, the depth maps of N-Cone, N-Cos and N-sine are corresponding respectively. The first column displays the ground truth images. The initial depth maps in the second column have large noisy spikes and excessively deteriorated. These filters have improved the noise depth maps, but their robustness is still limited. For the N-cone depth maps, the experimental parameters are set as $\zeta = 3$, $\lambda_0 = 50$. It can be seen there is almost no anti-noise ability by the WAGIF. While the GIF roughly removes some of the salient noise. The WGIF, EGIF and the proposed WAGIF algorithm exhibit anti-noise ability better, especially the WAGIF. Although there are still some small noises at the edges, the AWGIF is more robust than other filters. For the N-Cos image sequences, set the parameters as $\zeta = 5$, $\lambda_0 = 700$. The GIF and WAGIF still result into noisy depth maps obviously. The WGIF, EGIF and the proposed AWGIF algorithm provide the much improved depth maps, but the depth edge of the WGIF and EGIF are not as smooth as the WAGIF. The WAGIF has better performance for robustness. For the N-Sine depth maps, $\zeta = 4$, $\lambda_0 = 3000$ are set for the experiment. The good anti-noise ability has been shown in all the filters, but the GIF and AWGIF depth maps still have the visible noise. The WGIF and EGIF are inferior to the proposed AWGIF in terms of edge-preserving and surface smoothing. In a word, the AWGIF has demonstrated greater robustness and succeeded in recovering the depth maps more accurately. 

Depth maps of noisy real image sequences have been shown in Figs. \ref{fig:RealNoise_Filter}. The first row to the third row correspond to the results generated by N-BA, N-KB and N-KT image sequences respectively. The first column is the guided images, and the second column is the initial depth maps which are full of highly noise. The parameters are set as $\zeta = 11$, $\lambda_0 = 400$. It can be seen from the experimental results that the GIF and WAGIF improve the noise depth map very little. The WGIF, EGIF and the proposed AWGIF improve the noise problem of the initial depth map to some extent. However, the distinction between near and far end points in the depth maps of the AWGIF are clearer than the other filters, so the AWGIF is best for improving the noisy initial depth map. 

\subsubsection{Quantitative analysis}\label{QuantitativeRo}
In order to quantitatively analyze the robustness against noise of the proposed AWGIF algorithm, the RMSE and CORR for the noise synthetic image sequences have been shown in Fig. \ref{fig:QM_synthetic}. For N-Cone images, the yellow bar in Fig. \ref{fig:QM_synthetic} (a) and (b) display that least RMSE and maximum CORR have been achieved by our proposed algorithm AWGIF. It demonstrates that the AWGIF provides the best depth map compared with the other guided filters. The RMSE and CORR values of N-Cos depth maps for different filters are shown in the yellow bar of Fig. \ref{fig:QM_synthetic} (c) and (d). The WGIF, EGIF, and the proposed AWGIF all provide smaller RMSE and larger CORR values. Particularly, the best values of RMSE and CORR have been provided by the AWGIF which has the best anti-noise capability. The yellow bar in Fig. \ref{fig:QM_synthetic} (e) and (f) show the RMSE and CORR of the N-Cos depth maps by guided filters. The large RMSE and small CORR values indicate that the initial depth map has a lot of noise. All the filters can reduce the RMSE value and improve the correlation, indicating that these filters have a good noise removal ability, especially the AWGIF has the best RMSE and CORR values which indicates that the AWGIF has the highest robustness. 

The RMSD of the real noise depth maps obtained by the guided filters are shown in Fig. \ref{fig:QM_real}(b). Blue, pink and purple bars represent the RMSD for the depth maps of N-BA, N-KB and N-KT images respectively. The RMSD values of the three groups noisy depth maps obtained by the AWGIF are all the highest, indicating that it has the strongest anti-noise performance. Compared with the GIF, WGIF, EGIF and WAGIF, the proposed AWGIF has the best robustness.

Clearly, through the above comparative analysis and robustness analysis, the proposed AWGIF algorithm outperforms other state-of-the-art guided filtering algorithms, which is effective for improving depth maps and provide the strongest robustness.

\subsection{Computational time analysis}\label{Computation}
Finally, the computational times of the improved depth map algorithms based on various guide filters are compared and analyzed. All experiments are carried by working in Matlab 2016b on a PC with Intel (R) Core @ 2.30 GHz and 8GB RAM. The computational times in seconds of each guided filters for the synthetic images (Cone, Cos, Sine, N-Cone, N-Cos, N-Sine) and the real images (BA, KB, KT, N-BA, N-KB, N-KT) have been described in Table \ref{ComputationalTime}. It can be seen that the computational time of the proposed AWGIF is slightly longer than those of GIF, WGIF and EGIF, but it's the most accurate among them. On the other hand, all of them are much faster than other filters \cite{ali2021guided}, and they are suitable for the real-time applications.

\section{Conclusions}\label{Conclusions}
In this paper, an adaptive weighted guided image filtering (AWGIF) was proposed for depth enhancement in the shape from focus. Unlike traditional SFF technologies, the proposed algorithm focused on extracting the structural information from the input images as a guidance to compensate for the incorrect depth estimation. An adaptive amplification factor of the detail layer was introduced by using the coefficient of the AWGIF to tackle the noise and enhance depth details. The experimental comparisons on synthetic and real image sequences validate the proposed method.

Many single-view depth estimation algorithms  were proposed \cite{1liz2018}. The proposed AWGIF can be applied to smoothen the background and enhance the foreground of an image \cite{1chenw2017} under the guidance of the estimated depth. This problem will be studied in our future research.

\section*{Acknowledgement}\label{cknowledgement}
\addcontentsline{toc}{section}{Acknowledgement}
The authors thank Usman Ali for kindly providing the research materials and helping us solve related problems.

This work was supported in part by the National Natural Science Foundation of China (Grant NO.61775172) and the Science Foundation of Education Department of Jiangxi Provincial (Grant No.GJJ201927).

% Numbered list
% Use the style of numbering in square brackets.
% If nothing is used, default style will be taken.
%\begin{enumerate}[a)]
%\item 
%\item 
%\item 
%\end{enumerate}  

% Unnumbered list
%\begin{itemize}
%\item 
%\item 
%\item 
%\end{itemize}  

% Description list
%\begin{description}
%\item[]
%\item[] 
%\item[] 
%\end{description}  

% Uncomment and use as the case may be
%\begin{theorem} 
%\end{theorem}

% Uncomment and use as the case may be
%\begin{lemma} 
%\end{lemma}

%% The Appendices part is started with the command \appendix;
%% appendix sections are then done as normal sections
%% \appendix

% To print the credit authorship contribution details
\printcredits

%% Loading bibliography style file
%\bibliographystyle{model1b-num-names}
\bibliographystyle{elsarticle-num}

% Loading bibliography database
\bibliography{refers/ourAWGIFref}

\vspace{1\baselineskip}
\par\noindent 
\parbox[t]{\linewidth}{
	\noindent {\bf Yuwen Li} \ received the B.S degree in Electrical Engineering from Nanchang Institute of technology, Nanchang, China, in 2009, the M.S. degree in Instrument Science and Engineering from Mechatronic Engineering School of Nanchang University, Nanchang, China, in 2012. Currently, she is pursuing the Ph.D. degree at the School of Machinery and Automation, Wuhan University of Science and Technology, Wuhan, China. Her research interests include computer vision, image processing and pattern recognition.}
\vspace{1\baselineskip}

\par\noindent 
\parbox[t]{\linewidth}{
	\noindent {\bf Zhengguo Li} \ (SM’03) received the B.Sci (Applied Mathematics). and M.Eng. (Automatic Control) from Northeastern University, Shenyang, China, in 1992 and 1995, respectively, and the Ph.D. degree (Automatic Control) from Nanyang Technological University, Singapore, in 2001. Currently, he is with the Agency for Science, Technology and Research, Singapore. His current research interests include computational photography, video processing $\&$ delivery, robotics, and switched and impulsive control.}
\vspace{1\baselineskip}

\par\noindent 
\parbox[t]{\linewidth}{
	\noindent {\bf Chaobing Zheng} \ was born in Wuhan, China. He received the M.Eng. degree in electrical and computer engineering from Wuhan University of Science and Technology, China, in 2018. At present, he is pursuing her ph.D. degree in control science and engineering in Wuhan University of Science and Technology. His current research interests include image processing and pattern recognition.}
\vspace{1\baselineskip}

\par\noindent 
\parbox[t]{\linewidth}{
	\noindent {\bf Shiqian Wu} \ (M’02-SM’05) received the B.Eng. and M.Eng. degrees from the Huazhong University of Science and Technology (HUST), Wuhan, China, in 1985 and 1988, respectively, and the Ph.D. degree from Nanyang Technological University, Singapore in 2001. He is currently a Professor with the Institute of Robotics and Intelligent Systems, School of Information Science and Engineering, Wuhan University of Science and Technology, Director, Hubei Province Key Laboratory of Intelligent Information Processing and Real-Time Industrial Systems, Wuhan, China. His current research interests include image processing, pattern recognition, machine vision and artificial intelligence.}
\vspace{1\baselineskip}

\end{document}